\documentclass[12pt]{article}
\usepackage{amsmath}
\usepackage{graphicx}
\usepackage{enumerate}
\usepackage{natbib}
\usepackage{url} 
\usepackage[flushleft]{threeparttable}
\usepackage{multirow}
\usepackage{verbatim}
\newcommand{\blind}{1}

\usepackage{varwidth}
\usepackage{amsmath}
\usepackage{numprint}
\usepackage{algorithm}
\usepackage{algpseudocode}
\usepackage{multirow}
\usepackage[pdfborder={0 0 0}, colorlinks=false]{hyperref}
\usepackage{mdframed}

\algrenewcommand\algorithmicrequire{\textbf{Input:}}
\algrenewcommand\algorithmicensure{\textbf{Output:}}
\usepackage{amsmath,amssymb}

\usepackage{mathrsfs}
\usepackage{bm, xcolor}
\usepackage{mathrsfs}
\usepackage{verbatim}
\usepackage{longtable}
\usepackage{threeparttable}
\usepackage{color}
\usepackage{amsmath}

\usepackage{enumerate}
\usepackage{amssymb,bbm}
\usepackage{amsbsy}
\usepackage{amsmath}
\usepackage{amsthm}
\usepackage{amsfonts}
\usepackage{latexsym}
\usepackage{xifthen} 
\usepackage{color}
\usepackage{manfnt}
\usepackage{ifthen}
\usepackage{bm}
\usepackage{caption}
\usepackage{subcaption}

\makeatletter
\def\singlespace{\def\baselinestretch{1}\@normalsize}

\newtheorem{lemma}{Lemma}
\newtheorem{theorem}{Theorem}

\newtheorem{corollary}{Corollary}
\@addtoreset{equation}{section}

\newtheorem{assumption}{Assumption}

\renewcommand{\hat}{\widehat}

\def\singlespace{\def\baselinestretch{1}\@normalsize}

\makeatother

\def\newpage{\vfill\eject}

\newdimen\biblioindent    \biblioindent=30pt

 at 7truept

\def\beqr{\begin{eqnarray}}
	\def\eeqr{\end{eqnarray}}
\def\beqrs{\begin{eqnarray*}}
	\def\eeqrs{\end{eqnarray*}}

\def\beq{\begin{equation}}
	\def\eeq{\end{equation}}
\def\beqn{\begin{eqnarray}}
	\def\eeqn{\end{eqnarray}}
\def\beqnn{\begin{eqnarray*}}
	\def\eeqnn{\end{eqnarray*}}

\theoremstyle{definition}

\makeatletter

\newcommand{\Rmnum}[1]{\expandafter\@slowromancap\romannumeral #1@}
\makeatother

\usepackage{chngcntr}
\usepackage{apptools}
\AtAppendix{\counterwithin{lemma}{section}}

\ifdefined\pdfminorversion\pdfminorversion=4\fi

\begin{document}
\addtocontents{toc}{\protect\setcounter{tocdepth}{-1}}

	\def\spacingset#1{\renewcommand{\baselinestretch}%
		{#1}\small\normalsize} \spacingset{1}
	\if1\blind
	{
		\title{\bf \Large A Statistical Framework for Alignment with Biased AI Feedback}
		\author{Xintao Xia\\
			Center for Data Science, Zhejiang University\\
			and \\
			Zhiqiu Xia\\
			Department of Electrical and Computer Engineering, Rutgers University\\
			and \\
			Linjun Zhang\\
			Department of Statistics, Rutgers University\\
			and \\
			Zhanrui Cai\\
			Faculty of Business and Economics, The University of Hong Kong}
		\maketitle
	} \fi
	
	\if0\blind
	{
		\bigskip
		\bigskip
		\bigskip
		\begin{center}
			{\LARGE\bf A Statistical Framework for Alignment with Biased AI Feedback}
		\end{center}
		\medskip
	} \fi
	
	\bigskip
	
	\begin{abstract}
		Modern alignment pipelines are increasingly replacing expensive human preference labels with evaluations from large language models (LLM-as-a-Judge). However, AI labels can be systematically biased compared to high-quality human feedback datasets. In this paper, we develop two debiased alignment methods within a general framework that accommodates heterogeneous response-generation distributions and external human-feedback sources. Debiased Direct Preference Optimization (DB-DPO) augments standard DPO with a residual-based correction and density-ratio reweighting to mitigate systematic bias, while retaining DPO’s computational efficiency. Debiased Identity Preference Optimization (DIPO) directly estimates human preference probabilities without imposing a parametric reward model. We provide theoretical guarantees for both methods: DB-DPO offers a practical and computationally efficient solution for large-scale alignment, whereas the DIPO preference-probability estimator attains the semiparametric efficiency bound for fixed target and reference policies under the stated regularity, nuisance, and evaluator-accuracy conditions. This efficiency claim concerns policy evaluation, not the policy-learning procedure. Empirical studies on sentiment generation, summarization, and single-turn dialogue demonstrate that the proposed methods improve the reported alignment metrics and approach oracle baselines trained on all available clean labels, including model-based proxies where human labels are unavailable.
	\end{abstract}
	
	\noindent%
	{\it Keywords:} Large language models; Preference learning; LLM-as-a-Judge; Debiasing; Statistical efficiency.
	\vfill
	
	\newpage
	\spacingset{1.9} 
	
	\section{Introduction}
	\label{sec:introduction}

	Modern large language models (LLMs) are trained through a multi-stage pipeline and require a large amount of high-quality data. After large-scale self-supervised pretraining and supervised fine-tuning, a final alignment stage is typically applied to steer model behavior toward human preferences using reinforcement learning from feedback \citep{ouyang2022training}. This alignment stage plays an important role in determining the qualitative behavior of the deployed LLMs. The dominant paradigm for alignment is \emph{Reinforcement Learning from Human Feedback} (RLHF), which operationalizes human preferences through preference-based learning \citep{christiano2017deep}. 
	
	In a standard RLHF pipeline, human annotators provide pairwise comparisons between candidate responses generated by a reference policy. These comparisons are used to train a reward model that approximates latent human preferences. Afterward, the language model is optimized to maximize the learned reward, subject to a regularization constraint that limits deviation from the reference policy. This optimization can be carried out using Proximal Policy Optimization (PPO; \citealp{schulman2017proximal}). While effective, PPO-based RLHF is known to be sensitive to reward model misspecification and requires careful tuning to achieve optimal performance. \cite{rafailov2023direct} proposed \emph{Direct Preference Optimization} (DPO), which exploits a closed-form characterization of the optimal policy under the Bradley--Terry (BT) preference model \citep{bradley1952rank} and reduces alignment to a regularized maximum likelihood estimation problem. Related work on \emph{Identity Preference Optimization} (IPO; \citealp{azar2024general}) directly maximizes preference probabilities without introducing an explicit reward function and admits a sampled optimization objective.
	
	However, collecting high-quality human annotations is usually expensive and slow. Modern alignment pipelines increasingly replace or supplement human feedback with automated AI evaluators, known as \emph{LLM-as-a-Judge} \citep{zheng2023judging,liu2023g}. Specifically, a well-calibrated LLM is prompted to evaluate candidate responses and produce pairwise preference labels or scores. Empirical studies demonstrate that, when properly prompted, LLM judges can correlate well with aggregate human judgments across a range of tasks, including reasoning, summarization, dialogue, and safety assessment \citep{wu2025meta, yu2025rip}. The ability to generate large volumes of preference data at low marginal cost has motivated \emph{Reinforcement Learning from AI Feedback} (RLAIF), where AI-generated preferences serve as the primary signal for model alignment.
	
	Despite its practical appeal, the LLM-as-a-Judge paradigm introduces fundamental challenges. Specifically, AI-generated preferences are not guaranteed to reflect true human preferences and may exhibit systematic biases. For example, \cite{zheng2023judging} showed that LLM judges can have verbosity bias (preference for longer answers), self-enhancement bias (judges favoring models similar to themselves), and position bias (preference for answers in the first position).  As a result, naively treating AI-generated preferences as unbiased substitutes for human preferences can lead to misalignment. Simply mixing a small number of human-labeled comparisons with a large volume of AI-labeled data does not, in general, correct these biases and may even reinforce them.

	Combining predictions with a labeled residual correction has a long history in statistics. Difference and generalized regression estimators in survey sampling use auxiliary information to improve estimation \citep{cassel1976generalized,sarndal1992model}. Related augmentation principles arise in inverse probability weighting for missing data \citep{robins1994estimation}. Modern prediction-powered inference \citep{angelopoulos2023prediction,angelopoulos2023ppi++} builds on this prediction-and-correction structure; its survey-sampling roots and extensions for informative labeling are discussed by \citet{mozer2026difference} and \citet{datta2025prediction}. 
	
	We study the central question: \emph{How can abundant but potentially biased AI-labeled data be efficiently combined with limited but accurate human-labeled data, so that we can recover human-aligned preferences with theoretical guarantees?} In this paper, we develop a principled statistical debiasing framework for preference-based alignment under heterogeneous feedback sources, including both human and AI datasets. Our central contribution is to address an important alignment problem: how to use abundant but biased AI feedback together with limited reliable human feedback when their response-generation laws differ. Building on ideas from sampling and semiparametric inference, the main contribution of our paper is the problem's importance and the resulting alignment framework. Specifically:
	\begin{itemize}
		\item We propose \emph{Debiased Direct Preference Optimization} (DB-DPO), a computationally efficient extension of the widely used DPO framework that augments the standard objective with a residual-based correction and density-ratio reweighting. DB-DPO mitigates systematic discrepancies between AI-judge and human preference distributions while remaining practical for large-scale alignment tasks.
		
		\item We also propose \emph{Debiased Identity Preference Optimization} (DIPO), which directly estimates preference probabilities without imposing a reward model. We demonstrate that the DIPO preference-probability estimator achieves the semiparametric efficiency bound for fixed target and reference policies under the stated regularity conditions.
		
		\item We establish regret bounds for DB-DPO and finite-step DIPO, with an explicit computational gap for DIPO, and an asymptotic expansion for fixed-policy DIPO evaluation. These results characterize the roles of evaluator bias, density-ratio estimation, policy complexity, and computation in preference alignment.
		
		\item Extensive numerical experiments further demonstrate that both methods consistently mitigate judge-induced bias and improve downstream alignment performance, outperforming naive mixtures of clean labels with biased AI feedback.
	\end{itemize}
	
	The remainder of the paper is organized as follows. Section~\ref{sec:formulation} formulates the problem and introduces the data structure. Section~\ref{sec:method} presents the DB-DPO and DIPO methodologies. Section~\ref{sec:theory} establishes the main theoretical results, including suboptimality and efficiency guarantees. Section~\ref{sec:experiments} reports empirical studies on sentiment generation, summarization, and dialogue. Section~\ref{sec:conclusion} concludes with a discussion of implications and future directions. Technical proofs and additional experimental details are deferred to the supplement.

	\section{Preliminaries}
\label{sec:formulation}
\subsection{Preference Data }

Let $\mathbb P_X$ denote the prompt distribution. For a given prompt $X\sim\mathbb P_X$, two responses $Y^{(1)}, Y^{(2)} \sim \pi_{\mathrm{Gen}}^{\mathrm{AI}}(\cdot \mid X)$ are generated independently from a response policy $\pi_{\mathrm{Gen}}^{\mathrm{AI}}$, which typically corresponds to a supervised fine-tuned language model that will be subsequently aligned. The following conditional probability represents human preferences over response pairs
\begin{equation*}
	g(Y^{(1)}, Y^{(2)} \mid X)
	:= \mathbb{P}_{\mathrm{Human}}\big(Y^{(1)} \succ Y^{(2)} \mid X\big),
\end{equation*}
where $Y^{(1)} \succ Y^{(2)}$ indicates that a human evaluator prefers the first response. The associated human preference label is defined as $Z := \mathbb{I}(Y^{(1)} \succ Y^{(2)})$, where $\mathbb{I}(\cdot)$ is the indicator function. When the preference labels are generated by an AI evaluation model (LLM-as-a-Judge), we define the corresponding label as $\hat{Z}:= \mathbb{I}(Y^{(1)} \succ Y^{(2)})$ under the probability distribution $\mathbb{P}_{\mathrm{AI}}$, where the AI evaluator induces an approximate preference probability
\begin{equation*}
	\tilde{g}(Y^{(1)}, Y^{(2)} \mid X)
	:= \mathbb{P}_{\mathrm{AI}}\big(Y^{(1)} \succ Y^{(2)} \mid X\big),
\end{equation*}
and $\tilde{g}$ denotes the AI-generated preferences, which may not align with human preferences.

For our analysis, both data sources share the prompt law $\mathbb P_X$. We consider two sources of preference data: human-labeled data and AI-labeled data. Human-labeled data consist of samples $\mathcal D_{\mathrm{Hum}}= \{(X_i, Y_i^{(1)}, Y_i^{(2)}, Z_i)\}_{i=1}^{n}$, where $Z_i$ is generated according to the human preference mechanism $g(\cdot)$. We can collect such data directly or from external sources, such as WebGPT \citep{nakano2021webgpt}, whose response-generation policy is denoted by $\pi_{\mathrm{Gen}}^{\mathrm{Hum}}$. AI-labeled data consist of samples $\mathcal{D}_{\mathrm{AI}}= \{(X_i, Y_i^{(1)}, Y_i^{(2)}, \hat{Z}_i)\}_{i=n+1}^{n+N}$. where $\hat{Z}_i\in\{0,1\}$ is generated according to the AI preference mechanism $\tilde{g}(\cdot)$. Since AI-based annotation is relatively inexpensive, our analysis focuses on regimes where $N$ is larger than $n$.

\subsection{Preference-Based Alignment}

We briefly review a standard preference-based alignment formulation. Suppose
\begin{equation}
	\label{eq:bt_model}
	\mathbb{P}(Z = 1 \mid X, Y^{(1)}, Y^{(2)})
	= \sigma\big(r(X, Y^{(1)}) - r(X, Y^{(2)})\big),
\end{equation}
where $r(X,Y)$ is a latent human reward function and $\sigma(t) = \{1+\exp(-t)\}^{-1}$ is the logistic function. Model \eqref{eq:bt_model} corresponds to the well-known Bradley--Terry preference model \citep{bradley1952rank}. Proximal Policy Optimization (PPO, \citealp{schulman2017proximal}) first fits a reward model $r_\phi$ from preference data and then solves the following problem
\begin{equation}
	\label{eq:ppo}
	\max_{\pi\in\Pi}\Big\{  \mathbb{E}_{X,Y\sim\pi(\cdot\mid X)}[r_\phi(X,Y)]
	-\beta\,D_{\mathrm{KL}}(\pi\|\pi_{\mathrm{ref}})\Big\},
\end{equation}
where $\pi_{\mathrm{ref}}$ is a fixed reference policy, $\beta$ controls the regularization strength, and $\Pi$ denotes the policy class. Throughout, $D_{\mathrm{KL}}(\pi\|\pi_{\mathrm{ref}})$ abbreviates $\mathbb E_{X\sim\mathbb P_X}D_{\mathrm{KL}}\{\pi(\cdot\mid X)\|\pi_{\mathrm{ref}}(\cdot\mid X)\}$.

Direct Preference Optimization (DPO; \citealp{rafailov2023direct}) bypasses fitting a reward model by exploiting a closed-form characterization of the optimizer of \eqref{eq:ppo} at the population level. Under standard regularity conditions, the optimal policy satisfies $\pi(Y\mid X)\ \propto\ \pi_{\mathrm{ref}}(Y\mid X)\exp\{\beta^{-1}r_\phi(X,Y)\}$, which implies the reward--policy relation
\begin{equation}
	\label{eq:reward_form}
	r_\phi(X,Y)=\beta\log\frac{\pi(Y\mid X)}{\pi_{\mathrm{ref}}(Y\mid X)}+\beta\log W(X),
\end{equation}
where $W(X)$ is a normalizing term and independent of $Y$. Substituting \eqref{eq:reward_form} into the Bradley--Terry likelihood yields the DPO objective
\begin{equation}
	\label{eq:dpo_pop}
	L_{\mathrm{DPO}}(\pi;\pi_{\mathrm{ref}})
	:=-\mathbb{E}\Bigg[
	\log\sigma\Big(
	\beta(2Z-1)\big\{\Delta_\pi(X;Y^{(1)},Y^{(2)})-\Delta_{\mathrm{ref}}(X;Y^{(1)},Y^{(2)})\big\}
	\Big)
	\Bigg],
\end{equation}
where $\Delta_\pi(X;Y^{(1)},Y^{(2)}):=\log\pi(Y^{(1)}\mid X)-\log\pi(Y^{(2)}\mid X)$ and $\Delta_{\mathrm{ref}}(X;Y^{(1)},Y^{(2)}) := \log \pi_{\mathrm{ref}}(Y^{(1)}\mid X)-\log \pi_{\mathrm{ref}}(Y^{(2)}\mid X)$. Identity Preference Optimization (IPO; \citealp{azar2024general}) provides an alternative preference-based formulation that does not rely on the BT model assumption. IPO seeks a policy that maximizes preference over a reference policy, subject to regularization:
\begin{equation*}
	\max_{\pi\in\Pi}\ 
	\mathbb{P}\big(\pi \succ \pi_{\mathrm{ref}}\big)
	-\beta D_{\mathrm{KL}}(\pi\|\pi_{\mathrm{ref}})
\end{equation*}
where $\mathbb{P}(\pi\succ\pi_{\mathrm{ref}})$ denotes the expected probability that a response sampled from policy $\pi$ is preferred by a human over a response sampled from the reference policy $\pi_{\text{ref}}$. We use DPO and IPO as the two main alignment backbones for developing debiasing methods.

\section{Debiasing Framework for Preference Alignment}
\label{sec:method}

This section develops preference-alignment methods that correct AI feedback bias.

\subsection{Debiased Direct Preference Optimization}
\label{sec:ddpo}

Let $L_{\mathrm{DPO}}(\pi;\pi_{\mathrm{ref}})$ denote the population DPO objective in \eqref{eq:dpo_pop}. For each data point,  we define the DPO negative log-likelihood contribution as
\begin{equation*}
	\ell_{\mathrm{DPO}}(\pi;X,Y^{(1)},Y^{(2)},Z)
	:=
	-\log \sigma\!\Big(
	\beta(2Z-1)\big\{\Delta_{\pi}(X;Y^{(1)},Y^{(2)})-\Delta_{\mathrm{ref}}(X;Y^{(1)},Y^{(2)})\big\}
	\Big).
\end{equation*}
The goal of DPO is to learn an optimal policy 
\begin{equation}
	\label{eq:pi*_DPO}
	\pi^*=\arg\min_{\pi\in\Pi}\mathbb{E}_{X\sim\mathbb P_X,(Y^{(1)},Y^{(2)})\sim\pi_{\text{Gen}}^{\text{AI}}}\ell_{\mathrm{DPO}}(\pi;X,Y^{(1)},Y^{(2)},Z).    
\end{equation}
Using the AI-labeled data $\mathcal{D}_{\mathrm{AI}}$, the empirical DPO loss is
\begin{align*}
	\hat L_{\mathrm{DPO}}(\pi;\pi_{\mathrm{ref}})
	:&=-\frac{1}{N}\sum_{i=n+1}^{n+N}\log\sigma\Big(
	\beta(2\hat Z_i-1)\big\{\Delta_\pi(X_i;Y_i^{(1)},Y_i^{(2)})-\Delta_{\mathrm{ref}}(X_i;Y_i^{(1)},Y_i^{(2)})\big\}
	\Big)\\
	&=\frac{1}{N}\sum_{i=n+1}^{n+N}\ell_{\mathrm{DPO}}(\pi;X_i,Y_i^{(1)},Y_i^{(2)},\hat Z_i).
\end{align*}
This objective is biased because $\hat Z_i$ is generated under the AI model $\tilde g$ rather than the ground-truth mechanism $g$. We correct this bias using the human-labeled dataset. Moreover, since human-labeled pairs may be generated under a different response mechanism $\pi_{\mathrm{Gen}}^{\mathrm{Hum}}$ than the AI-generated data $\pi_{\mathrm{Gen}}^{\mathrm{AI}}$, we further account for the resulting distributional shift by incorporating an appropriate density-ratio correction:
\begin{equation}
	\label{eq:dratio_ddpo}
	w(Y^{(1)},Y^{(2)}\mid X)
	:=
	\frac{\pi_{\mathrm{Gen}}^{\mathrm{AI}}(Y^{(1)}\mid X)}{\pi_{\mathrm{Gen}}^{\mathrm{Hum}}(Y^{(1)}\mid X)}
	\cdot
	\frac{\pi_{\mathrm{Gen}}^{\mathrm{AI}}(Y^{(2)}\mid X)}{\pi_{\mathrm{Gen}}^{\mathrm{Hum}}(Y^{(2)}\mid X)}.
\end{equation}
If $\pi_{\mathrm{Gen}}^{\mathrm{Hum}}$ is unavailable, we estimate it from data and replace $w$ with an estimator $\hat w$ constructed from the estimated $\pi_{\mathrm{Gen}}^{\mathrm{Hum}}$. We employ the sample splitting technique so that $\hat w$ is independent of the evaluation sample \citep{cai2024asymptotic, cai2022model}. For notational simplicity, we write $w_i$ for either $w(\cdot)$ or $\hat w(\cdot)$ evaluated at $(X_i,Y_i^{(1)},Y_i^{(2)})$. Thus, the bias induced by replacing $Z$ with $\hat Z$ can be estimated by the weighted difference: 
\begin{equation}
	\label{eq:dpo_bias_term}
	\hat L_{\mathrm{DPO}}^{B}(\pi;\pi_{\mathrm{ref}})
	:=
	\frac{1}{n}\sum_{i=1}^{n} w_i
	\Big\{
	\ell_{\mathrm{DPO}}(\pi;X_i,Y_i^{(1)},Y_i^{(2)},\hat Z_i)
	-
	\ell_{\mathrm{DPO}}(\pi;X_i,Y_i^{(1)},Y_i^{(2)},Z_i)
	\Big\},
\end{equation}
where $\hat{Z}_i$ is generated from $(X_i, Y_i^{(1)}, Y_i^{(2)})$ by the AI evaluator. The debiased loss is then
\begin{equation}
	\label{eq:ddpo_loss}
	\hat L_{\mathrm{DB\text{-}DPO}}(\pi;\pi_{\mathrm{ref}})
	:=
	\hat L_{\mathrm{DPO}}(\pi;\pi_{\mathrm{ref}})
	-
	\hat L_{\mathrm{DPO}}^{B}(\pi;\pi_{\mathrm{ref}}),
	\qquad
\end{equation}
and the estimated policy is defined as $\hat\pi_{\mathrm{DB\text{-}DPO}} :=\arg\min_{\pi\in\Pi}\hat L_{\mathrm{DB\text{-}DPO}}(\pi;\pi_{\mathrm{ref}})$. Algorithm \ref{alg:ddpo} summarizes the procedure.

\begin{algorithm}[ht]
	\caption{Debiased Direct Preference Optimization (DB-DPO)}
	\label{alg:ddpo}
	\begin{algorithmic}[1]
		\Require $\mathcal D_{\mathrm{Hum}}=\{(X_i,Y_i^{(1)},Y_i^{(2)},Z_i)\}_{i=1}^{n}$, $\mathcal{D}_{\mathrm{AI}}=\{(X_i,Y_i^{(1)},Y_i^{(2)},\hat Z_i)\}_{i=n+1}^{n+N}$, reference policy $\pi_{\mathrm{ref}}$, KL weight $\beta$, total steps $T$. 
		\State Obtain an AI label $\hat Z_i$ for each human comparison pair.
		\State Compute $w_i$ using \eqref{eq:dratio_ddpo} (or $w_i=\hat w_i$ via sample splitting).
		\State Form $\hat L_{\mathrm{DPO}}^{B}(\pi;\pi_{\mathrm{ref}})$ using \eqref{eq:dpo_bias_term}.
		\State Apply $T$ stochastic-gradient steps to $\hat L_{\mathrm{DB\text{-}DPO}}(\pi;\pi_{\mathrm{ref}})$ in \eqref{eq:ddpo_loss}, obtaining $\pi^{(T)}$.
		\State \Return $\pi^{(T)}$.
	\end{algorithmic}
\end{algorithm}

\subsection{Debiased Identity Preference Optimization }
\label{sec: debiased_ipo}

Although DPO provides a practical and computationally attractive approach to preference alignment, it fundamentally relies on the BT assumption. This assumption can be problematic when preferences are nearly deterministic, as it may lead to overfitting the preference data while ignoring the KL regularization term. To address the limitation, \citet{azar2024general} proposed \emph{Identity Preference Optimization} (IPO), which seeks an optimal policy $\pi^*$ by maximizing the KL-regularized preference objective
\begin{equation}
	\label{eq:online_ipo}
	\begin{aligned}
		\max_{\pi\in\Pi}\quad 
		& \mathbb{P}(\pi(\cdot)\succ \pi_{\text{ref}}(\cdot))
		- \beta D_{\text{KL}}\!\big(\pi(\cdot)\,\|\,\pi_{\text{ref}}(\cdot)\big), \\
		\text{with}\quad
		& \mathbb{P}(\pi(\cdot)\succ \pi_{\text{ref}}(\cdot))
		:= \mathbb{E}_{X\sim \mathbb P_X}\!\left[
		\mathbb{E}_{Y\sim\pi(\cdot\mid X), Y^{\prime}\sim\pi_{\text{ref}}(\cdot\mid X)}
		\mathbb{P}_{\text{Human}}(Y\succ Y^{\prime}\mid X)
		\right].
	\end{aligned}
\end{equation}
Write $P(\pi)=\mathbb P(\pi(\cdot)\succ\pi_{\mathrm{ref}}(\cdot))$. The DIPO direct sample uses the common prompt law and newly generated target and reference responses. In our setting, we employ an AI evaluator to estimate the preference probability. For each prompt $X$, the AI evaluator outputs either a binary preference label $\hat{Z}$ or a pair of evaluation scores $s_{\mathrm{eva}}(X, Y^{(1)})$ and $s_{\mathrm{eva}}(X, Y^{(2)})$ for the candidate responses. Specifically, for each prompt $X_i$, we generate a response $Y_i \sim \pi(\cdot \mid X_i)$ from the target policy and two responses $Y_i^{(1)},Y_i^{(2)} \sim \pi_{\mathrm{ref}}(\cdot \mid X_i)$ from the reference policy. Then the AI evaluator compares the responses and outputs a preference probability. This procedure yields the following direct estimator:
\begin{equation}
	\label{eq:PDM}
	\hat{P}_{\text{DM}}(\pi):=\frac{1}{2N}\sum_{i=n+1}^{n+N}\{\tilde{g}(Y_i, Y_i^{(1)}\mid X_i)+\tilde{g}( Y_i, Y_i^{(2)}\mid X_i)\}.
\end{equation}
The estimator $\hat{P}_{\mathrm{DM}}(\pi)$ directly estimates the unobservable human preference probability. However, it may be systematically biased when the AI evaluator is misaligned with human preferences. Given the human preference dataset $\mathcal D_{\mathrm{Hum}}$, we aim to construct a debiased estimator of the human preference probability that accounts for the discrepancy between human and AI preferences. Recall that $Z$ and $\hat{Z}$ represent the human- and AI-labeled preference labels. Here, we characterize the bias as follows:
\begin{equation*}
	\begin{split}
		&\mathbb E\{\hat{P}_{\text{DM}}(\pi)\}-\mathbb{P}(\pi(\cdot)\succ \pi_{\text{ref}}(\cdot))=\mathbb{E}_{X\sim\mathbb P_X}[\mathbb{E}_{Y^{(1)}\sim\pi(\cdot\mid X),Y^{(2)}\sim\pi_{\text{ref}}(\cdot\mid X)}(\hat{Z}-Z)]\\
		&=\mathbb{E}_{X\sim\mathbb P_X}[\mathbb{E}_{(Y^{(1)},Y^{(2)})\sim\pi_{\text{Gen}}^{\text{Hum}}(\cdot\mid X)}(\hat{Z}-Z)\frac{\pi(Y^{(1)}\mid X)}{\pi_{\text{Gen}}^{\text{Hum}}(Y^{(1)}\mid X)}\frac{\pi_{\text{ref}}(Y^{(2)}\mid X)}{\pi_{\text{Gen}}^{\text{Hum}}(Y^{(2)}\mid X)}].
	\end{split}
\end{equation*}
This expression represents the bias as a weighted expectation of $\hat{Z}-Z$, where the weight adjusts for the distributional mismatch between the human- and AI-labeled datasets. Exchangeability of the human response pair and these identities justify symmetrizing over $(Y^{(1)},Y^{(2)})$. Thus, we obtain the following bias estimator based on the dataset $\mathcal D_{\mathrm{Hum}}$:
\begin{equation*}
	\begin{split}
		\widehat{\mathrm{Bias}}(\pi):=&\frac{1}{2n}\sum_{i=1}^{n}(\hat{Z}_i-Z_i)\frac{\pi(Y_i^{(1)}\mid X_i)}{\pi_{\text{Gen}}^{\text{Hum}}(Y_i^{(1)}\mid X_i)}\frac{\pi_{\text{ref}}(Y_i^{(2)}\mid X_i)}{\pi_{\text{Gen}}^{\text{Hum}}(Y_i^{(2)}\mid X_i)}\\
		&+\frac{1}{2n}\sum_{i=1}^{n}(Z_i-\hat{Z}_i)\frac{\pi(Y_i^{(2)}\mid X_i)}{\pi_{\text{Gen}}^{\text{Hum}}(Y_i^{(2)}\mid X_i)}\frac{\pi_{\text{ref}}(Y_i^{(1)}\mid X_i)}{\pi_{\text{Gen}}^{\text{Hum}}(Y_i^{(1)}\mid X_i)}.
	\end{split}
\end{equation*}
When the human generation policy is unknown, replace it with an estimate fitted on an independent nuisance-training sample. We define a debiased estimator of the IPO preference term by subtracting the estimated bias,
\begin{equation}\label{eq:est_dipo}
	\hat{P}_{\text{DIPO}}(\pi):=\hat{P}_{\text{DM}}(\pi)-\widehat{\text{Bias}}(\pi).
\end{equation}

For policy learning, let
$$
\overline P_{\mathrm{DM}}(\pi)
=\mathbb E_{X,\,Y\sim\pi(\cdot\mid X),\,Y'\sim\pi_{\mathrm{ref}}(\cdot\mid X)}
\tilde g(Y,Y'\mid X),
$$
where the expectation conditions on independently fitted nuisances. We seek to maximize the conditional mean objective
\begin{equation}\label{eq:dipo}
	\overline J(\pi)
	=\overline P_{\mathrm{DM}}(\pi)-\widehat{\mathrm{Bias}}(\pi)
	-\beta D_{\mathrm{KL}}(\pi\|\pi_{\mathrm{ref}}),\qquad \pi\in\Pi.
\end{equation}
The human sample is held fixed in this objective. Algorithm~\ref{alg:dipo} estimates its direct gradient using fresh prompts and responses at each step and returns $\hat\pi_{\mathrm{DIPO}}=\pi^{(T)}$. Theorem~\ref{thm:regret_dipo} bounds this finite-step output with an explicit computational gap. The estimator in \eqref{eq:est_dipo} remains the object of fixed-policy evaluation. The theoretical results use untruncated weights and the preference scores in \eqref{eq:PDM}. The final evaluation does not update the policy. Including this evaluation, the procedure uses $T+1$ fresh batches, each of size $N$.

\begin{algorithm}[ht]
	\caption{Debiased Identity Preference Optimization (DIPO)}
	\label{alg:dipo}
	\begin{algorithmic}[1]
		\Require Reference policy $\pi_{\text{ref}}$, human-labeled dataset $\mathcal D_{\mathrm{Hum}}=\{(X_i,Y^{(1)}_i,Y^{(2)}_i,Z_i,\hat Z_i)\}_{i=1}^n$, KL weight $\beta$, AI batch size $N$, clipping range $[c_{\min},c_{\max}]$, step sizes $\{\eta_t\}_{t=1}^T$, total steps $T$.
		\State Use the known human generation policy, or an estimate fitted on independent data.
		\State Initialize policy $\pi^{(0)}\leftarrow \pi_{\text{ref}}$.
		\For{$t=1,\dots,T$}
		\State Sample a set of prompts $\{X_i\}_{i=n+1}^{n+N}$ from the prompt distribution $\mathbb P_X$.
		\State Sampling: $\{Y_i \sim \pi^{(t-1)}(\cdot \mid X_i)\}_{i=n+1}^{n+N}$ and $\{(Y_i^{(1)},Y_i^{(2)}) \sim \pi_{\text{ref}}(\cdot \mid X_i)\}_{i=n+1}^{n+N}$.
		\State Compute the direct-model preference estimate, $\hat{P}_{\mathrm{DM}}(\pi^{(t-1)})$, as defined in \eqref{eq:PDM}.
		\State Compute importance ratios with clipping
		\[
		w^{(1)}_i =\text{clip}\!\left(\frac{\pi^{(t-1)}(Y_i^{(1)}\mid X_i)}{\pi_{\text{Gen}}^{\text{Hum}}(Y_i^{(1)}\mid X_i)}\frac{\pi_{\text{ref}}(Y_i^{(2)}\mid X_i)}{\pi_{\text{Gen}}^{\text{Hum}}(Y_i^{(2)}\mid X_i)},\,c_{\min},\,c_{\max}\right),
		\]
		\[
		w^{(2)}_i = \text{clip}\!\left(\frac{\pi^{(t-1)}(Y_i^{(2)}\mid X_i)}{\pi_{\text{Gen}}^{\text{Hum}}(Y_i^{(2)}\mid X_i)}\frac{\pi_{\text{ref}}(Y_i^{(1)}\mid X_i)}{\pi_{\text{Gen}}^{\text{Hum}}(Y_i^{(1)}\mid X_i)},\,c_{\min},\,c_{\max}\right).
		\]
		\[
		\widehat{\mathrm{Bias}}^{(t-1)}
		\;=\;
		\frac{1}{2n}\sum_{i=1}^{n}(\hat Z_i- Z_i)\,w^{(1)}_i
		\;+\;
		\frac{1}{2n}\sum_{i=1}^{n}( Z_i-\hat Z_i)\,w^{(2)}_i.
		\]
		\State Record the corrected sampled value minus the KL penalty at $\pi^{(t-1)}$.
		\State Update the policy parameters by a gradient-ascent step of size $\eta_t$.
		\EndFor
		\State Evaluate $\pi^{(T)}$ on an additional fresh batch of $N$ prompts and responses and record its corrected, regularized score.
		\State \Return Policy iterate $\pi^{(T)}$.
	\end{algorithmic}
\end{algorithm}

For policy training, we also consider an additive human-feedback term when $\pi_{\mathrm{Gen}}^{\mathrm{Hum}}$ is known. Motivated by the use of tunable combinations in PPI++ \citep{angelopoulos2023ppi++}, let
\begin{align*}
	\widehat{P}_{\text{Hum}}(\pi):=&\frac{1}{2n}\sum_{i=1}^{n}\bigg[Z_i\frac{\pi(Y_i^{(1)}\mid X_i)}{\pi_{\text{Gen}}^{\text{Hum}}(Y^{(1)}_i\mid X_i)}\frac{\pi_{\text{ref}}(Y^{(2)}_i\mid X_i)}{\pi_{\text{Gen}}^{\text{Hum}}(Y^{(2)}_i\mid X_i)}\\
	&+(1-Z_i)\frac{\pi(Y_i^{(2)}\mid X_i)}{\pi_{\text{Gen}}^{\text{Hum}}(Y^{(2)}_i\mid X_i)}\frac{\pi_{\text{ref}}(Y^{(1)}_i\mid X_i)}{\pi_{\text{Gen}}^{\text{Hum}}(Y^{(1)}_i\mid X_i)}\bigg]
\end{align*}
be an estimator of $\mathbb{P}(\pi(\cdot)\succ \pi_{\text{ref}}(\cdot))$. Note that we have
\begin{align*}
	&\mathbb{E}_{X\sim\mathbb P_X,(Y^{(1)},Y^{(2)})\sim\pi_{\text{Gen}}^{\text{Hum}}(\cdot\mid X)}\bigg[Z\frac{\pi(Y^{(1)}\mid X)}{\pi_{\text{Gen}}^{\text{Hum}}(Y^{(1)}\mid X)}\frac{\pi_{\text{ref}}(Y^{(2)}\mid X)}{\pi_{\text{Gen}}^{\text{Hum}}(Y^{(2)}\mid X)}\bigg]\\
	=&\mathbb{E}_{X\sim\mathbb P_X,Y^{(1)}\sim\pi(\cdot\mid X),Y^{(2)}\sim\pi_{\text{ref}}(\cdot\mid X)}[Z]=\mathbb{P}(\pi(\cdot)\succ \pi_{\text{ref}}(\cdot))
\end{align*}
and
\begin{align*}
	&\mathbb{E}_{X\sim\mathbb P_X,(Y^{(1)},Y^{(2)})\sim\pi_{\text{Gen}}^{\text{Hum}}(\cdot\mid X)}\bigg[(1-Z)\frac{\pi(Y^{(2)}\mid X)}{\pi_{\text{Gen}}^{\text{Hum}}(Y^{(2)}\mid X)}\frac{\pi_{\text{ref}}(Y^{(1)}\mid X)}{\pi_{\text{Gen}}^{\text{Hum}}(Y^{(1)}\mid X)}\bigg]\\
	=&\mathbb{E}_{X\sim\mathbb P_X,Y^{(1)}\sim\pi(\cdot\mid X),Y^{(2)}\sim\pi_{\text{ref}}(\cdot\mid X)}[Z]=\mathbb{P}(\pi(\cdot)\succ \pi_{\text{ref}}(\cdot)).
\end{align*}
Thus, we have $\mathbb{E}(\widehat{P}_{\text{Hum}}(\pi))=\mathbb{P}(\pi(\cdot)\succ \pi_{\text{ref}}(\cdot))$. For a tuning parameter $\lambda$, define the additive training objective
\begin{equation}
	\widehat{P}_{\text{DIPO}}^{+}(\pi):=\widehat{P}_{\text{DIPO}}(\pi)+\lambda \widehat{P}_{\text{Hum}}(\pi).
\end{equation}

The parameter $\lambda$ controls the contribution of the human-only term. When both component estimators are unbiased, and $\lambda$ is fixed, the expectation of this unnormalized objective is $(1+\lambda)P(\pi)$. Thus, it generally does not estimate the preference probability for nonzero $\lambda$. For fixed $\lambda\ge0$, the corresponding population training objective with unchanged KL coefficient is $(1+\lambda)P(\pi)-\beta D_{\mathrm{KL}}(\pi\|\pi_{\mathrm{ref}})$. Its maximizer is therefore that of $P(\pi)-\{\beta/(1+\lambda)\}D_{\mathrm{KL}}(\pi\|\pi_{\mathrm{ref}})$. Thus $\lambda=1$ halves the effective KL coefficient: comparisons at the same nominal $\beta$ also reflect this change in regularization. We do not claim a finite-sample variance improvement for this additive modification. The tuning parameter may be selected on separate validation data according to policy performance.

\section{Theoretical Properties}
\label{sec:theory}

\subsection{Theoretical Results for DB-DPO}
\label{sec:thm_ddpo}

We begin by stating a set of regularity conditions.

\begin{assumption}[Reward realizability]\label{assump:realizability}
	Let $\Pi$ denote the policy class. The oracle reward function $r^{*}(\cdot,\cdot)$ belongs to the reward class $\mathcal{R}=\{\beta \log\{\pi(Y\mid X)/\pi_{\mathrm{ref}}(Y\mid X)\}+ \beta\log W(X):\pi \in\Pi\}$, where $W(X)>0$ is a common prompt-dependent normalization term, and the reward class is uniformly bounded: $\sup_{r\in\mathcal R}\|r\|_\infty\le C<\infty$.
\end{assumption}

\begin{assumption}[Coverage and finite-moment weights]\label{assump:coverage}
	The reference policy satisfies $\pi_{\mathrm{ref}}(Y\mid X)\ge\epsilon>0$, and $\pi_{\mathrm{Gen}}^{\mathrm{AI}}(Y\mid X)\ge\epsilon\pi_{\mathrm{ref}}(Y\mid X)$. The $\pi_{\mathrm{Gen}}^{\mathrm{AI}}(\cdot\mid X)$ is absolutely continuous with respect to the $\pi_{\mathrm{Gen}}^{\mathrm{Hum}}(\cdot\mid X)$. Let $w=w(Y^{(1)},Y^{(2)}\mid X)$ be defined in \eqref{eq:dratio_ddpo} and $\hat w$ be its independently fitted estimator. Assume $\mathbb E[w^\delta+\hat w^\delta]\le M_\delta<\infty$ for some $\delta>2$.
\end{assumption}

\begin{assumption}[Model complexity]\label{assump:model-complexity}
	The DB-DPO criterion classes and the DIPO human-correction class induced by $\Pi$ satisfy the ordinary Rademacher complexity bounds in Appendix~\ref{sec:rad-complexity}, with a finite complexity index $v\ge1$. Rademacher complexity is defined as in \citet{bartlett2002rademacher}.
\end{assumption}

\begin{assumption}[Alignment between AI and human preferences]\label{assump:aif_hf}
	The AI evaluation model and human preferences are aligned in the sense that $\mathbb{P}(Z \neq \hat Z \mid X, Y^{(1)}, Y^{(2)}) = O_p(\nu_n)$.
\end{assumption}

Assumption~\ref{assump:realizability} expresses the oracle reward as a scaled log-ratio of the optimal and reference policies, up to an $X$-dependent term, yielding population equivalence between PPO and DPO under the BT model \citep{rafailov2023direct}. Assumption~\ref{assump:coverage} ensures response coverage while allowing unbounded weights with finite $\delta$-th moments for $\delta>2$. Related overlap conditions are standard in causal inference \citep{imbens2015causal} and off-policy evaluation \citep{uehara2022review,zhan2024policy}. In Assumption~\ref{assump:model-complexity}, the Rademacher complexity index $v$ governs uniform convergence and suboptimality rates. Assumption~\ref{assump:aif_hf} quantifies AI--human label agreement. We allow $\nu_n$ to depend on $n$ to model a regime in which AI evaluators improve with increasing training data as the study size grows. Smaller $\nu_n$ corresponds to closer alignment and fewer human labels needed for bias correction, while constant $\nu_n$ allows persistent disagreement. The norm-based Rademacher bounds of \citet{bartlett2017spectrally} for DNNs and \citet{edelman2022inductive} for Transformer self-attention provide examples for Assumption~\ref{assump:model-complexity}. These bounds have no explicit parameter-count factor; the norm quantities and remaining logarithmic dimension and sample-size factors can be included in $v$. Appendix~\ref{sec:rad-examples} summarizes the two results and verifies the conditions.

\begin{theorem}[Suboptimality gap for DB-DPO]\label{thm:subopt-gap}
	Under the BT model \eqref{eq:bt_model} and Assumptions~\ref{assump:realizability}--\ref{assump:aif_hf}, the suboptimality gap of the DB-DPO estimator satisfies
	\begin{equation}\label{eq:subopt-dpo}
		\begin{split}
			&\Big|\mathbb{E}_{X\sim\mathbb P_X}\Big[\mathbb{E}_{Y\sim\pi^{*}(\cdot\mid X)} r(X,Y)-\mathbb{E}_{Y\sim\hat\pi_{\mathrm{DB\text{-}DPO}}(\cdot\mid X)} r(X,Y)\Big]\Big| \\
			=&O_p\bigg(\frac{1}{\beta\epsilon}\bigg(\frac{v}{N}\bigg)^{1/4}+\frac{1}{\beta\epsilon}\bigg(\frac{v\nu_n}{n}\bigg)^{1/4}+\frac{1}{\beta\epsilon}\nu_n^{1/2}\,\|\hat w - w\|_2\bigg),
		\end{split}
	\end{equation}
	where $\pi^{*}$ denotes the oracle optimal policy defined in \eqref{eq:pi*_DPO} and
	$$
	\|\hat w - w\|_2:=\Bigg\{\mathbb{E}_{X\sim\mathbb P_X,\,(Y^{(1)},Y^{(2)})\sim\pi_{\mathrm{Gen}}^{\mathrm{Hum}}(\cdot\mid X)}\big[w(Y^{(1)},Y^{(2)}\mid X)-\hat w(Y^{(1)},Y^{(2)}\mid X)\big]^2\Bigg\}^{1/2}.
	$$
\end{theorem}

Theorem~\ref{thm:subopt-gap} gives a stochastic-order bound on the reward-value difference relative to the regularized oracle. The first term reflects estimation error from the AI-labeled dataset of size $N$, and the second term captures the contribution of the human-labeled dataset of size $n$, scaled by the disagreement rate $\nu_n$. The third term quantifies the impact of density-ratio estimation. It vanishes when the density ratio is known and used, including when the two generation policies are known to coincide. This term is analogous to the cross-term appearing in double machine learning estimators \citep{chernozhukov2018double}. The factor $1/\beta$ reflects policy sensitivity to reward error, while the DB-DPO loss derivative with respect to the log-policy contrast contains $\beta$. Thus, choosing $\beta$ also depends on the policy class and optimization procedure; see \citet{wu2024beta,xu2025doubly}. In many practical settings, the AI-labeled dataset is substantially larger than the human-labeled dataset, while the disagreement rate $\nu_n$ is small but non-negligible. In this regime, the debiased estimator substantially reduces the bias inherited from the AI evaluation model. 

\begin{corollary}[DB-DPO under local Rademacher complexity]
	\label{cor:ddpo-local}
	Under the conditions of Theorem~\ref{thm:subopt-gap}, assume also the local Rademacher bounds in Appendix~\ref{sec:local-ddpo}. Then
	\begin{equation}\label{eq:subopt-dpo-local}
		\begin{split}
			&\bigg|\mathbb E_{X\sim\mathbb P_X}\bigg[ \mathbb E_{Y\sim\pi^*(\cdot\mid X)}r(X,Y) -\mathbb E_{Y\sim\hat\pi_{\mathrm{DB\text{-}DPO}}(\cdot\mid X)}r(X,Y)\bigg]\bigg|\\
			=&O_p\bigg[\frac1{\beta\epsilon}\bigg\{ \sqrt{\frac vN}+\sqrt{\frac{v\nu_n}{n}}+\nu_n^{1/2}\|\hat w-w\|_2\bigg\}\bigg].
		\end{split}
	\end{equation}
\end{corollary}

Corollary~\ref{cor:ddpo-local} improves the fourth-root sampling terms in Theorem~\ref{thm:subopt-gap} to square-root terms under the local Rademacher bounds in Appendix \ref{sec:local-ddpo}, following the localization framework of \citet{bartlett2005local}. Appendix~\ref{sec:local-ddpo} verifies these conditions for deep ReLU DNNs and multilayer, multihead Transformers with frozen pretrained representations and trainable linear reward heads, under the stated regularity conditions.

\subsection{Theoretical Results for DIPO}

We begin by characterizing the DIPO estimator through an influence function expansion \citep{van2000asymptotic}, which clarifies how bias correction and variance reduction arise from combining AI-labeled and human-labeled preference data. We first consider the estimation of the preference probability $\mathbb{P}(\pi(\cdot)\succ\pi_{\mathrm{ref}}(\cdot))$ using \emph{only human-labeled data}. For fixed-policy evaluation, we consider the score function defined by
\begin{equation}
	\label{eq:influence}
	\begin{split}
		&\psi(X,Y^{(1)},Y^{(2)},Z;\pi,\pi_{\mathrm{ref}},\pi_{\mathrm{Gen}}^{\mathrm{Hum}},g):=\mathbb{E}_{Y\sim\pi(\cdot\mid X),\,Y'\sim\pi_{\mathrm{ref}}(\cdot\mid X)}
		\big\{g(Y,Y'\mid X)\big\} \\
		&+\frac{1}{2}\{Z-g(Y^{(1)},Y^{(2)}\mid X)\}
		\frac{\pi(Y^{(1)}\mid X)\pi_{\mathrm{ref}}(Y^{(2)}\mid X)
			-\pi(Y^{(2)}\mid X)\pi_{\mathrm{ref}}(Y^{(1)}\mid X)}
		{\pi_{\mathrm{Gen}}^{\mathrm{Hum}}(Y^{(1)}\mid X)\pi_{\mathrm{Gen}}^{\mathrm{Hum}}(Y^{(2)}\mid X)} .
	\end{split}
\end{equation}
The first term corresponds to a direct plug-in estimator of the preference probability, while the second term augments this estimate with a weighted residual. This augmentation corrects for model misspecification and yields a doubly robust estimating equation: consistency holds if either $g$ or $\pi_{\mathrm{Gen}}^{\mathrm{Hum}}$ is known or correctly specified.

\begin{lemma}\label{lem:semi}
	Suppose $n$ i.i.d.\ samples $\{(X_i,Y_i^{(1)},Y_i^{(2)},Z_i)\}_{i=1}^n$ are observed from $\mathcal D_{\mathrm{Hum}}$. Keep $\pi$ and $\pi_{\mathrm{ref}}$ fixed. Suppose that Assumption~\ref{assump:weakcoverage} hold, $0<g(y_2,y_1\mid x)=1-g(y_1,y_2\mid x)<1$ almost surely. If $\psi(\cdot)$ has finite second moment, the efficient influence function for estimating $P(\pi)$ is $\psi(X,Y^{(1)},Y^{(2)},Z;\pi,\pi_{\mathrm{ref}},\pi_{\mathrm{Gen}}^{\mathrm{Hum}},g)-P(\pi)$.
\end{lemma}

Lemma~\ref{lem:semi} centers the score in \eqref{eq:influence} to obtain the influence function for the fixed-policy preference probability. In contrast to \citet{xu2025doubly}, we treat $\pi$ and $\pi_{\mathrm{ref}}$ as known and regard $g$ and $\pi_{\mathrm{Gen}}^{\mathrm{Hum}}$ as nuisance components. The score in \eqref{eq:influence} naturally motivates the DIPO estimator as a debiased construction: the first term provides a model-based estimate via $g$, and the second term serves as a bias correction, removing systematic bias induced by AI evaluation. We impose the following additional assumptions.

\begin{assumption}[Finite-moment coverage]\label{assump:weakcoverage}
	For every $\pi\in\Pi$, we have $\pi(\cdot\mid X),\pi_{\mathrm{ref}}(\cdot\mid X)\ll\pi_{\mathrm{Gen}}^{\mathrm{Hum}}(\cdot\mid X)$ with positive fitted probabilities on the support. The envelope $\bar w$ of the true and fitted weights defined in \eqref{eq:dipo-weight-moment} satisfies $\mathbb E\bar w^\delta\le M_{\delta}<\infty$ for $\delta>2$ almost surely. The optimal policy $\pi^*$ defined in \eqref{eq:online_ipo} is contained in the policy class $\Pi$.
\end{assumption}

Assumption~\ref{assump:weakcoverage} permits unbounded importance weights and imposes no uniform lower bound on generation probabilities. The moment condition concerns the full pairwise weights uniformly over policies. It also ensures that the policy class contains the regularized population optimizer.

\begin{theorem}[Asymptotic expansion of DIPO]\label{thm:mse}
	Under Assumptions~\ref{assump:aif_hf}, \ref{assump:weakcoverage}, fix $\pi\in\Pi$. Conditional on independent nuisance fitting, the human and AI evaluation samples are independent, with i.i.d. observations within each source. The estimator $\hat P_{\mathrm{DIPO}}(\pi)$ defined in \eqref{eq:est_dipo} admits the expansion
	\begin{equation}
		\label{eq:mse}
		\begin{split}
			&\hat P_{\mathrm{DIPO}}(\pi)\\
			=&
			\frac{1}{2}\sum_{a=1}^2\frac{1}{N}\sum_{i=n+1}^{n+N}
			g(Y_i,Y_i^{(a)}\mid X_i) \\
			&+\frac{1}{2n}\sum_{i=1}^n
			\frac{\pi(Y_i^{(1)}\mid X_i)\pi_{\mathrm{ref}}(Y_i^{(2)}\mid X_i)
				-\pi_{\mathrm{ref}}(Y_i^{(1)}\mid X_i)\pi(Y_i^{(2)}\mid X_i)}
			{\pi_{\mathrm{Gen}}^{\mathrm{Hum}}(Y_i^{(1)}\mid X_i)\pi_{\mathrm{Gen}}^{\mathrm{Hum}}(Y_i^{(2)}\mid X_i)}
			\{Z_i-g(Y_i^{(1)},Y_i^{(2)}\mid X_i)\} \\
			&-\frac{1}{2n}\sum_{i=1}^n
			\frac{\pi(Y_i^{(1)}\mid X_i)\pi_{\mathrm{ref}}(Y_i^{(2)}\mid X_i)
				-\pi_{\mathrm{ref}}(Y_i^{(1)}\mid X_i)\pi(Y_i^{(2)}\mid X_i)}
			{\pi_{\mathrm{Gen}}^{\mathrm{Hum}}(Y_i^{(1)}\mid X_i)\pi_{\mathrm{Gen}}^{\mathrm{Hum}}(Y_i^{(2)}\mid X_i)}
			\{\hat Z_i-\tilde g(Y_i^{(1)},Y_i^{(2)}\mid X_i)\}\\
			&+R ,
		\end{split}
	\end{equation}
	where, conditionally on the nuisance training data, with probability at least $1-\alpha$,
	\begin{equation*}
		|R|\le\frac{C}{\sqrt\alpha}\bigg[
		\frac{\|\tilde g-g\|}{\sqrt N}
		+\frac{M_{\delta}^{1/\delta}
			\|\tilde g-g\|_{2\delta/(\delta-2)}
			+e_{\delta}}{\sqrt n}\bigg]+e_{\delta}
		\|\tilde g-g\|_{\delta/(\delta-1)}.
	\end{equation*}
	Here $e_{\delta}$ is defined in \eqref{eq:moment-dipo-nuisance}, $\|h\|_{q}=(\mathbb E|h|^q)^{1/q}$,
	and the original target-law norm is
	$$
	\|\tilde g-g\|^2=\mathbb E_{X,Y\sim\pi(\cdot\mid X),Y'\sim\pi_{\mathrm{ref}}(\cdot\mid X)}|\tilde g(Y,Y'\mid X)-g(Y,Y'\mid X)|^2.
	$$
\end{theorem}

Theorem~\ref{thm:mse} concerns the original estimator with a binary AI label in the human correction. Its third leading term records AI evaluation noise and is generally of order $n^{-1/2}$. The remainder depends on the sample sizes $(n,N)$ and the quality of the nuisance estimators $(\tilde g,\widehat\pi_{\mathrm{Gen}}^{\mathrm{Hum}})$. The preference-model error is measured under the target and human comparison laws, while $e_{\delta}$ measures error in the joint importance weights. When the generation mechanism is known, $e_{\delta}=0$. The final product term captures the interaction between these two sources of error and is characteristic of debiased estimation procedures \citep{chernozhukov2018double}. We next compare the exact-probability estimator of the fixed-policy preference probability with an efficient human-only estimator. For fixed-policy evaluation, we define an estimator using exact evaluator probabilities by
\begin{equation}\label{eq:dipo-probability}
	\begin{aligned}
		\hat P_{\mathrm{DIPO}}^{\mathrm{prob}}(\pi)
		&=\hat P_{\mathrm{DM}}(\pi)
		-\frac{1}{2n}\sum_{i=1}^n
		\{\tilde g(Y_i^{(1)},Y_i^{(2)}\mid X_i)-Z_i\}\\
		&\quad\times
		\frac{
			\pi(Y_i^{(1)}\mid X_i)\pi_{\mathrm{ref}}(Y_i^{(2)}\mid X_i)
			-\pi(Y_i^{(2)}\mid X_i)\pi_{\mathrm{ref}}(Y_i^{(1)}\mid X_i)
		}{
			\hat\pi_{\mathrm{Gen}}^{\mathrm{Hum}}(Y_i^{(1)}\mid X_i)
			\hat\pi_{\mathrm{Gen}}^{\mathrm{Hum}}(Y_i^{(2)}\mid X_i)
		}.
	\end{aligned}
\end{equation}
Here $\hat\pi_{\mathrm{Gen}}^{\mathrm{Hum}}$ is fitted on independent data and replaced by $\pi_{\mathrm{Gen}}^{\mathrm{Hum}}$ when known. The evaluator probabilities $\tilde g$ are distinct from the binary labels $\hat Z_i$.

\begin{corollary}[Efficiency of fixed-policy evaluation]
	\label{cor:efficiency_comparison}
	Under the conditions of Theorem~\ref{thm:mse} and Lemma~\ref{lem:semi}, keep both $\pi$ and $\pi_{\mathrm{ref}}$ fixed and use $\hat P_{\mathrm{DIPO}}^{\mathrm{prob}}(\pi)$ from \eqref{eq:dipo-probability}. Suppose $\|\tilde g-g\|+\|\tilde g-g\|_{2\delta/(\delta-2)}+e_{\delta}=o_p(1)$ and $e_{\delta}\|\tilde g-g\|_{\delta/(\delta-1)}=o_{L^2}(n^{-1/2})$, where $o_{L^2}$ refers to mean-square convergence over the training data. Let $\hat P_{\mathrm{IPO}}(\pi)$ be a regular human-only estimator with the efficient influence function in Lemma~\ref{lem:semi}. Define $\mu_\pi(X)=\mathbb E_{Y\sim\pi(\cdot\mid X) Y'\sim\pi_{\mathrm{ref}}(\cdot\mid X)}g(Y,Y'\mid X)$, $V_X=\operatorname{Var}\{\mu_\pi(X)\}$, and $V_{\mathrm A}=\operatorname{Var}\{\sum_{a=1}^2g(Y,Y^{(a)}\mid X)/2\}$. If $\liminf_{n,N\to\infty}\{V_X-(n/N)V_{\mathrm A}\}>0$, then, for all sufficiently large $n,N$,
	$$
	\mathrm{MSE}(\hat P_{\mathrm{DIPO}}^{\mathrm{prob}}(\pi))<\mathrm{MSE}(\hat P_{\mathrm{IPO}}(\pi)).
	$$
	In particular, this condition holds when $N/n\to\infty$ and $V_X>0$. 
	
	Additionally, assume that the estimator is regular under local perturbations of the evaluation law, conditional on the independent nuisance-training data. When $N/n\to\infty$, $\hat P_{\mathrm{DIPO}}^{\mathrm{prob}}(\pi)$ attains the limiting semiparametric efficiency bound for fixed-policy evaluation based on the $n$ human observations and $N$ additional prompts.
\end{corollary}

Theorem~\ref{thm:mse} concerns the original binary-label estimator at a fixed policy; Corollary~\ref{cor:efficiency_comparison} concerns its exact-probability version. Appendix~\ref{sec:computational-errors} gives the Monte Carlo extension. Finally, we establish a regret bound for the policy returned after $T$ updates with fresh AI sampling. The conditional mean objective is
\begin{align*}
	\overline J(\pi)=\mathbb E_{X\sim\mathbb P_X,\,Y\sim\pi(\cdot\mid X),\,
		Y'\sim\pi_{\mathrm{ref}}(\cdot\mid X)}
	\tilde g(Y,Y'\mid X)-\widehat{\mathrm{Bias}}(\pi)
	-\beta D_{\mathrm{KL}}(\pi\|\pi_{\mathrm{ref}}),
\end{align*}
where the expectation holds the fitted nuisances and the human sample fixed. Define its fresh-batch score by
\begin{equation}\label{eq:dipo-checkpoint-score}
	\begin{split}
		\widehat J_t=\frac1{2N}\sum_{i=n+1}^{n+N}\sum_{a=1}^2\tilde g(Y_{i,t},Y_{i,t}^{(a)}\mid X_{i,t})-\widehat{\mathrm{Bias}}(\pi^{(t)})-\beta D_{\mathrm{KL}}(\pi^{(t)}\|\pi_{\mathrm{ref}}),\qquad 0\le t\le T.
	\end{split}
\end{equation}

\begin{theorem}[Regret bound for finite-step DIPO]\label{thm:regret_dipo}
	Under Assumptions~\ref{assump:model-complexity} and~\ref{assump:weakcoverage}, use independently fitted nuisances and suppose $K:=\sup_{\pi\in\Pi}D_{\mathrm{KL}}(\pi\|\pi_{\mathrm{ref}})<\infty$. Let $T\ge1$ be a deterministic iteration budget and $\{\pi^{(t)}\}_{t=0}^T\subset\Pi$ follow the sampling strategy in Algorithm~\ref{alg:dipo} with observations within each batch conditionally i.i.d.\ given the fitted nuisances, human sample, and preceding history. Define the nonnegative computational gap by
	\begin{equation}\label{eq:dipo-computation-gap}
		\varepsilon_{\mathrm{opt},T}=\bigg\{\sup_{\pi\in\Pi}\overline J(\pi)-\max_{0\le t\le T}\overline J(\pi^{(t)})\bigg\}+\bigg\{\max_{0\le t\le T}\widehat J_t-\widehat J_T\bigg\}.
	\end{equation}
	Then the final iterate satisfies. 
	\begin{align*}
		&\left|P(\pi^*)
		-P(\pi^{(T)})\right|\\
		&\quad\le \varepsilon_{\mathrm{opt},T}
		+O_p\bigg\{\beta K+\sqrt{\frac vn}+\frac vn
		+\sqrt{\frac{\log\{2(T+1)\}}N}
		+e_{\delta}\|\tilde g-g\|_{\delta/(\delta-1)}\bigg\}.
	\end{align*}
\end{theorem}

Theorem~\ref{thm:regret_dipo} bounds the regret of the final DIPO iterate. The first term of \eqref{eq:dipo-computation-gap} measures how closely the best visited policy approaches the supremum of the conditional mean objective $\overline J$ over $\Pi$. The second term of \eqref{eq:dipo-computation-gap} is the final iterate's empirical score shortfall relative to the best evaluated checkpoint; it is zero when $\widehat J_T=\max_{0\le t\le T}\widehat J_t$. It is not assumed to vanish merely because $T$ updates are performed. The AI term follows from conditional concentration across the $T+1$ fresh evaluations; $N$ is the size of each batch, not the total number of AI observations. The bound allows nonvanishing nuisance errors. If the generation mechanism is known and its true weights are used, the nuisance-product term is zero even when the evaluator remains biased. With consistent estimated weights, this term vanishes without evaluator consistency because $\|\tilde g-g\|_{\delta/(\delta-1)}\le1$.

Appendix~\ref{sec:moment-weights} extends the policy bounds to $1<\delta<2$, including the local refinement for DB-DPO. Appendix~\ref{sec:computational-errors} quantifies evaluator-label Monte Carlo error for fixed-policy evaluation, defines the finite-step computational gaps, and bounds the effects of approximate criterion values. Appendix~\ref{sec:dipo-implementation} proves the conditional score-function gradient identity for \eqref{eq:dipo}; this identity does not itself bound the computational gap.

\section{Numerical Experiments}
\label{sec:experiments}

We evaluate DB-DPO and DIPO on controlled sentiment generation, summarization, and single-turn dialogue using biased AI feedback and a limited set of reliable target labels. The experiments introduce position, verbosity, and self-enhancement biases, respectively, and use estimated density ratios. We compare the proposed methods with mixed-label baselines and study the effects of density-ratio correction and target-label availability. Section~\ref{sec:main-ablations} examines the effects of density-ratio correction, the target-labeled data fraction, the DIPO+ tuning parameter $\lambda$, and importance-weight clipping thresholds.

\subsection{Experimental Setup}

\paragraph{Tasks and feedback sources.}
We use IMDb for sentiment generation, TL;DR for summarization, and Anthropic HH for single-turn dialogue \citep{imdb,summarization_data,anthropichh}. For sentiment generation, the policy completes a movie-review prefix with positive sentiment. The AI evaluator \texttt{twitter-\allowbreak roberta-\allowbreak base-\allowbreak sentiment-\allowbreak latest} \citep{loureiro2022timelms} is adjusted to favor the first-listed response when the two sentiment scores are close. It agrees with the gold reward model $63.0\%$ of the time on the AI-labeled data. For summarization, \texttt{Qwen3-1.7B} \citep{qwen3} receives a length-based adjustment favoring longer summaries, yielding $63.1\%$ agreement with the target labels. For dialogue, each AI comparison pair contains responses sampled at temperatures $0.1$ and $1.0$. The evaluator \texttt{Qwen3-1.7B} is adjusted to favor responses with higher log-likelihood under its own model, yielding $67.1\%$ agreement with the target labels. These constructions introduce controlled, response-dependent evaluator biases.

For sentiment generation, while the IMDb dataset lacks native preference pairs, we construct the target-labeled pairs from reviews that share common prefixes and annotate them using a gold reward model, \texttt{sentiment-\allowbreak roberta-\allowbreak large-\allowbreak english} \citep{siebert}. For summarization and dialogue, we use \emph{claude-haiku-5.5} (``Haiku'') as a proxy for human annotators to provide preference labels for the response pairs in $\mathcal D_{\mathrm{Hum}}$. These labels represent the model's preferences rather than direct human annotations. Target-labeled pairs come from existing preference datasets, whereas AI pairs are sampled from the SFT policy. We estimate density ratios using a cross-fitted classifier distinguishing the two pair sources, following the sample-splitting construction in Section~\ref{sec:method}. The training details are described in Appendix~\ref{sec:training-details}. The proposed methods use fitted weights throughout the main comparisons, with a default target-labeled fraction of $20\%$. Section~\ref{sec:main-ablations} examines estimated versus unit weights and varies the target-labeled fraction.

\paragraph{Methods and evaluation.}
We separate offline and online methods. DPO and Sampled IPO are offline baselines using the available mixed labels; online IPO is the corresponding baseline for DIPO. DPO* and IPO* use clean target labels throughout and serve as empirical oracle benchmarks. DIPO+ adds the target-only term, implemented with fitted weights. Its default is $\lambda=1$, except in the sensitivity study. DIPO and DIPO+ use the default importance-weight clipping interval $[0.8,1.25]$. All methods use common evaluation settings within each task; the offline and online training procedures remain distinct.

Sentiment evaluation uses the gold reward model on $5{,}000$ held-out prefixes, reporting the average sentiment score and win rate against SFT continuations at temperature $1.0$. Summarization evaluates summaries generated at temperature $0.1$ on $5{,}888$ held-out prompts against the dataset references. Dialogue evaluates responses generated at temperature $0.7$ on $2{,}602$ held-out prompts against the dataset's chosen responses. For summarization and dialogue, evaluation under Haiku measures alignment with the proxy target preferences. We also report win rates under \emph{gpt-6-luna} (``GPT''), excluded from data generation and training, to assess transfer across judges. Both judges are queried without reasoning, and candidate order is randomized to reduce systematic position bias.

The tables report sample means with approximate $95\%$ confidence-interval half-widths, calculated as $1.96s/\sqrt m$, where $m$ is the number of test prompts and $s$ is the sample standard deviation of the prompt-level scores. These intervals quantify evaluation uncertainty conditional on a single trained policy and do not measure variability across training runs.

\subsection{Main Results}

\paragraph{Controlled sentiment generation.}
Table~\ref{tab:revised-sentiment} reports average sentiment scores and win rates against SFT. DB-DPO improves over mixed-label DPO from $92.62\%$ to $95.59\%$ in average score and from $83.13\%$ to $86.83\%$ in win rate. DPO* remains higher on both metrics, and DB-DPO narrows the gap using limited target labels. In the online comparison, DIPO improves the average score from $77.99\%$ to $81.36\%$ and the win rate from $70.22\%$ to $73.52\%$. DIPO+ yields further gains, reaching $95.49\%$ and $86.77\%$, respectively. Thus both the bias correction and the additional target-label term improve the observed performance under the constructed position bias.

\begin{table}[!htbp]
	\centering
	\begin{tabular}{lcc}
		\hline
		Models & Average Score & Win Rate \\ \hline
		SFT                      & 51.89\%{\scriptsize$\pm$1.4} & 50.00\% \\ \hline
		\multicolumn{3}{l}{\textit{Offline}} \\
		DPO              & 92.62\%{\scriptsize$\pm$0.7} & 83.13\%{\scriptsize$\pm$1.0} \\
		Sampled IPO              & 74.88\%{\scriptsize$\pm$1.2} & 69.24\%{\scriptsize$\pm$1.3} \\
		DPO*                     & 97.47\%{\scriptsize$\pm$0.4} & 89.91\%{\scriptsize$\pm$0.8} \\
		DB-DPO                     & 95.59\%{\scriptsize$\pm$0.6} & 86.83\%{\scriptsize$\pm$0.9} \\ \hline
		\multicolumn{3}{l}{\textit{Online}} \\
		IPO                      & 77.99\%{\scriptsize$\pm$1.1} & 70.22\%{\scriptsize$\pm$1.3} \\
		IPO*                     & 94.89\%{\scriptsize$\pm$0.6} & 89.48\%{\scriptsize$\pm$0.8} \\
		DIPO                     & 81.36\%{\scriptsize$\pm$1.1} & 73.52\%{\scriptsize$\pm$1.2} \\
		DIPO+                    & 95.49\%{\scriptsize$\pm$0.6} & 86.77\%{\scriptsize$\pm$0.9} \\ \hline
	\end{tabular}
	\caption{Sentiment generation under position-biased AI feedback. Scores and win rates are percentages; win rates compare policy and SFT continuations at temperature $1.0$. Values after $\pm$ are the reported $95\%$ confidence-interval half-widths.}
	\label{tab:revised-sentiment}
\end{table}

\paragraph{Summarization.}
Table~\ref{tab:revised-summarization} shows that DB-DPO improves over mixed-label DPO under both judges: the Haiku win rate increases from $71.81\%$ to $74.59\%$, and the GPT win rate increases from $71.94\%$ to $74.10\%$. DIPO improves the Haiku win rate from $65.58\%$ for online IPO to $69.83\%$, while its GPT result is slightly lower, $67.03\%$ versus $67.65\%$. DIPO+ reaches $75.75\%$ under Haiku and $72.21\%$ under GPT, improving over online IPO under both evaluators. The DPO* and IPO* benchmarks remain higher than their corresponding debiased methods under both judges.

\begin{table}[!htbp]
	\centering
	\begin{tabular}{lcc}
		\hline
		Models & Win Rate (Haiku) & Win Rate (GPT) \\ \hline
		SFT    & 49.86\%{\scriptsize$\pm$1.3} & 48.81\%{\scriptsize$\pm$1.3} \\ \hline
		\multicolumn{3}{l}{\textit{Offline}} \\
		DPO              & 71.81\%{\scriptsize$\pm$1.1} & 71.94\%{\scriptsize$\pm$1.1} \\
		Sampled IPO             & 65.74\%{\scriptsize$\pm$1.2} & 63.60\%{\scriptsize$\pm$1.2} \\
		DPO*                     & 81.56\%{\scriptsize$\pm$1.0} & 77.16\%{\scriptsize$\pm$1.1} \\
		DB-DPO                   & 74.59\%{\scriptsize$\pm$1.1} & 74.10\%{\scriptsize$\pm$1.1} \\ \hline
		\multicolumn{3}{l}{\textit{Online}} \\
		IPO                      & 65.58\%{\scriptsize$\pm$1.2} & 67.65\%{\scriptsize$\pm$1.2} \\
		IPO*                     & 78.31\%{\scriptsize$\pm$1.1} & 77.16\%{\scriptsize$\pm$1.1} \\
		DIPO                     & 69.83\%{\scriptsize$\pm$1.2} & 67.03\%{\scriptsize$\pm$1.2} \\
		DIPO+                    & 75.75\%{\scriptsize$\pm$1.1} & 72.21\%{\scriptsize$\pm$1.1} \\ \hline
	\end{tabular}
	\caption{Summarization under verbosity-biased AI feedback. Win rates (\%) compare generated summaries with reference summaries at temperature $0.1$. Values after $\pm$ are the reported $95\%$ confidence-interval half-widths.}
	\label{tab:revised-summarization}
\end{table}

\paragraph{Single-turn dialogue.}
Table~\ref{tab:revised-dialogue} reports results under the self-enhancement bias. DB-DPO improves over mixed-label DPO from $57.03\%$ to $59.67\%$ under Haiku and from $53.77\%$ to $55.38\%$ under GPT. DIPO improves the Haiku win rate from $58.84\%$ to $60.59\%$, while its GPT win rate changes from $54.73\%$ to $57.07\%$. DIPO+ obtains $62.68\%$ and $57.61\%$, exceeding online IPO under both judges. 

Overall, DB-DPO and DIPO+ achieve higher observed scores than their corresponding mixed-label baselines across the reported tasks and evaluators. DIPO also improves over IPO under Haiku in both summarization and dialogue, although it does not uniformly outperform IPO under GPT. Because Haiku supplies the proxy target labels, improvements under Haiku support the intended objective of alignment with that preference source. GPT evaluation separately assesses transfer across evaluators, where gains need not be uniform. These results support the practical value of combining biased AI feedback with limited human or proxy annotations, while highlighting how performance depends on the evaluation criterion.

\begin{table}[!htbp]
	\centering
	\begin{tabular}{lcc}
		\hline
		Models & Win Rate (Haiku) & Win Rate (GPT) \\ \hline
		SFT    & 51.10\%{\scriptsize$\pm$1.9} & 52.50\%{\scriptsize$\pm$1.9} \\ \hline
		\multicolumn{3}{l}{\textit{Offline}} \\
		DPO              & 57.03\%{\scriptsize$\pm$1.9} & 53.77\%{\scriptsize$\pm$1.9} \\
		Sampled IPO              & 56.74\%{\scriptsize$\pm$1.9} & 54.96\%{\scriptsize$\pm$1.9} \\
		DPO*                     & 64.49\%{\scriptsize$\pm$1.8} & 58.92\%{\scriptsize$\pm$1.9} \\
		DB-DPO                   & 59.67\%{\scriptsize$\pm$1.9} & 55.38\%{\scriptsize$\pm$1.9} \\ \hline
		\multicolumn{3}{l}{\textit{Online}} \\
		IPO                      & 58.84\%{\scriptsize$\pm$1.9} & 54.73\%{\scriptsize$\pm$1.9} \\
		IPO*                     & 63.59\%{\scriptsize$\pm$1.8} & 59.07\%{\scriptsize$\pm$1.9} \\
		DIPO                     & 60.59\%{\scriptsize$\pm$1.9} & 57.07\%{\scriptsize$\pm$1.9} \\
		DIPO+                    & 62.68\%{\scriptsize$\pm$1.9} & 57.61\%{\scriptsize$\pm$1.9} \\ \hline
	\end{tabular}
	\caption{Dialogue under self-enhancement-biased AI feedback. Win rates (\%) compare generated responses with the dataset's chosen responses at temperature $0.7$. Values after $\pm$ are the reported $95\%$ confidence-interval half-widths.}
	\label{tab:revised-dialogue}
\end{table}

\paragraph{Importance weighting.}
\label{sec:main-ablations}\label{sec:revised-ablations}
\label{sec:ablation-weights}
Table~\ref{tab:ablation-weights} compares fitted and unit density ratios on the summarization task. In the unit setting, we set the classifier-estimated density ratio in \eqref{eq:dratio_ddpo} to one. DIPO and DIPO+ retain the policy-to-reference likelihood ratio $\pi_\theta(y\mid x)/\pi_{\mathrm{ref}}(y\mid x)$. For each method, using the fitted density ratio yields higher observed win rates under both judges. For DIPO, the Haiku and GPT win rates increase from $68.78\%$ and $65.15\%$ to $69.83\%$ and $67.03\%$, respectively. Some differences are small relative to the reported uncertainty. This comparison measures the effect of reweighting for differences between the two response sources.

\begin{table}[htbp]
	\centering
	\begin{tabular}{llcc}
		\hline
		Method & Density ratio & Win Rate (Haiku) & Win Rate (GPT) \\ \hline
		DB-DPO & Unit    & 73.99\%{\scriptsize$\pm$1.1} & 73.57\%{\scriptsize$\pm$1.1} \\
		DB-DPO & Fitted  & 74.59\%{\scriptsize$\pm$1.1} & 74.10\%{\scriptsize$\pm$1.1} \\ \hline
		DIPO   & Unit    & 68.78\%{\scriptsize$\pm$1.2} & 65.15\%{\scriptsize$\pm$1.2} \\
		DIPO   & Fitted  & 69.83\%{\scriptsize$\pm$1.2} & 67.03\%{\scriptsize$\pm$1.2} \\ \hline
		DIPO+  & Unit    & 74.62\%{\scriptsize$\pm$1.1} & 71.60\%{\scriptsize$\pm$1.2} \\
		DIPO+  & Fitted  & 75.75\%{\scriptsize$\pm$1.1} & 72.21\%{\scriptsize$\pm$1.1} \\ \hline
	\end{tabular}
	\caption{Density ratio ablation on summarization. Unit sets the classifier-estimated density ratio to one. Fitted uses its estimated value. DIPO and DIPO+ retain the policy-to-reference likelihood ratio in both settings. Win rates are reported as percentages with approximate 95\% confidence-interval half-widths in percentage points.}
	\label{tab:ablation-weights}
\end{table}

\paragraph{Target-label availability.}
\label{sec:ablation-human}
Using a fixed pool of $14{,}764$ summarization prompts, we vary the target-labeled fraction over $5\%$, $10\%$, $20\%$, and $30\%$, keeping the remaining settings fixed. Table~\ref{tab:ablation-human} shows that DB-DPO and DIPO+ improve over their corresponding mixed-label baselines at every tested fraction under both judges, including at $5\%$. Increasing the fraction improves DB-DPO under both judges and DIPO+ under Haiku. DIPO also improves over IPO under Haiku at each fraction, but its performance is not monotone and its GPT win rate decreases at $30\%$. The contribution of additional target annotations therefore varies across methods and evaluators.

\begin{table}[htbp]
	\centering
	\renewcommand{\baselinestretch}{1}\selectfont
	\begin{tabular}{lcccc}
		\hline
		\multicolumn{5}{c}{Win Rate (Haiku)} \\
		Method & 5\% & 10\% & 20\% & 30\% \\ \hline
		\multicolumn{5}{l}{\textit{Offline}} \\
		DPO & 67.10{\scriptsize$\pm$1.2} & 67.87{\scriptsize$\pm$1.2} & 71.81{\scriptsize$\pm$1.1} & 72.49{\scriptsize$\pm$1.1} \\
		DPO* & \multicolumn{4}{c}{81.56{\scriptsize$\pm$1.0}} \\
		DB-DPO & 69.50{\scriptsize$\pm$1.2} & 73.35{\scriptsize$\pm$1.1} & 74.59{\scriptsize$\pm$1.1} & 76.25{\scriptsize$\pm$1.1} \\ \hline
		\multicolumn{5}{l}{\textit{Online}} \\
		IPO & 65.36{\scriptsize$\pm$1.2} & 65.51{\scriptsize$\pm$1.2} & 65.58{\scriptsize$\pm$1.2} & 67.91{\scriptsize$\pm$1.2} \\
		IPO* & \multicolumn{4}{c}{78.31{\scriptsize$\pm$1.1}} \\
		DIPO & 68.98{\scriptsize$\pm$1.2} & 69.12{\scriptsize$\pm$1.2} & 69.83{\scriptsize$\pm$1.2} & 68.16{\scriptsize$\pm$1.2} \\
		DIPO+ & 72.04{\scriptsize$\pm$1.1} & 75.19{\scriptsize$\pm$1.1} & 75.75{\scriptsize$\pm$1.1} & 76.55{\scriptsize$\pm$1.1} \\ \hline
	\end{tabular}
	\par\medskip
	\begin{tabular}{lcccc}
		\hline
		\multicolumn{5}{c}{Win Rate (GPT)} \\
		Method & 5\% & 10\% & 20\% & 30\% \\ \hline
		\multicolumn{5}{l}{\textit{Offline}} \\
		DPO & 69.48{\scriptsize$\pm$1.2} & 70.30{\scriptsize$\pm$1.2} & 71.94{\scriptsize$\pm$1.1} & 72.86{\scriptsize$\pm$1.1} \\
		DPO* & \multicolumn{4}{c}{77.16{\scriptsize$\pm$1.1}} \\
		DB-DPO & 69.89{\scriptsize$\pm$1.2} & 72.50{\scriptsize$\pm$1.1} & 74.10{\scriptsize$\pm$1.1} & 74.80{\scriptsize$\pm$1.1} \\ \hline
		\multicolumn{5}{l}{\textit{Online}} \\
		IPO & 67.27{\scriptsize$\pm$1.2} & 66.30{\scriptsize$\pm$1.2} & 67.65{\scriptsize$\pm$1.2} & 68.78{\scriptsize$\pm$1.2} \\
		IPO* & \multicolumn{4}{c}{77.16{\scriptsize$\pm$1.1}} \\
		DIPO & 68.16{\scriptsize$\pm$1.2} & 65.88{\scriptsize$\pm$1.2} & 67.03{\scriptsize$\pm$1.2} & 63.50{\scriptsize$\pm$1.2} \\
		DIPO+ & 71.60{\scriptsize$\pm$1.2} & 72.30{\scriptsize$\pm$1.1} & 72.21{\scriptsize$\pm$1.1} & 71.91{\scriptsize$\pm$1.1} \\ \hline
	\end{tabular}
	\caption{Sensitivity to the fraction of target-labeled data on summarization. All entries are percentages, with approximate 95\% confidence-interval half-widths in percentage points.}
	\label{tab:ablation-human}
\end{table}

\paragraph{The DIPO+ tuning parameter.}
\label{sec:ablation-lambda}
Table~\ref{tab:ablation-lambda} varies $\lambda\in\{0,0.5,1,1.5\}$ in the composite objective $\widehat P_{\mathrm{DIPO}}+\lambda\widehat P_{\mathrm{Hum}}$, using estimated density ratios and holding the remaining settings fixed. The case $\lambda=0$ is DIPO, while $\lambda=1$ is the default DIPO+ configuration. Adding the target-only term improves the observed win rates under both judges. Increasing $\lambda$ from $1$ to $1.5$ leaves Haiku performance nearly unchanged and modestly improves the GPT result. This supports $\lambda=1$ as a reasonable default without establishing its optimality.

\begin{table}[htbp]
	\centering
	\begin{tabular}{lcc}
		\hline
		Method & Win Rate (Haiku) & Win Rate (GPT) \\ \hline
		\multicolumn{3}{l}{\textit{$\widehat P_{\mathrm{DIPO}}+\lambda\widehat P_{\mathrm{Hum}}$ with $\widehat w$}} \\
		$\lambda=0$ (DIPO)             & 69.83\%{\scriptsize$\pm$1.2} & 67.03\%{\scriptsize$\pm$1.2} \\
		$\lambda=0.5$                  & 72.63\%{\scriptsize$\pm$1.1} & 69.60\%{\scriptsize$\pm$1.2} \\
		$\lambda=1$ (DIPO+)            & 75.75\%{\scriptsize$\pm$1.1} & 72.21\%{\scriptsize$\pm$1.1} \\
		$\lambda=1.5$                  & 75.70\%{\scriptsize$\pm$1.1} & 73.06\%{\scriptsize$\pm$1.1} \\ \hline
	\end{tabular}
	\caption{Sensitivity to the DIPO+ tuning parameter on summarization, using estimated density ratios. Entries are win rates with approximate 95\% confidence-interval half-widths.}
	\label{tab:ablation-lambda}
\end{table}

\paragraph{Importance-weight clipping.}
\label{sec:ablation-clipping}
Table~\ref{tab:ablation-clipping} compares the clipping intervals $[c,1/c]$ for $c\in\{0.9,0.8,0.7\}$ in Algorithm~\ref{alg:dipo}, keeping the fitted density ratios and remaining settings fixed. The default $[0.8,1.25]$ gives the highest observed DIPO+ win rates under both judges. DIPO has its highest Haiku win rate at $c=0.9$ and its highest GPT win rate at $c=0.8$, with modest differences between these two settings. Widening the interval to $[0.7,1/0.7]$ reduces the observed win rates for both methods under both judges. These results characterize sensitivity within the tested ranges; they do not imply that a single clipping interval is optimal across tasks or evaluators.

\begin{table}[htbp]
	\centering 
	\begin{tabular}{llcc}
		\hline
		Method & Clip bound & Win Rate (Haiku) & Win Rate (GPT) \\ \hline
		DIPO   & $[0.9, 1/0.9]$           & 70.14\%{\scriptsize$\pm$1.2} & 66.90\%{\scriptsize$\pm$1.2} \\
		DIPO   & $[0.8, 1.25]$            & 69.83\%{\scriptsize$\pm$1.2} & 67.03\%{\scriptsize$\pm$1.2} \\
		DIPO   & $[0.7, 1/0.7]$           & 68.79\%{\scriptsize$\pm$1.2} & 65.39\%{\scriptsize$\pm$1.2} \\ \hline
		DIPO+  & $[0.9, 1/0.9]$           & 74.73\%{\scriptsize$\pm$1.1} & 71.82\%{\scriptsize$\pm$1.1} \\
		DIPO+  & $[0.8, 1.25]$            & 75.75\%{\scriptsize$\pm$1.1} & 72.21\%{\scriptsize$\pm$1.1} \\
		DIPO+  & $[0.7, 1/0.7]$           & 74.50\%{\scriptsize$\pm$1.1} & 71.60\%{\scriptsize$\pm$1.2} \\ \hline
	\end{tabular}
	\caption{Sensitivity to importance-weight clipping on summarization. Entries are win rates with approximate 95\% confidence-interval half-widths.}
	\label{tab:ablation-clipping}
\end{table}

\section{Conclusion}
\label{sec:conclusion}

This paper investigates preference-based alignment of large language models under heterogeneous and potentially biased feedback sources. Motivated by the widespread adoption of LLM-as-a-Judge in reinforcement learning from AI feedback, we develop a statistical debiasing framework that integrates abundant AI-generated preference labels with a limited number of reliable human annotations. We propose two methods, Debiased Direct Preference Optimization (DB-DPO) and Debiased Identity Preference Optimization (DIPO), tailored to modern preference optimization objectives. Our theoretical analysis establishes suboptimality bounds for DB-DPO, fixed-policy asymptotic expansions and efficiency guarantees for DIPO, and a regret bound for finite-step DIPO with an explicit computational gap. Empirical results across sentiment generation, summarization, and dialogue tasks demonstrate that the proposed methods substantially reduce misalignment induced by biased AI feedback and narrow the gap to clean-label oracle baselines. Some experiments use model-based proxies for human labels, and their win rates measure agreement with the specified evaluator.

Several directions for future work remain. Extending the proposed debiasing framework to multi-turn dialogue and long-horizon reinforcement learning is an important challenge, as is relaxing the assumption that human-labeled comparisons are independently obtained. Incorporating adaptive or active data collection strategies and studying interactions between preference debiasing and safety or normative constraints are also promising avenues. More broadly, as AI-generated feedback becomes an important component of scalable alignment pipelines, statistical debiasing techniques will play critical roles in robust and reliable model alignment.

\bibliographystyle{apalike}
\bibliography{ref}

@article{christiano2017deep,
  title={Deep reinforcement learning from human preferences},
  author={Christiano, Paul F and Leike, Jan and Brown, Tom and Martic, Miljan and Legg, Shane and Amodei, Dario},
  journal={Advances in neural information processing systems},
  volume={30},
  year={2017}
}

@book{van2000asymptotic,
  title={Asymptotic statistics},
  author={van der Vaart, Aad W},
  volume={3},
  year={2000},
  publisher={Cambridge university press}
}

@article{chernozhukov2018double,
  title={Double/debiased machine learning for treatment and structural parameters: Double/debiased machine learning},
  author={Chernozhukov, Victor and Chetverikov, Denis and Demirer, Mert and Duflo, Esther and Hansen, Christian and Newey, Whitney and Robins, James},
  journal={The Econometrics Journal},
  volume={21},
  number={1},
  year={2018},
  publisher={Oxford University Press}
}

@inproceedings{azar2024general,
  title={A general theoretical paradigm to understand learning from human preferences},
  author={Azar, Mohammad Gheshlaghi and Guo, Zhaohan Daniel and Piot, Bilal and Munos, Remi and Rowland, Mark and Valko, Michal and Calandriello, Daniele},
  booktitle={International Conference on Artificial Intelligence and Statistics},
  pages={4447--4455},
  year={2024},
  organization={PMLR}
}

@article{xu2025doubly,
  title={Doubly Robust Alignment for Large Language Models},
  author={Xu, Erhan and Ye, Kai and Zhou, Hongyi and Zhu, Luhan and Quinzan, Francesco and Shi, Chengchun},
  journal={arXiv preprint arXiv:2506.01183},
  year={2025}
}

@article{rafailov2023direct,
  title={Direct preference optimization: Your language model is secretly a reward model},
  author={Rafailov, Rafael and Sharma, Archit and Mitchell, Eric and Manning, Christopher D and Ermon, Stefano and Finn, Chelsea},
  journal={Advances in neural information processing systems},
  volume={36},
  pages={53728--53741},
  year={2023}
}

@article{yu2025rip,
  title={Rip: Better models by survival of the fittest prompts},
  author={Yu, Ping and Yuan, Weizhe and Golovneva, Olga and Wu, Tianhao and Sukhbaatar, Sainbayar and Weston, Jason and Xu, Jing},
  journal={arXiv preprint arXiv:2501.18578},
  year={2025}
}

@book{imbens2015causal,
  title={Causal inference in statistics, social, and biomedical sciences},
  author={Imbens, Guido W and Rubin, Donald B},
  year={2015},
  publisher={Cambridge university press}
}

@article{uehara2022review,
  title={A review of off-policy evaluation in reinforcement learning},
  author={Uehara, Masatoshi and Shi, Chengchun and Kallus, Nathan},
  journal={arXiv preprint arXiv:2212.06355},
  year={2022}
}

@article{zhan2024policy,
  title={Policy learning with adaptively collected data},
  author={Zhan, Ruohan and Ren, Zhimei and Athey, Susan and Zhou, Zhengyuan},
  journal={Management Science},
  volume={70},
  number={8},
  pages={5270--5297},
  year={2024},
  publisher={INFORMS}
}

@article{zheng2023judging,
  title={Judging llm-as-a-judge with mt-bench and chatbot arena},
  author={Zheng, Lianmin and Chiang, Wei-Lin and Sheng, Ying and Zhuang, Siyuan and Wu, Zhanghao and Zhuang, Yonghao and Lin, Zi and Li, Zhuohan and Li, Dacheng and Xing, Eric and others},
  journal={Advances in neural information processing systems},
  volume={36},
  pages={46595--46623},
  year={2023}
}

@article{ouyang2022training,
  title={Training language models to follow instructions with human feedback},
  author={Ouyang, Long and Wu, Jeffrey and Jiang, Xu and Almeida, Diogo and Wainwright, Carroll and Mishkin, Pamela and Zhang, Chong and Agarwal, Sandhini and Slama, Katarina and Ray, Alex and others},
  journal={Advances in neural information processing systems},
  volume={35},
  pages={27730--27744},
  year={2022}
}

@article{schulman2017proximal,
  title={Proximal policy optimization algorithms},
  author={Schulman, John and Wolski, Filip and Dhariwal, Prafulla and Radford, Alec and Klimov, Oleg},
  journal={arXiv preprint arXiv:1707.06347},
  year={2017}
}

@article{nakano2021webgpt,
  title={Webgpt: Browser-assisted question-answering with human feedback},
  author={Nakano, Reiichiro and Hilton, Jacob and Balaji, Suchir and Wu, Jeff and Ouyang, Long and Kim, Christina and Hesse, Christopher and Jain, Shantanu and Kosaraju, Vineet and Saunders, William and others},
  journal={arXiv preprint arXiv:2112.09332},
  year={2021}
}

@article{angelopoulos2023prediction,
  title={Prediction-powered inference},
  author={Angelopoulos, Anastasios N and Bates, Stephen and Fannjiang, Clara and Jordan, Michael I and Zrnic, Tijana},
  journal={Science},
  volume={382},
  number={6671},
  pages={669--674},
  year={2023},
  publisher={American Association for the Advancement of Science}
}

@article{angelopoulos2023ppi++,
  title={Ppi++: Efficient prediction-powered inference},
  author={Angelopoulos, Anastasios N and Duchi, John C and Zrnic, Tijana},
  journal={arXiv preprint arXiv:2311.01453},
  year={2023}
}

@article{wu2024beta,
   title={$\beta$-DPO: Direct Preference Optimization with Dynamic $\beta$},
  author={Wu, Junkang and Xie, Yuexiang and Yang, Zhengyi and Wu, Jiancan and Gao, Jinyang and Ding, Bolin and Wang, Xiang and He, Xiangnan},
  journal={Advances in Neural Information Processing Systems},
  volume={37},
  pages={129944--129966},
  year={2024}
}

@inproceedings{imdb,
    title = "Learning Word Vectors for Sentiment Analysis",
    author = "Maas, Andrew L.  and
      Daly, Raymond E.  and
      Pham, Peter T.  and
      Huang, Dan  and
      Ng, Andrew Y.  and
      Potts, Christopher",
    editor = "Lin, Dekang  and
      Matsumoto, Yuji  and
      Mihalcea, Rada",
    booktitle = "Proceedings of the 49th Annual Meeting of the Association for Computational Linguistics: Human Language Technologies",
    month = jun,
    year = "2011",
    address = "Portland, Oregon, USA",
    publisher = "Association for Computational Linguistics",
    url = "https://aclanthology.org/P11-1015/",
    pages = "142--150"
}

@article{gpt_neo,
  title={The Pile: An 800GB Dataset of Diverse Text for Language Modeling},
  author={Gao, Leo and Biderman, Stella and Black, Sid and Golding, Laurence and Hoppe, Travis and Foster, Charles and Phang, Jason and He, Horace and Thite, Anish and Nabeshima, Noa and others},
  journal={arXiv preprint arXiv:2101.00027},
  year={2020}
}

@article{siebert,
title = {More than a Feeling: Accuracy and Application of Sentiment Analysis},
journal = {International Journal of Research in Marketing},
volume = {40},
number = {1},
pages = {75-87},
year = {2023},
issn = {0167-8116},
doi = {https://doi.org/10.1016/j.ijresmar.2022.05.005},
url = {https://www.sciencedirect.com/science/article/pii/S0167811622000477},
author = {Jochen Hartmann and Mark Heitmann and Christian Siebert and Christina Schamp}
}

@misc{qwen25,
      title={Qwen2.5 Technical Report}, 
      author={Qwen and An Yang and Baosong Yang and Beichen Zhang and Binyuan Hui and Bo Zheng and Bowen Yu and Chengyuan Li and Dayiheng Liu and Fei Huang and Haoran Wei and Huan Lin and Jian Yang and Jianhong Tu and Jianwei Zhang and Jianxin Yang and Jiaxi Yang and Jingren Zhou and Junyang Lin and Kai Dang and Keming Lu and Keqin Bao and Kexin Yang and Le Yu and Mei Li and Mingfeng Xue and Pei Zhang and Qin Zhu and Rui Men and Runji Lin and Tianhao Li and Tianyi Tang and Tingyu Xia and Xingzhang Ren and Xuancheng Ren and Yang Fan and Yang Su and Yichang Zhang and Yu Wan and Yuqiong Liu and Zeyu Cui and Zhenru Zhang and Zihan Qiu},
      year={2025},
      eprint={2412.15115},
      archivePrefix={arXiv},
      primaryClass={cs.CL},
      url={https://arxiv.org/abs/2412.15115}, 
}

@inproceedings{summarization_data,
  author = {Nisan Stiennon and Long Ouyang and Jeff Wu and Daniel M. Ziegler and Ryan Lowe and Chelsea Voss and Alec Radford and Dario Amodei and Paul Christiano},
  title = {Learning to summarize from human feedback},
  booktitle = {NeurIPS},
  year = 2020,
}

@inproceedings{summarization_sft_data,
title={The N+ Implementation Details of {RLHF} with {PPO}: A Case Study on {TL};{DR} Summarization},
author={Shengyi Huang and Michael Noukhovitch and Arian Hosseini and Kashif Rasul and Weixun Wang and Lewis Tunstall},
booktitle={First Conference on Language Modeling},
year={2024},
url={https://openreview.net/forum?id=kHO2ZTa8e3}
}

@misc{anthropichh,
      title={Training a Helpful and Harmless Assistant with Reinforcement Learning from Human Feedback}, 
      author={Yuntao Bai and Andy Jones and Kamal Ndousse and Amanda Askell and Anna Chen and Nova DasSarma and Dawn Drain and Stanislav Fort and Deep Ganguli and Tom Henighan and Nicholas Joseph and Saurav Kadavath and Jackson Kernion and Tom Conerly and Sheer El-Showk and Nelson Elhage and Zac Hatfield-Dodds and Danny Hernandez and Tristan Hume and Scott Johnston and Shauna Kravec and Liane Lovitt and Neel Nanda and Catherine Olsson and Dario Amodei and Tom Brown and Jack Clark and Sam McCandlish and Chris Olah and Ben Mann and Jared Kaplan},
      year={2022},
      eprint={2204.05862},
      archivePrefix={arXiv},
      primaryClass={cs.CL},
      url={https://arxiv.org/abs/2204.05862}, 
}

@inproceedings{wu2025meta,
  title={Meta-rewarding language models: Self-improving alignment with llm-as-a-meta-judge},
  author={Wu, Tianhao and Yuan, Weizhe and Golovneva, Olga and Xu, Jing and Tian, Yuandong and Jiao, Jiantao and Weston, Jason E and Sukhbaatar, Sainbayar},
  booktitle={Proceedings of the 2025 Conference on Empirical Methods in Natural Language Processing},
  pages={11548--11565},
  year={2025}
}

@article{bradley1952rank,
  title={Rank analysis of incomplete block designs: I. the method of paired comparisons},
  author={Bradley, Ralph Allan and Terry, Milton E},
  journal={Biometrika},
  volume={39},
  number={3/4},
  pages={324--345},
  year={1952},
  publisher={JSTOR}
}

@article{cai2024asymptotic,
  title={Asymptotic distribution-free independence test for high-dimension data},
  author={Cai, Zhanrui and Lei, Jing and Roeder, Kathryn},
  journal={Journal of the American Statistical Association},
  volume={119},
  number={547},
  pages={1794--1804},
  year={2024},
  publisher={Taylor \& Francis}
}

@article{cai2022model,
  title={Model-free prediction test with application to genomics data},
  author={Cai, Zhanrui and Lei, Jing and Roeder, Kathryn},
  journal={Proceedings of the National Academy of Sciences},
  volume={119},
  number={34},
  pages={e2205518119},
  year={2022},
  publisher={National Academy of Sciences}
}

@inproceedings{liu2023g,
  title={G-Eval: NLG Evaluation using Gpt-4 with Better Human Alignment},
  author={Liu, Yang and Iter, Dan and Xu, Yichong and Wang, Shuohang and Xu, Ruochen and Zhu, Chenguang},
  booktitle={Proceedings of the 2023 Conference on Empirical Methods in Natural Language Processing},
  pages={2511--2522},
  year={2023}
}

@inproceedings{bartlett2017spectrally,
  title={Spectrally-normalized margin bounds for neural networks},
  author={Bartlett, Peter L. and Foster, Dylan J. and Telgarsky, Matus J.},
  booktitle={Advances in Neural Information Processing Systems},
  volume={30},
  year={2017},
  url={https://papers.nips.cc/paper/7204-spectrally-normalized-margin-bounds-for-neural-networks.pdf}
}

@inproceedings{edelman2022inductive,
  title={Inductive Biases and Variable Creation in Self-Attention Mechanisms},
  author={Edelman, Benjamin L. and Goel, Surbhi and Kakade, Sham and Zhang, Cyril},
  booktitle={Proceedings of the 39th International Conference on Machine Learning},
  series={Proceedings of Machine Learning Research},
  volume={162},
  pages={5793--5831},
  year={2022},
  publisher={PMLR},
  url={https://proceedings.mlr.press/v162/edelman22a.html}
}

@article{bartlett2005local,
  title={Local {R}ademacher complexities},
  author={Bartlett, Peter L. and Bousquet, Olivier and Mendelson, Shahar},
  journal={The Annals of Statistics},
  volume={33},
  number={4},
  pages={1497--1537},
  year={2005},
  doi={10.1214/009053605000000282},
  url={https://www.stat.berkeley.edu/~bartlett/papers/bbm-lrc-05.pdf}
}

@inproceedings{cortes2010importance,
  title={Learning Bounds for Importance Weighting},
  author={Cortes, Corinna and Mansour, Yishay and Mohri, Mehryar},
  booktitle={Advances in Neural Information Processing Systems},
  volume={23},
  pages={442--450},
  year={2010},
  url={https://papers.nips.cc/paper_files/paper/2010/hash/59c33016884a62116be975a9bb8257e3-Abstract.html}
}

@article{cassel1976generalized,
  author={Cassel, Claes M. and S{\"a}rndal, Carl E. and Wretman, Jan H.},
  title={Some results on generalized difference estimation and generalized regression estimation for finite populations},
  journal={Biometrika},
  volume={63}, number={3}, pages={615--620}, year={1976},
  doi={10.1093/biomet/63.3.615}
}

@article{robins1994estimation,
  author={Robins, James M. and Rotnitzky, Andrea and Zhao, Lue Ping},
  title={Estimation of Regression Coefficients When Some Regressors Are Not Always Observed},
  journal={Journal of the American Statistical Association},
  volume={89}, number={427}, pages={846--866}, year={1994},
  doi={10.1080/01621459.1994.10476818}
}

@book{sarndal1992model,
  author={S{\"a}rndal, Carl-Erik and Swensson, Bengt and Wretman, Jan},
  title={Model Assisted Survey Sampling},
  publisher={Springer}, year={1992},
  doi={10.1007/978-1-4612-4378-6}
}

@article{mozer2026difference,
  author={Mozer, Reagan},
  title={{PPI} is the Difference Estimator: Recognizing the Survey Sampling Roots of Prediction-Powered Inference},
  journal={arXiv preprint arXiv:2603.19160}, year={2026}
}

@article{datta2025prediction,
  author={Datta, Jyotishka and Polson, Nicholas G.},
  title={Prediction-Powered Inference with Inverse Probability Weighting},
  journal={arXiv preprint arXiv:2508.10149}, year={2025}
}

@article{bartlett2002rademacher,
  title={Rademacher and Gaussian Complexities:
         Risk Bounds and Structural Results},
  author={Bartlett, Peter L. and Mendelson, Shahar},
  journal={Journal of Machine Learning Research},
  volume={3},
  pages={463--482},
  year={2002}
}

@inproceedings{loureiro2022timelms,
  author = {Loureiro, Daniel and Barbieri, Francesco and Neves, Leonardo and Espinosa Anke, Luis and Camacho-Collados, Jose},
  title = {{TimeLMs}: Diachronic Language Models from {Twitter}},
  booktitle = {Proceedings of the 60th Annual Meeting of the Association for Computational Linguistics: System Demonstrations},
  year = {2022},
  pages = {251--260},
  publisher = {Association for Computational Linguistics},
  doi = {10.18653/v1/2022.acl-demo.25},
  url = {https://aclanthology.org/2022.acl-demo.25/}
}

@article{qwen3,
  author = {Yang, An and Li, Anfeng and Yang, Baosong and others},
  title = {{Qwen3} Technical Report},
  journal = {arXiv preprint arXiv:2505.09388},
  year = {2025},
  url = {https://arxiv.org/abs/2505.09388}
}

\clearpage
\appendix

\pdfbookmark[0]{Supplementary Material}{supplement}
\begingroup
\spacingset{1}
\begin{center}
{\Large\bfseries Supplementary Material for\\[0.5em]
A Statistical Framework for Alignment with Biased AI Feedback\par}
\end{center}
\endgroup
\bigskip

\addtocontents{toc}{\protect\setcounter{tocdepth}{3}}
\tableofcontents
\clearpage

\allowdisplaybreaks
\section{Theoretical Proofs}

\subsection{Proofs of the Main Results}
\label{sec:main-proofs}

\paragraph{Sampling, nuisance fitting and norms.}
We first clarify the conventions used throughout the proofs. Let $\mathcal T$ denote the independent training data used to fit $\widehat\pi_{\mathrm{Gen}}^{\mathrm{Hum}}$, the importance weights, and, when estimated, the evaluator $\tilde g$. Any pretrained representation used below is also fixed or trained on $\mathcal T$. The human comparison tuples and human labels $(U_i,Z_i)$ are independent of $\mathcal T$. The paired AI labels are generated using the fitted evaluator; the augmented human observations are i.i.d. conditional on $\mathcal T$. For DB-DPO and fixed-policy DIPO evaluation, the two evaluation samples are also independent conditional on $\mathcal T$. In Theorem~\ref{thm:regret_dipo}, each new DIPO batch instead follows the stipulated sampling law conditional on $\mathcal T$, the human sample, and the preceding optimization history. Its target-response law depends on the current policy. Fresh simulation uses random variables independent of this history. Unless stated otherwise, expectations and empirical-process bounds condition on $\mathcal T$, so fitted nuisances are treated as fixed. Moment bounds hold almost surely over $\mathcal T$, with fixed constants, and all suprema are assumed measurable. Mean-square error and $o_{L^2}$ conclusions also average over training randomness. Both sources have prompt law $\mathbb P_X$. Under the human comparison law, $Y^{(1)},Y^{(2)}\mid X$ are independent draws from $\pi_{\mathrm{Gen}}^{\mathrm{Hum}}(\cdot\mid X)$. Write $U=(X,Y^{(1)},Y^{(2)})$ and let $\mathbb E_{\mathrm{Hum}}$ denote expectation under this law, including the paired labels when present. Thus
$$
\|h\|_{q,\mathrm{Hum}}
=\{\mathbb E_{\mathrm{Hum}}|h|^q\}^{1/q},\qquad 1\le q<\infty.
$$
The subscript-free $\|h\|_q$ and the weight moments in the main text use this human law. The unindexed $\|\tilde g-g\|$ instead uses the target/reference comparison law defined in Theorem~\ref{thm:mse}. For DB-DPO, the AI response pair consists of independent draws from $\pi_{\mathrm{Gen}}^{\mathrm{AI}}(\cdot\mid X)$. For DIPO's direct term, conditional on $X$, the responses $Y\sim\pi(\cdot\mid X)$ and $Y^{(1)},Y^{(2)}\sim\pi_{\mathrm{ref}}(\cdot\mid X)$ are independent. The notation $\mathbb E_{\mathrm{AI}}$ and $\|\cdot\|_{q,\mathrm{AI}}$ refers to the corresponding AI or direct sampling law for the method under consideration.

The paired labels have conditional means $\mathbb E(Z\mid U,\mathcal T)=g(Y^{(1)},Y^{(2)}\mid X)$ and $\mathbb E(\hat Z\mid U,\mathcal T)=\tilde g(Y^{(1)},Y^{(2)}\mid X)$; they need not be conditionally independent. For the symmetrized DIPO loss, we use the comparison convention $h(y_2,y_1\mid x)=1-h(y_1,y_2\mid x)$ for $h=g,\tilde g$. Assumption~\ref{assump:aif_hf} is understood uniformly over comparisons:
$$
\operatorname*{ess\,sup}_{X,Y^{(1)},Y^{(2)}}\mathbb P(Z\ne\hat Z\mid X,Y^{(1)},Y^{(2)},\mathcal T)=O_p(\nu_n),\qquad 0<\nu_n\le1.
$$
Here the stochastic order is over $\mathcal T$; the essential supremum covers both comparison laws where the assumption is used.

We first establish the theoretical results for DB-DPO.

\begin{proof}[Proof of Theorem \ref{thm:subopt-gap}]
	Write $\hat\pi=\hat\pi_{\mathrm{DB\text{-}DPO}}$ and condition on the independent nuisance-training sample. For $\pi\in\Pi$, use the induced reward
	\begin{equation}\label{eq:rpi}
		r^\pi(Y,X)=\beta\log\frac{\pi(Y\mid X)}{\pi_{\mathrm{ref}}(Y\mid X)} +\beta\log W(X),
	\end{equation}
	and write $r^*(Y,X)=r(X,Y)=r^{\pi^*}(Y,X)$, as in Assumption~\ref{assump:realizability}. Let $l(\pi)$ be the population log-likelihood under the AI generation distribution and human labels, and let $\mathbb E_{n,N}l(\pi)=-\hat L_{\mathrm{DB\text{-}DPO}}(\pi;\pi_{\mathrm{ref}})$. The definitions of $f_\pi$, $\sigma_\pi$, $a_i(\pi)$ and $b_i(\pi)$ in Appendix~\ref{sec:rad-complexity} give
	$$
	\mathbb E_{n,N}l(\pi)-\mathbb E_{n,N}l(\pi^*)=\frac1N\sum_{i=n+1}^{n+N}a_i(\pi)+\frac1n\sum_{i=1}^n b_i(\pi).
	$$
	Under the BT model, the conditional score at the true reward difference has mean zero. Since all candidate rewards are uniformly bounded, the negative log-likelihood has second derivative bounded below by a positive constant on all candidate reward differences. Taylor's theorem therefore gives
	$$
	l(\pi^*)-l(\pi)\ge c\sigma_\pi^2,\qquad \pi\in\Pi,
	$$
	for a constant $c>0$. We center the fitted empirical loss about its conditional mean. Ordinary symmetrization and \eqref{eq:rad-ddpo-ai}--\eqref{eq:rad-ddpo-human} give
	\begin{align*}
		&\mathbb E\sup_{\pi\in\Pi}\Big| \mathbb E_{n,N}l(\pi)-\mathbb E_{n,N}l(\pi^*)-\mathbb E\{\mathbb E_{n,N}l(\pi)-\mathbb E_{n,N}l(\pi^*)\}\Big|\\
		\le& 2\mathbb E\sup_{\pi\in\Pi}\bigg|\frac1N\sum_{i=n+1}^{n+N}\varepsilon_i a_i(\pi)\bigg|+2\mathbb E\sup_{\pi\in\Pi}\bigg|\frac1n\sum_{i=1}^n\varepsilon_i b_i(\pi)\bigg|\le C\mu_n,
	\end{align*}
	where
	$$ \mu_n=\sqrt{\frac vN}+\sqrt{\frac{v\nu_n}{n}}.$$
	Markov's inequality yields a uniform $O_p(\mu_n)$ bound, which also holds at the data-dependent $\hat\pi$. The remaining population bias comes only from estimating $w$. The identity $\log\sigma(t)-\log\sigma(-t)=t$ and change of measure give
	\begin{align*}
		&\mathbb E\{\mathbb E_{n,N}l(\pi)-\mathbb E_{n,N}l(\pi^*)\}
		-\{l(\pi)-l(\pi^*)\}\\
		=&\mathbb E_{X\sim\mathbb P_X,\,(Y^{(1)},Y^{(2)})\sim\pi_{\mathrm{Gen}}^{\mathrm{Hum}}(\cdot\mid X)}\big[(\hat w-w)(Z-\hat Z)f_\pi(X,Y^{(1)},Y^{(2)})\big].
	\end{align*}
	Since each human generation probability is at most one, the retained coverage conditions imply $w\ge\epsilon^4$ on the human support. Uniform conditional disagreement and Cauchy--Schwarz bound its absolute value, simultaneously for every $\pi$, by
	$$
	O_p(\nu_n)\|\hat w-w\|_2\{\mathbb E_{X\sim\mathbb P_X,\,(Y^{(1)},Y^{(2)})\sim\pi_{\mathrm{Gen}}^{\mathrm{Hum}}(\cdot\mid X)}f_\pi^2\}^{1/2}\le O_p(\nu_n\epsilon^{-2})\|\hat w-w\|_2\sigma_\pi.
	$$
	Here $\|\hat w-w\|_2$ is exactly the norm in the theorem statement. Empirical optimality and the curvature inequality imply, with $\sigma=\sigma_{\hat\pi}$,
	$$
	c\sigma^2\le O_p(\mu_n)+O_p(\nu_n\epsilon^{-2})\|\hat w-w\|_2\sigma.
	$$
	Solving this quadratic inequality gives
	$$
	\sigma=O_p\bigg(\mu_n^{1/2}+\nu_n\epsilon^{-2}\|\hat w-w\|_2\bigg)
	=O_p\bigg\{\bigg(\frac vN\bigg)^{1/4}+\bigg(\frac{v\nu_n}{n}\bigg)^{1/4}+\nu_n^{1/2}\|\hat w-w\|_2\bigg\},
	$$
	where the last, weaker bound uses $0<\nu_n\le1$ and fixed $\epsilon$. We now convert the reward-difference error to the stated value gap. Put $d_X(Y)=r^{\hat\pi}(Y,X)-r^*(Y,X)$ and center it under $\pi_{\mathrm{Gen}}^{\mathrm{AI}}(\cdot\mid X)$:
	$$
	\bar d_X(Y)=d_X(Y)
	-\mathbb E_{Y'\sim\pi_{\mathrm{Gen}}^{\mathrm{AI}}(\cdot\mid X)}d_X(Y'),
	\qquad
	\sigma_X^2=\mathbb E[f_{\hat\pi}^2\mid X].
	$$
	Conditional independence of the two responses gives
	$\sigma_X^2=2\mathbb E_{Y\sim\pi_{\mathrm{Gen}}^{\mathrm{AI}}}\bar d_X(Y)^2$. The coverage condition gives $\pi_{\mathrm{Gen}}^{\mathrm{AI}}(Y\mid X)\ge\epsilon^2$, so
	$$
	\max_Y|\bar d_X(Y)|\le\frac{\sigma_X}{\sqrt2\epsilon}.
	$$
	For $0\le t\le1$, let $\pi_t(\cdot\mid X)$ be the policy with reference $\pi_{\mathrm{ref}}$ and reward $(1-t)r^*(\cdot,X)+t r^{\hat\pi}(\cdot,X)$. Its endpoints are $\pi^*$ and $\hat\pi$. Differentiation gives
	$$
	\bigg|\frac{d}{dt}\mathbb E_{Y\sim\pi_t(\cdot\mid X)}r^*(Y,X)\bigg|=\frac1\beta|\operatorname{Cov}_{Y\sim\pi_t(\cdot\mid X)}(r^*(Y,X),\bar d_X(Y))|\le\frac C\beta\max_Y|\bar d_X(Y)|.
	$$
	Integrating over $t$ and $X$, and using $\mathbb E_X\sigma_X\le(\mathbb E_X\sigma_X^2)^{1/2}=\sigma$, yields
	$$
	|J(\pi^*)-J(\hat\pi)|\le\frac C{\beta\epsilon}\sigma,
	\qquad
	J(\pi)=\mathbb E_{X\sim\mathbb P_X,\,Y\sim\pi(\cdot\mid X)}r^*(Y,X).
	$$
	Combining the last two bounds proves \eqref{eq:subopt-dpo}.
	\hfill $\blacksquare$
\end{proof}

Before presenting the main theoretical results for DIPO, we first establish the influence function expansion.

\begin{proof}[Proof of Lemma \ref{lem:semi}]
	For notational simplicity, we denote
	\begin{equation*}
		\begin{split}
			&\psi(W)\\
			=&\psi(X, Y^{(1)}, Y^{(2)}, Z;\pi,\pi_{\text{ref}},\pi_{\text{Gen}}^{\text{Hum}}, g)\\
			=&\mathbb{E}_{Y\sim\pi(\cdot\mid X),Y^{\prime}\sim\pi_{\text{ref}}(\cdot\mid X)}g(Y,Y^{\prime}\mid X)\\
			&+\frac{1}{2} \frac{\pi(Y^{(1)} \mid X)\pi_{\text{ref}}(Y^{(2)} \mid X)-\pi(Y^{(2)} \mid X)\pi_{\text{ref}}(Y^{(1)} \mid X)}{\pi_{\text{Gen}}^{\text{Hum}}(Y^{(1)}\mid X)\pi_{\text{Gen}}^{\text{Hum}}(Y^{(2)} \mid X)}(Z - g(Y^{(1)}, Y^{(2)}\mid  X)).
		\end{split}
	\end{equation*}
	where $W=(X, Y^{(1)}, Y^{(2)}, Z)$ denotes the i.i.d.\ copies of the following generation model. The $X$ is generated from the distribution $\mathbb P_X$; the $Y^{(1)}, Y^{(2)}\mid X$ are generated from the distribution $\pi_{\text{Gen}}^{\text{Hum}}(\cdot\mid X)$; the $Z$ is generated from the distribution $Z=1$ with probability $\mathbb{P}_{\text{Human}}(Y^{(1)}\succ Y^{(2)}\mid X)=g(Y^{(1)},Y^{(2)}\mid X)$. We use $M$ to denote the model. Thus, the model contains three types of parameters:
	
	\begin{enumerate}
		\item[(i)] parameters $\gamma$ governing the density $f_{X}$ of $X$;
		\item[(ii)] parameters $b$ governing the paired-data generation $\pi_{\text{Gen}}^{\text{Hum}}(\cdot\mid\cdot)$ for 
		$(Y^{(1)}, Y^{(2)})$;
		\item[(iii)] parameters $\eta$ governing the preference probability $\mathbb{P}_{\text{Human}}(Y^{(1)}\succ Y^{(2)}\mid X)$ denoted by $g(Y^{(1)},Y^{(2)}\mid X)$.
	\end{enumerate}
	
	Thus, the likelihood function is
	\begin{equation}
		\label{eq:likelihood}
		\ell(W;\gamma,b,\eta)=f_{\gamma}(X)\pi_b(Y^{(1)}\mid X)\pi_b(Y^{(2)}\mid X)g_{\eta}(Y^{(1)},Y^{(2)},X)^{Z}(1-g_{\eta}(Y^{(1)},Y^{(2)},X))^{1-Z},
	\end{equation}
	where we use $f_{\gamma}(X)$, $\pi_b(Y\mid X)$, $g_{\eta}(Y^{(1)},Y^{(2)}\mid X)$ to denote submodel components for $f_{X}$, $\pi_{\text{Gen}}^{\text{Hum}}$, $g$ indexed by the parameters $\gamma$, $b$, $\eta$. Let $(\gamma_0,b_0,\eta_0)$ denote the true parameters, so that $f_{\gamma_0}=f_{X}$, $\pi_{b_0}=\pi_{\text{Gen}}^{\text{Hum}}$, and $g_{\eta_0}=g$. Along these submodels, define the target functional by
	\begin{equation*}
		\begin{split}
			&p(\pi;\gamma,b,\eta)=\int g_\eta(Y,Y^{\prime}\mid X) \pi(Y\mid X) \pi_{\text{ref}}(Y^{\prime}\mid X)f_\gamma(X)dXdYdY^{\prime},
		\end{split}
	\end{equation*}
	Let $p^{*}(\pi)=p(\pi;\gamma_0,b_0,\eta_0)=P(\pi)$ denote the target at the true parameter values. The target does not depend on $b$ because the target and reference policies are fixed. Unless stated otherwise, expectations are under the true law, and derivatives are evaluated at $(\gamma_0,b_0,\eta_0)$ after differentiation. We verify the pathwise derivative identity and tangent-space membership. 
	
	\paragraph{Step 1:} The term $E\left[\{\psi(W)-p^{*}(\pi)\}\nabla\log \ell(W; \gamma,b,\eta)\right]$ is the first-order derivative of $p(\pi;\gamma,b,\eta)$ with respect to $\gamma,b,\eta$ evaluated at $(\gamma_0,b_0,\eta_0)$. Note that for the second part, 
	\begin{equation*}
		E\left[p^{*}(\pi)\nabla\log \ell(W; \gamma,b,\eta)\right]=p^{*}(\pi)\nabla E\left[1\right]=0. 
	\end{equation*}
	It remains to show that
	\begin{align*}
		\text{(i)}\quad
		\mathbb{E}\bigg[
		\psi(W)\,
		\frac{\partial}{\partial \gamma} \log \ell(W; \gamma, b, \eta)\bigg]
		&=
		\left.\frac{\partial}{\partial \gamma} p(\pi;\gamma,b,\eta)\right|_{\gamma = \gamma_0},
		\\[0.5em]
		\text{(ii)}\quad
		\mathbb{E}\bigg[
		\psi(W)\,
		\frac{\partial}{\partial b} \log \ell(W; \gamma, b, \eta)
		\bigg]
		&=
		\left.\frac{\partial}{\partial b} p(\pi;\gamma,b,\eta)\right|_{b = b_0},
		\\[0.5em]
		\text{(iii)}\quad
		\mathbb{E}\bigg[
		\psi(W)\,
		\frac{\partial}{\partial \eta} \log \ell(W; \gamma, b, \eta)
		\bigg]
		&=
		\left.\frac{\partial}{\partial \eta} p(\pi;\gamma,b,\eta)\right|_{\eta = \eta_0}.
	\end{align*}
	Let $w = (x, y_1, y_2, z)$ denote a realization of $W = (X, Y^{(1)}, Y^{(2)}, Z)$. By \eqref{eq:likelihood}, we have
	\begin{equation*}
		\begin{split}
			\log \ell(w; \gamma, b, \eta)=&
			\log f_{\gamma}(x)+ \log \pi_{b}(y_1 \mid x)+ \log \pi_{b}(y_2 \mid x)\\
			&+ z \log g_{\eta}(y_1, y_2, x) + (1 - z)\log\bigl\{1 - g_{\eta}(y_1, y_2, x)\bigr\}.
		\end{split}
	\end{equation*}
	A direct calculation yields
	\begin{align*}
		\left.\frac{\partial}{\partial \gamma} \log \ell(w; \gamma, b, \eta)\right|_{\gamma = \gamma_0}&=\frac{1}{f_X(x)}\,\left.\frac{\partial}{\partial \gamma} f_{\gamma}(x)\right|_{\gamma = \gamma_0}, \\
		\left.\frac{\partial}{\partial b} \log \ell(w; \gamma, b, \eta)\right|_{b = b_0}&=\frac{1}{\pi_{\text{Gen}}^{\text{Hum}}(y_1 \mid x)}\left.\frac{\partial}{\partial b} \pi_b(y_1 \mid x)\right|_{b=b_0}+\frac{1}{\pi_{\text{Gen}}^{\text{Hum}}(y_2 \mid x)}
		\left.\frac{\partial}{\partial b} \pi_b(y_2 \mid x)\right|_{b=b_0}, \\
		\left.\frac{\partial}{\partial \eta} \log \ell(w; \gamma, b, \eta)\right|_{\eta = \eta_0}
		&=\left\{\frac{z}{g(y_1, y_2\mid x)}-\frac{1 - z}{1 - g(y_1, y_2\mid x)}\right\}\left.\frac{\partial}{\partial \eta} g_{\eta}(y_1, y_2, x)\right|_{\eta=\eta_0}.
	\end{align*}
	In what follows, we omit the evaluations at $(\gamma_0, b_0, \eta_0)$ from the notation for brevity.
	
	\textit{Term (i)}:
	\begin{align*}
		&\mathbb{E}\left[\psi(W)\frac{\partial}{\partial \gamma} \log \ell(W; \gamma, b, \eta)\right]\\
		&=\frac{1}{2}\sum_{x, y_1, y_2}\mathbb{E}_{z \sim g(y_1, y_2\mid  x)}\biggl[\Bigl\{\frac{\pi(y_1 \mid x)\pi_{\text{ref}}(y_2 \mid x)}{\pi_{\text{Gen}}^{\text{Hum}}(y_1 \mid x)\pi_{\text{Gen}}^{\text{Hum}}(y_2 \mid x)}-\frac{\pi(y_2 \mid x)\pi_{\text{ref}}(y_1 \mid x)}{\pi_{\text{Gen}}^{\text{Hum}}(y_2 \mid x)\pi_{\text{Gen}}^{\text{Hum}}(y_1 \mid x)}\Bigr\}\\
		&\times(z - g(y_1, y_2\mid x))\pi_{\text{Gen}}^{\text{Hum}}(y_1 \mid x)\pi_{\text{Gen}}^{\text{Hum}}(y_2 \mid x)\frac{\partial}{\partial \gamma} f_{\gamma}(x)\biggr] \\
		&+\sum_{x,y_1,y_2}\mathbb{E}_{Y\sim\pi(\cdot\mid x),Y^{\prime}\sim\pi_{\text{ref}}(\cdot\mid x)}g(Y,Y^{\prime}\mid x)\pi_{\text{Gen}}^{\text{Hum}}(y_1 \mid x)\pi_{\text{Gen}}^{\text{Hum}}(y_2 \mid x)\frac{\partial}{\partial \gamma} f_{\gamma}(x).
	\end{align*}
	Note that $\mathbb{E}_{z \sim g(y_1, y_2\mid x)}\bigl\{z - g(y_1, y_2\mid x)\bigr\} = 0$, we have that the first term vanishes. Hence
	\begin{align*}
		&\mathbb{E}\left[\psi(W)\frac{\partial}{\partial \gamma} \log \ell(W; \gamma, b, \eta)\right]\\
		=&\sum_{x,y_1,y_2}\mathbb{E}_{Y\sim\pi(\cdot\mid x),Y^{\prime}\sim\pi_{\text{ref}}(\cdot\mid x)}g(Y,Y^{\prime}\mid x)\pi_{\text{Gen}}^{\text{Hum}}(y_1 \mid x)\pi_{\text{Gen}}^{\text{Hum}}(y_2 \mid x)\frac{\partial}{\partial \gamma} f_{\gamma}(x)\\
		=&\sum_{x}\mathbb{E}_{Y\sim\pi(\cdot\mid x),Y^{\prime}\sim\pi_{\text{ref}}(\cdot\mid x)}g(Y,Y^{\prime}\mid x)\frac{\partial}{\partial \gamma} f_{\gamma}(x)\\
		=&\int g(Y,Y^{\prime}\mid X) \pi(Y\mid X) \pi_{\text{ref}}(Y^{\prime}\mid X)\frac{\partial}{\partial \gamma} f_{\gamma}(X)dXdYdY^{\prime}=\frac{\partial}{\partial \gamma} p(\pi;\gamma,b,\eta).
	\end{align*}
	Thus, Term (i) is established.
	
	\textit{Term (ii)}:
	\begin{align*}
		&\mathbb{E}\!\left[\psi(W)\frac{\partial}{\partial b} \log \ell(W; \gamma, b, \eta)\right]\\
		&=\frac{1}{2}\sum_{x, y_1, y_2}\mathbb{E}_{z \sim g(y_1, y_2\mid x)}\biggl[\Bigl\{\frac{\pi(y_1 \mid x)\pi_{\text{ref}}(y_2 \mid x)}{\pi_{\text{Gen}}^{\text{Hum}}(y_1 \mid x)\pi_{\text{Gen}}^{\text{Hum}}(y_2 \mid x)}-\frac{\pi(y_2 \mid x)\pi_{\text{ref}}(y_1 \mid x)}{\pi_{\text{Gen}}^{\text{Hum}}(y_2 \mid x)\pi_{\text{Gen}}^{\text{Hum}}(y_1 \mid x)}\Bigr\}\\
		&\times(z - g(y_1, y_2\mid x))\pi_{\text{Gen}}^{\text{Hum}}(y_1 \mid x)\pi_{\text{Gen}}^{\text{Hum}}(y_2 \mid x)f_X(x)\bigg\{\frac{1}{\pi_{\text{Gen}}^{\text{Hum}}(y_1 \mid x)}\frac{\partial}{\partial b} \pi_b(y_1 \mid x)\\
		&+\frac{1}{\pi_{\text{Gen}}^{\text{Hum}}(y_2 \mid x)}\frac{\partial}{\partial b} \pi_b(y_2 \mid x)\bigg\}\biggr] \\
		&+\sum_{x, y_1, y_2}\mathbb{E}_{Y\sim\pi(\cdot\mid x),Y^{\prime}\sim\pi_{\text{ref}}(\cdot\mid x)}g(Y,Y^{\prime}\mid x)\\
		&\times\pi_{\text{Gen}}^{\text{Hum}}(y_1 \mid x)\pi_{\text{Gen}}^{\text{Hum}}(y_2 \mid x)f_X(x)\bigg\{\frac{1}{\pi_{\text{Gen}}^{\text{Hum}}(y_1 \mid x)}\frac{\partial}{\partial b} \pi_b(y_1 \mid x)+\frac{1}{\pi_{\text{Gen}}^{\text{Hum}}(y_2 \mid x)}\frac{\partial}{\partial b} \pi_b(y_2 \mid x)\bigg\}.
	\end{align*}
	Note that $\mathbb{E}_{Z \sim g(y_1, y_2\mid x)}\bigl\{Z - g(y_1, y_2\mid x)\bigr\} = 0$, we have that the first term vanishes. Thus, it remains to consider the second term,
	\begin{align*}
		&\sum_{x, y_1, y_2}\mathbb{E}_{Y\sim\pi(\cdot\mid x),Y^{\prime}\sim\pi_{\text{ref}}(\cdot\mid x)}g(Y,Y^{\prime}\mid x)\\
		&\times\pi_{\text{Gen}}^{\text{Hum}}(y_1 \mid x)\pi_{\text{Gen}}^{\text{Hum}}(y_2 \mid x)f_X(x)\bigg\{\frac{1}{\pi_{\text{Gen}}^{\text{Hum}}(y_1 \mid x)}\frac{\partial}{\partial b} \pi_b(y_1 \mid x)+\frac{1}{\pi_{\text{Gen}}^{\text{Hum}}(y_2 \mid x)}\frac{\partial}{\partial b} \pi_b(y_2 \mid x)\bigg\}\\
		=&\sum_{x, y_1, y_2}\mathbb{E}_{Y\sim\pi(\cdot\mid x),Y^{\prime}\sim\pi_{\text{ref}}(\cdot\mid x)}g(Y,Y^{\prime}\mid x)f_X(x)\frac{\partial}{\partial b}\{\pi_b(y_1 \mid x)\pi_b(y_2 \mid x)\}|_{b=b_0}\\
		=&\frac{\partial}{\partial b}\left.\bigg\{\sum_{x, y_1, y_2}\mathbb{E}_{Y\sim\pi(\cdot\mid x),Y^{\prime}\sim\pi_{\text{ref}}(\cdot\mid x)}g(Y,Y^{\prime}\mid x)f_X(x)\pi_b(y_1 \mid x)\pi_b(y_2 \mid x)\bigg\}\right|_{b=b_0}\\
		=&\frac{\partial}{\partial b}\left.\bigg\{\sum_{x}\mathbb{E}_{Y\sim\pi(\cdot\mid x),Y^{\prime}\sim\pi_{\text{ref}}(\cdot\mid x)}g(Y,Y^{\prime}\mid x)f_X(x)\bigg\}\right|_{b=b_0}=
		\left.\frac{\partial}{\partial b} p(\pi;\gamma,b,\eta)\right|_{b = b_0}.
	\end{align*}
	Thus, Term (ii) is established.
	
	\textit{Term (iii)}:
	\begin{align*}
		&\mathbb{E}\!\left[\psi(W)\frac{\partial}{\partial \eta} \log \ell(W; \gamma, b, \eta)\right]\\
		&=\frac{1}{2}\sum_{x, y_1, y_2}\mathbb{E}_{z \sim g(y_1, y_2\mid x)}\biggl[\Bigl\{\frac{\pi(y_1 \mid x)\pi_{\text{ref}}(y_2 \mid x)}{\pi_{\text{Gen}}^{\text{Hum}}(y_1 \mid x)\pi_{\text{Gen}}^{\text{Hum}}(y_2 \mid x)}-\frac{\pi(y_2 \mid x)\pi_{\text{ref}}(y_1 \mid x)}{\pi_{\text{Gen}}^{\text{Hum}}(y_2 \mid x)\pi_{\text{Gen}}^{\text{Hum}}(y_1 \mid x)}\Bigr\}\\
		&\times(z - g(y_1, y_2\mid x))\pi_{\text{Gen}}^{\text{Hum}}(y_1 \mid x)\pi_{\text{Gen}}^{\text{Hum}}(y_2 \mid x)f_X(x)\left\{\frac{z}{g(y_1, y_2\mid x)}-\frac{1 - z}{1 - g(y_1, y_2\mid  x)}\right\}\\
		&\times\frac{\partial}{\partial \eta} g_{\eta}(y_1, y_2, x)\biggr] \\
		&+\mathbb{E}_{z \sim g(y_1, y_2\mid x)}\sum_{x, y_1, y_2}\mathbb{E}_{Y\sim\pi(\cdot\mid x),Y^{\prime}\sim\pi_{\text{ref}}(\cdot\mid x)}g(Y,Y^{\prime}\mid x)\\
		&\times\pi_{\text{Gen}}^{\text{Hum}}(y_1 \mid x)\pi_{\text{Gen}}^{\text{Hum}}(y_2 \mid x)f_X(x)\left\{\frac{z}{g(y_1, y_2\mid x)}-\frac{1 - z}{1 - g(y_1, y_2\mid x)}\right\}\frac{\partial}{\partial \eta} g_{\eta}(y_1, y_2\mid x).
	\end{align*}
	The second term is zero because
	$$
	\mathbb{E}_{z \sim g(y_1, y_2\mid x)}\Bigl[\frac{z}{g(y_1, y_2\mid x)}-\frac{1 - z}{1 - g(y_1, y_2\mid x)}\Bigr] = 0.
	$$
	Moreover, using the fact
	\begin{align*}
		\mathbb{E}_{z \sim g(y_1, y_2\mid x)}\Bigl[(z - g(y_1, y_2\mid x))\Bigl\{\frac{z}{g(y_1, y_2\mid x)}-\frac{1 - z}{1 - g(y_1, y_2\mid x)}\Bigr\}\Bigr]= 1,
	\end{align*}
	we obtain
	\begin{align*}
		&\mathbb{E}\!\left[\psi(W)\frac{\partial}{\partial \eta} \log \ell(W; \gamma, b, \eta)\right]\\
		=&\frac{1}{2}\sum_{x, y_1, y_2}\biggl[\Bigl\{\frac{\pi(y_1 \mid x)\pi_{\text{ref}}(y_2 \mid x)}{\pi_{\text{Gen}}^{\text{Hum}}(y_1 \mid x)\pi_{\text{Gen}}^{\text{Hum}}(y_2 \mid x)}-\frac{\pi(y_2 \mid x)\pi_{\text{ref}}(y_1 \mid x)}{\pi_{\text{Gen}}^{\text{Hum}}(y_2 \mid x)\pi_{\text{Gen}}^{\text{Hum}}(y_1 \mid x)}\Bigr\}\\
		&\times\pi_{\text{Gen}}^{\text{Hum}}(y_1 \mid x)\pi_{\text{Gen}}^{\text{Hum}}(y_2 \mid x)f_X(x)\frac{\partial}{\partial \eta} g_{\eta}(y_1, y_2, x)\biggr]\\
		=&\frac{1}{2}\sum_{x, y_1, y_2}\Bigl\{\pi(y_1 \mid x)\pi_{\text{ref}}(y_2 \mid x)-\pi(y_2 \mid x)\pi_{\text{ref}}(y_1 \mid x)\Bigr\}f_X(x)\frac{\partial}{\partial \eta} g_{\eta}(y_1, y_2, x)\\
		=&\sum_{x, y_1, y_2}\pi(y_1 \mid x)\pi_{\text{ref}}(y_2 \mid x)f_X(x)\frac{\partial}{\partial \eta} g_{\eta}(y_1, y_2, x)=\frac{\partial}{\partial \eta} p(\pi;\gamma,b,\eta),
	\end{align*}
	where the second-to-last equality uses the antisymmetry
	\[
	\frac{\partial}{\partial \eta} g_{\eta}(y_1, y_2, x)
	=
	-\,\frac{\partial}{\partial \eta} g_{\eta}(y_2, y_1, x).
	\]
	Thus, Term (iii) is established.
	
	\textbf{Step 2:} $\psi(W)-p^{*}(\pi)\in \mathcal{T}_{M}$, where $\mathcal{T}_{M}$ is the tangent space of model $M$.
	
	The tangent space $\mathcal T_M$ is the $L^2$ closure of scores of regular one-dimensional submodels through the true distribution. We first use bounded perturbations and then take their $L^2$ closures. The likelihood \eqref{eq:likelihood} gives the orthogonal sum
	$$
	\mathcal{T}_{M} = \mathcal{T}_{f} \oplus \mathcal{T}_{\pi} \oplus \mathcal{T}_{g},
	$$
	where $\mathcal{T}_{f}$, $\mathcal{T}_{\pi}$, and $\mathcal{T}_{g}$ denote the sets of score functions for all one–dimensional submodels passing through the marginal distribution $f_X(x)$, the generation policy $\pi_{\text{Gen}}^{\text{Hum}}$, and the preference probability $g$, respectively.
	
	For $\mathcal{T}_{f}$, let $\{ f_{\varepsilon}(x) \}$ be a one–dimensional submodel defined by  
	$$
	f_{\varepsilon}(x) = f_X(x)\,(1 + \varepsilon q(x)),
	$$
	where $q(x)$ satisfies $\sum_x f_X(x)\,q^2(x) < \infty$. Since $f_{\varepsilon}$ must be a valid probability mass function, we require
	$$
	\sum_x f_{\varepsilon}(x) = 1
	\quad \Longrightarrow \quad 
	\mathbb{E}[q(X)] = 0.
	$$
	The score function with respect to $\varepsilon$ is therefore
	$$
	\left.\frac{d}{d\varepsilon} \log f_{\varepsilon}(x)\right|_{\varepsilon=0} = q(x).
	$$
	Thus, the tangent space for $f_X(x)$ is
	$$
	\mathcal{T}_{f}=\{q(x):\ \mathbb{E}[q(X)] = 0,\ \sum_x f_X(x)\,q^2(x) < \infty\}.
	$$
	
	For the one–dimensional submodel of the conditional distribution:
	$$
	\pi_{\varepsilon}(y \mid x)=\pi_{\text{Gen}}^{\text{Hum}}(y \mid x)(1 + \varepsilon q(y,x)),
	$$
	where $q(y,x)$ satisfies $\sum_{y} q^2(y,x)\pi_{\text{Gen}}^{\text{Hum}}(y \mid x)<\infty$. Because $\pi_{\varepsilon}(\cdot \mid x)$ must be a valid conditional probability mass function,
	$$
	\sum_y \pi_{\varepsilon}(y \mid x) = 1\quad \Longrightarrow \quad \mathbb{E}_{Y \sim \pi_{\text{Gen}}^{\text{Hum}}(\cdot \mid x)}[q(Y,x)] = 0\quad \text{for all } x.
	$$
	The score function with respect to $\varepsilon$ is then
	$$
	\left.\frac{d}{d\varepsilon}\log\bigl\{ \pi_{\varepsilon}(y_1 \mid x)\,\pi_{\varepsilon}(y_2 \mid x) \bigr\}\right|_{\varepsilon = 0}= q(y_1,x) + q(y_2,x).
	$$
	Thus, the tangent space for the human generation policy is
	$$
	\mathcal T_\pi=\overline{\{q(Y^{(1)},X)+q(Y^{(2)},X):\quad\mathbb E[q(Y,X)\mid X]=0,\quad \mathbb E q(Y,X)^2<\infty\}},
	$$
	where $Y\mid X\sim\pi_{\mathrm{Gen}}^{\mathrm{Hum}}(\cdot\mid X)$. Here the expectations are under the human generation law, and the bar denotes closure in $L^2$. The two score contributions share the same function $q$ because both responses are drawn independently, conditional on $X$, from the same generation policy. The target and reference policies remain fixed.
	
	For $\mathcal{T}_{g}$, we derive the tangent space associated with the preference probability $g(y_1,y_2\mid x)$. Conditional on $(x,y_1,y_2)$, the distribution of $Z$ is
	$$
	\mathbb{P}(Z=z \mid x,y_1,y_2)=\bigl(g(y_1,y_2\mid x)\bigr)^z\bigl(1-g(y_1,y_2\mid x)\bigr)^{1-z},\qquad z \in \{0,1\}.
	$$
	Define a one–dimensional parametric submodel of $g$ by
	$$
	\mathrm{logit}\left(g_{\varepsilon}(y_1,y_2\mid x)\right)=\mathrm{logit}\!\left(g(y_1,y_2\mid x)\right)+\varepsilon\, q(y_1,y_2,x),
	$$
	where $q(y_2,y_1,x)=-q(y_1,y_2,x)$ preserves preference
	symmetry and
	$$
	\mathbb E_{\mathrm{Hum}}[g(U)\{1-g(U)\}q(U)^2]<\infty.
	$$
	For each $\varepsilon$, $g_{\varepsilon}(y_1,y_2\mid x)$ is a valid probability and $g_{0}=g$. The log-likelihood contribution of $Z$ is
	$$
	\ell_{\varepsilon}(z\mid x,y_1,y_2)=z \log g_{\varepsilon}(y_1,y_2\mid x)+(1-z)\log\!\bigl(1-g_{\varepsilon}(y_1,y_2\mid x)\bigr).
	$$
	Differentiating with respect to $\varepsilon$ gives
	\begin{equation*}
		\begin{split}
			\frac{d}{d\varepsilon}\,\ell_{\varepsilon}(z\mid x,y_1,y_2)&=\frac{z - g_{\varepsilon}(y_1,y_2\mid x)}{g_{\varepsilon}(1-g_{\varepsilon})}\cdot\frac{d}{d\varepsilon}g_{\varepsilon}(y_1,y_2\mid x)\\
			&=\{z - g_{\varepsilon}(y_1,y_2\mid x)\}q(y_1,y_2,x).
		\end{split}
	\end{equation*}
	Substituting and evaluating at $\varepsilon=0$ yields the score
	$$
	\left.\frac{d}{d\varepsilon}\ell_{\varepsilon}(z\mid x,y_1,y_2)\right|_{\varepsilon=0}=\bigl(z - g(y_1,y_2\mid x)\bigr)\, q(y_1,y_2,x).
	$$
	Taking the $L^2$ closure yields
	$$
	\mathcal T_g=\operatorname{cl}_{L^2}\big\{
	(Z-g(U))q(U):\ q(y_2,y_1,x)=-q(y_1,y_2,x),\mathbb E_{\mathrm{Hum}}[g(U)\{1-g(U)\}q(U)^2]<\infty\big\}.
	$$
	Bounded antisymmetric truncations approximate any such score in $L^2$.
	
	To verify that $\psi(W) - p^*(\pi)$ lies in $\mathcal{T}_{M}$, first note that
	\begin{equation*}
		\mathbb{E}_{Y\sim\pi(\cdot\mid X),Y^{\prime}\sim\pi_{\text{ref}}(\cdot\mid X)}g(Y,Y^{\prime}\mid X)-p^*(\pi)\in\mathcal{T}_{f},
	\end{equation*}
	and then
	\begin{equation*}
		\begin{split}
			&\frac{1}{2}\Bigl\{\frac{\pi(y_1 \mid x)\pi_{\text{ref}}(y_2 \mid x)}{\pi_{\text{Gen}}^{\text{Hum}}(y_1 \mid x)\pi_{\text{Gen}}^{\text{Hum}}(y_2 \mid x)}-\frac{\pi(y_2 \mid x)\pi_{\text{ref}}(y_1 \mid x)}{\pi_{\text{Gen}}^{\text{Hum}}(y_2 \mid x)\pi_{\text{Gen}}^{\text{Hum}}(y_1 \mid x)}\Bigr\}(Z - g(y_1, y_2\mid x))\in\mathcal{T}_{g}.
		\end{split}
	\end{equation*}
	The multiplier in this last display is antisymmetric, and the lemma's second-moment condition ensures its variance-weighted square is integrable. Thus it belongs to the stated tangent-space closure. This completes the proof.\hfill $\blacksquare$
\end{proof}

We then prove the main theorem, which establishes the asymptotic expansion of the proposed DIPO estimator.

\begin{proof}[Proof of Theorem \ref{thm:mse}]
	Condition on the independent nuisance-training data $\mathcal T$ and keep $\pi$ and $\pi_{\mathrm{ref}}$ fixed. Throughout this proof, population expectations and norms are conditional on $\mathcal T$. Thus the fitted weights and $\tilde g$ are fixed functions, and the evaluation observations are i.i.d. within each source. Let $\mathbb E_n$ and $\mathbb E_N$ denote the empirical averages over the human and AI evaluation samples, respectively, with population counterparts $\mathbb E_{\mathrm{Hum}}$ and $\mathbb E_{\mathrm{AI}}$. Write $U=(X,Y^{(1)},Y^{(2)})$, set $d=\tilde g-g$, and define
	$$
	a_\pi(U)=\frac{w_\pi^{(1)}(U)-w_\pi^{(2)}(U)}2,
	\qquad
	\widehat a_\pi(U)
	=\frac{\widehat w_\pi^{(1)}(U)-\widehat w_\pi^{(2)}(U)}2,
	$$
	using the joint weights in \eqref{eq:dipo-weight-moment}. For a direct observation, put
	$$
	A_d(X,Y,Y^{(1)},Y^{(2)})=\frac12\sum_{a=1}^2d(Y,Y^{(a)}\mid X).
	$$
	
	We first verify the cancellation between the population direct error and the weighted human correction. The comparison symmetry of $g$ and $\tilde g$ gives
	$$
	d(y_2,y_1\mid x)=-d(y_1,y_2\mid x).
	$$
	Under the human comparison law, $Y^{(1)}$ and $Y^{(2)}$ are exchangeable conditional on $X$. Since $w_\pi^{(2)}(x,y_1,y_2)=w_\pi^{(1)}(x,y_2,y_1)$, interchanging the two responses yields
	\begin{align*}
		\mathbb E_{\mathrm{Hum}}\{w_\pi^{(2)}(U)d(U)\}=\mathbb E_{\mathrm{Hum}}\{w_\pi^{(1)}(U)d(Y^{(2)},Y^{(1)}\mid X)\}=-\mathbb E_{\mathrm{Hum}}\{w_\pi^{(1)}(U)d(U)\}.
	\end{align*}
	It follows that $\mathbb E_{\mathrm{Hum}}(a_\pi d)=\mathbb E_{\mathrm{Hum}}\{w_\pi^{(1)}d\}$. The weight $w_\pi^{(1)}$ changes the human response-pair law to the target/reference comparison law. Therefore,
	\begin{align*}
		\mathbb E_{\mathrm{Hum}}\{w_\pi^{(1)}d\}=\mathbb E_{X,\,Y\sim\pi(\cdot\mid X),\,Y'\sim\pi_{\mathrm{ref}}(\cdot\mid X)}d(Y,Y'\mid X)=\mathbb E_{\mathrm{AI}}A_d.
	\end{align*}
	All these expectations are well-defined: $|d|\le1$, and the weight moment condition with $\delta>2$ implies integrability. We next derive the remainder exactly. From \eqref{eq:est_dipo}, we have
	$$
	\hat P_{\mathrm{DIPO}}(\pi)
	=\hat P_{\mathrm{DM}}(\pi)
	+\mathbb E_n\{\widehat a_\pi(Z-\widehat Z)\}.
	$$
	Substituting $\tilde g=g+d$ in the direct term, and using $Z-\widehat Z=(Z-g)-(\widehat Z-\tilde g)-d$, for the part of the correction involving $a_\pi$, gives
	\begin{align*}
		\hat P_{\mathrm{DIPO}}(\pi)=&\frac1{2N}\sum_{i=n+1}^{n+N}\sum_{a=1}^2g(Y_i,Y_i^{(a)}\mid X_i)+\mathbb E_n\{a_\pi(Z-g)\}-\mathbb E_n\{a_\pi(\widehat Z-\tilde g)\}\\
		&+\mathbb E_N A_d-\mathbb E_n(a_\pi d)+\mathbb E_n\{(\widehat a_\pi-a_\pi)(Z-\widehat Z)\}.
	\end{align*}
	By the definition of $a_\pi$, the first three terms are precisely the leading terms in \eqref{eq:mse}. Hence $R=\mathbb E_N A_d-\mathbb E_n(a_\pi d)+\mathbb E_n\{(\widehat a_\pi-a_\pi)(Z-\widehat Z)\}$. The conditional label means satisfy
	$$
	\mathbb E(Z-\widehat Z\mid U,\mathcal T)
	=g(U)-\tilde g(U)=-d(U),
	$$
	which does not require conditional independence of $Z$ and $\widehat Z$. Since $\widehat a_\pi-a_\pi$ is a function of $U$ and $\mathcal T$, iterated expectation gives
	$$
	\mathbb E_{\mathrm{Hum}}\{(\widehat a_\pi-a_\pi)(Z-\widehat Z)\}
	=-\mathbb E_{\mathrm{Hum}}\{(\widehat a_\pi-a_\pi)d\}.
	$$
	Centering the three empirical averages and using
	$\mathbb E_{\mathrm{AI}}A_d=\mathbb E_{\mathrm{Hum}}(a_\pi d)$, we obtain
	\begin{align*}
		R=&(\mathbb E_N-\mathbb E_{\mathrm{AI}})A_d-(\mathbb E_n-\mathbb E_{\mathrm{Hum}})(a_\pi d)+(\mathbb E_n-\mathbb E_{\mathrm{Hum}})\{(\widehat a_\pi-a_\pi)(Z-\widehat Z)\}\\&-\mathbb E_{\mathrm{Hum}}\{(\widehat a_\pi-a_\pi)d\}.
	\end{align*}
	In particular, we have $\mathbb E(R\mid\mathcal T)=-\mathbb E_{\mathrm{Hum}}\{(\widehat a_\pi-a_\pi)d\}$. We now bound the summands and the remaining bias. First, the pointwise inequality
	$$
	A_d^2\le\frac12\sum_{a=1}^2d(Y,Y^{(a)}\mid X)^2
	$$
	and the common marginal comparison law imply
	$$
	\|A_d\|_{2,\mathrm{AI}}^2
	\le\mathbb E_{X,\,Y\sim\pi,\,Y'\sim\pi_{\mathrm{ref}}}
	|d(Y,Y'\mid X)|^2
	=\|\tilde g-g\|^2.
	$$
	This argument allows the two comparisons in a direct observation to share the same target response $Y$. Next, $|a_\pi|\le\bar w$ and \eqref{eq:dipo-weight-moment} give $\|a_\pi\|_{\delta,\mathrm{Hum}}\le M_\delta^{1/\delta}$. H\"older's inequality with conjugate exponents $\delta/2$ and $\delta/(\delta-2)$ yields
	\begin{align*}
		\mathbb E_{\mathrm{Hum}}(a_\pi^2d^2)
		&\le\{\mathbb E_{\mathrm{Hum}}|a_\pi|^\delta\}^{2/\delta}
		\{\mathbb E_{\mathrm{Hum}}|d|^{2\delta/(\delta-2)}\}^{(\delta-2)/\delta},\\
		\|a_\pi d\|_{2,\mathrm{Hum}}
		&\le M_\delta^{1/\delta}
		\|d\|_{2\delta/(\delta-2),\mathrm{Hum}}.
	\end{align*}
	Also, the triangle inequality and \eqref{eq:moment-dipo-nuisance} give
	$$
	\|\widehat a_\pi-a_\pi\|_{\delta,\mathrm{Hum}}
	\le\frac12\sum_{a=1}^2
	\|\widehat w_\pi^{(a)}-w_\pi^{(a)}\|_{\delta,\mathrm{Hum}}
	\le e_\delta.
	$$
	Since $|Z-\widehat Z|\le1$ and $\delta>2$, monotonicity of probability-space norms implies
	\begin{align*}
		\|(\widehat a_\pi-a_\pi)(Z-\widehat Z)\|_{2,\mathrm{Hum}}\le\|\widehat a_\pi-a_\pi\|_{2,\mathrm{Hum}}\le\|\widehat a_\pi-a_\pi\|_{\delta,\mathrm{Hum}}\le e_\delta.
	\end{align*}
	Finally, H\"older's inequality with conjugate exponents $\delta$ and $\delta/(\delta-1)$ gives
	$$
	|\mathbb E(R\mid\mathcal T)|
	\le\|\widehat a_\pi-a_\pi\|_{\delta,\mathrm{Hum}}
	\|d\|_{\delta/(\delta-1),\mathrm{Hum}}
	\le e_\delta\|d\|_{\delta/(\delta-1),\mathrm{Hum}}.
	$$
	It remains to control the centered empirical terms. For any square-integrable human summand $h$, conditional i.i.d. sampling gives
	$$
	\mathbb E\big[\{(\mathbb E_n-\mathbb E_{\mathrm{Hum}})h\}^2
	\mid\mathcal T\big]
	=\frac1n\mathbb E_{\mathrm{Hum}}
	\{h-\mathbb E_{\mathrm{Hum}}h\}^2
	\le\frac{\|h\|_{2,\mathrm{Hum}}^2}{n}.
	$$
	The analogous inequality holds for the AI sample with $N$ and $\mathbb E_{\mathrm{AI}}$. Apply these inequalities to the three centered terms in $R$, followed by the conditional $L^2$ triangle inequality. The preceding bounds yield
	\begin{align*}
		\{\mathbb E\big[\{R-\mathbb E(R\mid\mathcal T)\}^2\mid\mathcal T\big]\}^{1/2}\le\frac{\|\tilde g-g\|}{\sqrt N}+\frac{M_\delta^{1/\delta}\|d\|_{2\delta/(\delta-2),\mathrm{Hum}}+e_\delta}{\sqrt n}.
	\end{align*}
	The two centered human averages use the same observations and need not be independent; the triangle inequality accounts for this dependence. Conditional Chebyshev's inequality now shows that, with probability at least $1-\alpha$ for any $0<\alpha<1$,
	\begin{align*}
		|R|\le\frac1{\sqrt\alpha}\bigg[\frac{\|\tilde g-g\|}{\sqrt N}+\frac{M_\delta^{1/\delta}\|d\|_{2\delta/(\delta-2),\mathrm{Hum}}+e_\delta}{\sqrt n}\bigg]+e_\delta\|d\|_{\delta/(\delta-1),\mathrm{Hum}}.
	\end{align*}
	Substituting $d=\tilde g-g$ and the norm convention stated at the beginning of the appendix gives the remainder bound in Theorem~\ref{thm:mse}.
	\hfill $\blacksquare$
\end{proof}

We next prove the efficiency comparison under the stated variance and mean-square remainder conditions.

\begin{proof}[Proof of Corollary \ref{cor:efficiency_comparison}]
	The conditional mean of the direct summand defining $V_{\mathrm A}$ is $\mu_\pi(X)$, so total variance gives $V_{\mathrm A}\ge V_X$. The positive variance-gap condition therefore implies $V_X>0$ and $N>n$ eventually. The proof of Theorem~\ref{thm:mse} bounds the conditional $L^2$ norm of the centered remainder by the three standard-error terms in that theorem. Since $|\tilde g-g|\le1$ and $e_{\delta}\le2M_{\delta}^{1/\delta}$, the assumed convergence in probability of these bounded nuisance norms also implies their convergence in mean square. The product condition and $N>n$ therefore give $R=o_{L^2}(n^{-1/2})$. For $\hat P_{\mathrm{DIPO}}^{\mathrm{prob}}(\pi)$, the algebra of Theorem~\ref{thm:mse} uses $Z-\tilde g(Y^{(1)},Y^{(2)}\mid X)$ in the correction, whose absolute value is at most one. Its remainder therefore obeys the same bound, and its expansion has no binary-label noise term. With
	$$
	A_g=\frac12\sum_{a=1}^2g(Y,Y^{(a)}\mid X),
	\qquad
	\xi_\pi=a_\pi(U)\{Z-g(U)\},
	$$
	where $a_\pi$ is defined in the preceding proof, we obtain
	$$
	\hat P_{\mathrm{DIPO}}^{\mathrm{prob}}(\pi)-P(\pi)=(\mathbb E_N-\mathbb E_{\mathrm{AI}})A_g+\mathbb E_n\xi_\pi+o_{L^2}(n^{-1/2}).
	$$
	Write $V_{\mathrm H}=\mathbb E_{\mathrm{Hum}}[a_\pi(U)^2g(U)\{1-g(U)\}]<\infty$. The two evaluation samples are independent, and $\mathbb E(\xi_\pi\mid X)=0$. Consequently,
	$$
	\mathrm{MSE}(\hat P_{\mathrm{DIPO}}^{\mathrm{prob}}(\pi))=\frac{V_{\mathrm A}}N+\frac{V_{\mathrm H}}n+o(n^{-1}).
	$$
	The regular human-only estimator has the efficient influence function
	$\mu_\pi(X)-P(\pi)+\xi_\pi$, so
	$$
	\sqrt n\{\hat P_{\mathrm{IPO}}(\pi)-P(\pi)\}
	\stackrel{d}{\to}N(0,V_X+V_{\mathrm H}).
	$$
	Let $G\sim N(0,V_X+V_{\mathrm H})$. For each fixed $M>0$,
	the function $x\mapsto x^2\wedge M=\min\{x^2,M\}$ is bounded
	and continuous. The preceding convergence in distribution
	therefore implies
	$$
	\mathbb E\big[\min\{n\{\hat P_{\mathrm{IPO}}(\pi)-P(\pi)\}^2,M\}\big]\longrightarrow \mathbb E[\min\{G^2,M\}].
	$$
	Since truncation can only decrease a nonnegative squared error,
	\begin{align*}
		n\mathrm{MSE}(\hat P_{\mathrm{IPO}}(\pi))=\mathbb E[n\{\hat P_{\mathrm{IPO}}(\pi)-P(\pi)\}^2]\ge\mathbb E[\min\{n\{\hat P_{\mathrm{IPO}}(\pi)-P(\pi)\}^2,M\}].
	\end{align*}
	Taking the lower limit as $n\to\infty$ gives, for every fixed $M>0$, we thus have 
	$$\liminf_{n\to\infty}n\mathrm{MSE}(\hat P_{\mathrm{IPO}}(\pi))\ge \mathbb E[\min\{G^2,M\}].$$
	As $M\to\infty$, $\min\{G^2,M\}$ increases to $G^2$. By the monotone convergence theorem, we have $\mathbb E[\min\{G^2,M\}]\longrightarrow\mathbb E G^2=V_X+V_{\mathrm H}$. Consequently,
	$$
	\liminf_{n\to\infty}
	n\,\mathrm{MSE}(\hat P_{\mathrm{IPO}}(\pi))
	\ge V_X+V_{\mathrm H}.
	$$
	This lower bound and the stated positive variance gap prove the strict comparison; no mean-square remainder condition for the human-only estimator is needed. The condition is needed because response sampling in $A_g$ can make $V_{\mathrm A}$ exceed $V_X$.
	
	In the augmented evaluation model, we condition on the external nuisance-training data. Target and reference response laws are fixed, and evaluator calls provide no additional modeled information about the human preference mechanism. The prompt-law score is shared by all $n+N$ prompts, whereas the preference score occurs only in the $n$ human observations. The orthogonal tangent components in Lemma~\ref{lem:semi} therefore give the efficiency bound
	$$
	\frac{V_{\mathrm H}}n+\frac{V_X}{n+N}.
	$$
	Indeed, writing $\mu_c(X)=\mu_\pi(X)-P(\pi)$, the efficient contributions under separate-sample averaging are $\xi_\pi+n\mu_c/(n+N)$ for a human observation and $N\mu_c/(n+N)$ for an additional prompt. Since $\mathbb E(\xi_\pi\mid X)=0$, their combined variance is the displayed bound. When $N/n\to\infty$, this is $V_{\mathrm H}/n+o(n^{-1})$. The bounded direct summand and the preceding expansion give
	$$
	\hat P_{\mathrm{DIPO}}^{\mathrm{prob}}(\pi)-P(\pi)=\mathbb E_n\xi_\pi+o_{L^2}(n^{-1/2}).
	$$
	Under the stated regularity condition, this attains the limiting bound for fixed-policy evaluation. The pointwise nuisance rates alone are not being used to infer regularity.
	\hfill $\blacksquare$
\end{proof}

Finally, we establish the regret bound for the finite-step DIPO output.

\begin{proof}[Proof of Theorem~\ref{thm:regret_dipo}]
	Condition on the independent nuisance-training sample $\mathcal T$. Recall the conditional mean objective $\overline J$ in \eqref{eq:dipo}, and let $\widehat J_t$ be the fresh corrected score at $\pi^{(t)}$ defined in \eqref{eq:dipo-checkpoint-score}. We first control the reused human sample, then the fresh AI evaluations, and finally the computational gap.
	
	By \eqref{eq:rad-dipo-human}, ordinary symmetrization gives
	\begin{align*}
		\mathbb E\bigg[\sup_{\pi\in\Pi}\bigg|\frac1n\sum_{i=1}^nB_i(\pi)-\mathbb E_{\mathrm{Hum}}B(\pi)\bigg|\,\bigg|\,\mathcal T\bigg]\le2\mathbb E\bigg[\sup_{\pi\in\Pi}\bigg|\frac1n\sum_{i=1}^n\varepsilon_iB_i(\pi)\bigg|\,\bigg|\,\mathcal T\bigg]\le C\bigg(\sqrt{\frac vn}+\frac vn\bigg).
	\end{align*}
	Markov's inequality therefore yields
	$$
	\sup_{\pi\in\Pi}\bigg|\frac1n\sum_{i=1}^nB_i(\pi)-\mathbb E_{\mathrm{Hum}}B(\pi)\bigg|=O_p\bigg(\sqrt{\frac vn}+\frac vn\bigg).
	$$
	This supremum permits every iterate to depend on the reused human sample. For each fixed $\pi$, exchangeability of the human response pair, comparison symmetry, and change of measure give
	\begin{align*}
		\overline P_{\mathrm{DM}}(\pi)-P(\pi)=\mathbb E_{\mathrm{Hum}}\{(\tilde g-g)w_\pi^{(1)}\},\quad
		\mathbb E_{\mathrm{Hum}}B(\pi)=-\mathbb E_{\mathrm{Hum}}\{(\tilde g-g)\hat w_\pi^{(1)}\}.
	\end{align*}
	The support and moment conditions ensure integrability. Adding these identities and applying H\"older's inequality gives
	$$
	\sup_{\pi\in\Pi}|\overline P_{\mathrm{DM}}(\pi)+\mathbb E_{\mathrm{Hum}}B(\pi)-P(\pi)|\le e_\delta\|\tilde g-g\|_{\delta/(\delta-1),\mathrm{Hum}}.
	$$
	Consequently, if $e_{\mathrm H}:=\sup_{\pi\in\Pi}|\overline P_{\mathrm{DM}}(\pi)-\widehat{\mathrm{Bias}}(\pi)-P(\pi)|$, then
	\begin{equation}\label{eq:dipo-human-deviation}
		e_{\mathrm H}=O_p\bigg\{\sqrt{\frac vn}+\frac vn+e_\delta\|\tilde g-g\|_{\delta/(\delta-1),\mathrm{Hum}}\bigg\}.
	\end{equation}
	
	Next, let $\mathcal F_t$ contain $\mathcal T$, the human sample, and all optimization and sampling history up to the formation of $\pi^{(t)}$, before its fresh evaluation batch. Thus $\pi^{(t)}$ is $\mathcal F_t$-measurable. In that batch, put
	$$
	A_{i,t}=\frac12\sum_{a=1}^2
	\tilde g(Y_{i,t},Y_{i,t}^{(a)}\mid X_{i,t}),
	\qquad i=n+1,\ldots,n+N.
	$$
	Conditional on $\mathcal F_t$, these variables are independent, take values in $[0,1]$, and have common mean $\overline P_{\mathrm{DM}}(\pi^{(t)})$. The human correction and KL penalty are already $\mathcal F_t$-measurable. Hence, for every $u>0$, the exponential bound for averages of bounded independent variables gives
	$$
	\mathbb P\{|\widehat J_t-\overline J(\pi^{(t)})|>u\mid\mathcal F_t\}\le2\exp(-2Nu^2).
	$$
	Take expectations and sum over $t=0,\ldots,T$. With $\Delta_{N,T}:=\max_{0\le t\le T}|\widehat J_t-\overline J(\pi^{(t)})|$, this gives
	\begin{equation}\label{eq:dipo-fresh-concentration}
		\mathbb P\bigg\{\Delta_{N,T}>\sqrt{\frac{\log\{2(T+1)/\alpha\}}{2N}}\bigg\}\le\alpha,\qquad 0<\alpha<1.
	\end{equation}
	The argument allows later policies to depend on earlier batches; independence across the whole trajectory is not required. In particular, $\Delta_{N,T}=O_p\{\sqrt{\log\{2(T+1)\}/N}\}$. By the definition of $\varepsilon_{\mathrm{opt},T}$ in \eqref{eq:dipo-computation-gap},
	\begin{align*}
		\overline J(\pi^*)-\overline J(\pi^{(T)})\le \sup_{\pi\in\Pi}\overline J(\pi)-\max_{0\le t\le T}\overline J(\pi^{(t)})+\max_{0\le t\le T}\widehat J_t-\widehat J_T+2\Delta_{N,T}=\varepsilon_{\mathrm{opt},T}+2\Delta_{N,T}.
	\end{align*}
	The two terms forming the computational gap are nonnegative. Write $J(\pi)=P(\pi)-\beta D_{\mathrm{KL}}(\pi\|\pi_{\mathrm{ref}})$. Since $\sup_{\pi\in\Pi}|J(\pi)-\overline J(\pi)|\le e_{\mathrm H}$, we obtain
	$$
	J(\pi^*)-J(\pi^{(T)})\le 2e_{\mathrm H}+2\Delta_{N,T}+\varepsilon_{\mathrm{opt},T}.
	$$
	Moreover, $\pi^*$ maximizes $J$, and every KL divergence lies in $[0,K]$. Therefore,
	\begin{equation}\label{eq:dipo-trajectory-comparison}
		-\beta K\le P(\pi^*)-P(\pi^{(T)})\le\beta K+2e_{\mathrm H}+2\Delta_{N,T}+\varepsilon_{\mathrm{opt},T}.
	\end{equation}
	Combining \eqref{eq:dipo-human-deviation}--\eqref{eq:dipo-trajectory-comparison} proves the theorem. Neither the stochastic bounds nor the bias identity requires nuisance consistency. With true weights, the product term is zero, and with $e_\delta=o_p(1)$ it vanishes because $|\tilde g-g|\le1$. These conclusions concern policy regret; the fixed-policy efficiency result retains its separate assumptions.
\end{proof}

\subsection{Rademacher complexity conditions}
\label{sec:rad-complexity}

This subsection specifies Assumption~\ref{assump:model-complexity} using ordinary Rademacher complexity for both methods. Let $\varepsilon_i$ be independent signs taking values $-1$ and $1$ with equal probability, independent of the observations and the nuisance-training sample. The three displayed bounds (\eqref{eq:rad-ddpo-ai}, \eqref{eq:rad-ddpo-human}, and \eqref{eq:rad-dipo-human}) use ordinary Rademacher complexity with the absolute-value convention. Expectations are over the observations and the signs, conditional on an independent sample used to fit the nuisance functions. The DB-DPO evaluation samples are i.i.d. within each source and independent across sources, conditional on nuisance fitting. For DIPO, the human comparisons and human labels are i.i.d. and independent of nuisance fitting, while the augmented observations containing the paired AI labels are i.i.d. conditional on nuisance fitting. Its fresh AI batches satisfy the conditional sampling requirements of Theorem~\ref{thm:regret_dipo}. Constants may depend on the fixed coverage and reward bounds and the fixed moment order and constants, but not on $n$, $N$, $T$, or $\nu_n$.

For DB-DPO, use $r^\pi$ defined in \eqref{eq:rpi} and define
$$
f_\pi(X,Y^{(1)},Y^{(2)})=r^\pi(Y^{(1)},X)-r^\pi(Y^{(2)},X)-r^*(Y^{(1)},X)+r^*(Y^{(2)},X),
$$
$$
\sigma_\pi^2=\mathbb E_{X\sim\mathbb P_X,(Y^{(1)},Y^{(2)})\sim\pi_{\mathrm{Gen}}^{\mathrm{AI}}(\cdot\mid X)}f_\pi^2.
$$
To simplify the notation, we denote
$$
\ell_i(\pi,z)=\log\sigma\big[(2z-1)\{r^\pi(Y_i^{(1)},X_i)-r^\pi(Y_i^{(2)},X_i)\}\big],
$$
let $a_i(\pi)=\ell_i(\pi,\hat Z_i)-\ell_i(\pi^*,\hat Z_i)$ and $b_i(\pi)=w_i(Z_i-\hat Z_i)f_\pi(X_i,Y_i^{(1)},Y_i^{(2)})$, where $w_i=\hat w(Y_i^{(1)},Y_i^{(2)}\mid X_i)$, or the true weight when known. Suppose that
\begin{align}
	\mathbb E\sup_{\pi\in\Pi}
	\bigg|\frac1N\sum_{i=n+1}^{n+N}\varepsilon_i a_i(\pi)\bigg|
	&\le C\sqrt{\frac vN},\label{eq:rad-ddpo-ai}\\
	\mathbb E\sup_{\pi\in\Pi}
	\bigg|\frac1n\sum_{i=1}^n\varepsilon_i b_i(\pi)\bigg|
	&\le C\sqrt{\frac{v\nu_n}{n}}.\label{eq:rad-ddpo-human}
\end{align}
In particular, the factor $\nu_n$ in \eqref{eq:rad-ddpo-human} is part of the complexity assumption on the weighted correction class.

For DIPO, let $B_i(\pi)$ be the human correction summand in \eqref{eq:est_dipo}, using fitted generation probabilities when needed, so that
$$
-\widehat{\mathrm{Bias}}(\pi)=\frac1n\sum_{i=1}^nB_i(\pi).
$$
Assume
\begin{equation}\label{eq:rad-dipo-human}
	\mathbb E\sup_{\pi\in\Pi}\bigg|
	\frac1n\sum_{i=1}^n\varepsilon_iB_i(\pi)\bigg|
	\le C\bigg(\sqrt{\frac vn}+\frac vn\bigg).
\end{equation}
This uniform bound permits the policy iterates to depend on the reused human sample. No uniform Rademacher condition is imposed on the sampled DIPO direct term. We control its fresh evaluations along the visited policies by \eqref{eq:dipo-fresh-concentration}; their prompts and both target and reference responses must be drawn after the evaluated policy is determined. Appendix~\ref{sec:computational-errors} gives the computational and loss-approximation terms.

\subsection{Examples of Rademacher complexity bounds}
\label{sec:rad-examples}

We summarize two norm-based results and give sufficient conditions for the bounds in Appendix~\ref{sec:rad-complexity}. For a class $\mathcal H$ of real-valued functions and a deterministic sample $o_1,\ldots,o_k$, define the ordinary empirical Rademacher complexity by
$$
\widehat{\mathfrak R}_k(\mathcal H)
=\mathbb E_\varepsilon\sup_{h\in\mathcal H}
\bigg|\frac1k\sum_{i=1}^k\varepsilon_i h(o_i)\bigg|,
$$
where $k$ is the sample size and the signs are defined in Appendix~\ref{sec:rad-complexity}. In the bounds below, $B_{\mathrm{env}}>0$ is a fixed uniform bound on the absolute values of the functions under consideration. In the applications, this bound concerns the bounded base and normalized correction classes. For a covering radius $\tau>0$, $\mathcal N(\tau,\mathcal F,d_{\mathrm{emp}})$ denotes the minimum number of balls of radius $\tau$ needed to cover a function class $\mathcal F$ in an empirical distance $d_{\mathrm{emp}}$, specified below. The bounds have no explicit parameter-count factor, but retain logarithmic dimension and sample-size factors and depend on depth through the stated norm quantities.

\paragraph{Example 1 (DNNs: \citealp{bartlett2017spectrally}).}

Let $L_{\mathrm{DNN}}$ denote the network depth, and let $\mathcal F_{\mathrm{DNN}}$ be the class of vector-valued output maps of these networks without bias terms. The input is a fixed Euclidean encoding of the prompt $X$. At layer $\ell$, the activation is fixed, vanishes at zero, and has Lipschitz constant $\kappa_\ell$. Theorem~3.3 in \cite{bartlett2017spectrally} applies when the weight matrices $\mathbf A_\ell$ satisfy
$$
\|\mathbf A_\ell\|_{\mathrm{op}}\le s_\ell,
\qquad
\|(\mathbf A_\ell-\mathbf A_\ell^{0})^\top\|_{2,1}\le \zeta_\ell,
$$
where $\mathbf A_\ell^{0}$ are fixed reference matrices of the same dimensions as $\mathbf A_\ell$, and $s_\ell,\zeta_\ell>0$ are fixed norm bounds. For any matrix, $\|\cdot\|_{\mathrm{op}}$ denotes its Euclidean operator norm and $\|\cdot\|_{2,1}$ the sum of its column Euclidean norms. Define the norm complexity
$$
\Gamma_{\mathrm{DNN}}=\bigg(\prod_{\ell=1}^{L_{\mathrm{DNN}}}\kappa_\ell s_\ell\bigg)\bigg\{\sum_{\ell=1}^{L_{\mathrm{DNN}}}(\zeta_\ell/s_\ell)^{2/3}\bigg\}^{3/2}.
$$
Let $B_{\mathrm{in}}$ bound the Euclidean norm of the input encoding, and let $d_{\max}$ be the maximum input or layer dimension. Their joint output covering bound is
\begin{gather*}
	\log\mathcal N(\tau,\mathcal F_{\mathrm{DNN}},d_{2,k})
	\le\frac{B_{\mathrm{in}}^2\Gamma_{\mathrm{DNN}}^2\log(2d_{\max}^2)}{\tau^2},\\
	d_{2,k}(F,G)^2
	=\frac1k\sum_{i=1}^k\|F(X_i)-G(X_i)\|_2^2.
\end{gather*}
Here $F,G\in\mathcal F_{\mathrm{DNN}}$, and $d_{2,k}$ is the normalized empirical Euclidean distance on $X_1,\ldots,X_k$. Consequently, any class $\mathcal H$ of real-valued functions obtained by fixed, uniformly Lipschitz transformations of these output vectors, with envelope $B_{\mathrm{env}}$, satisfies
$$
\widehat{\mathfrak R}_k(\mathcal H)\le\frac C{\sqrt k}\big\{B_{\mathrm{env}}+B_{\mathrm{in}}\Gamma_{\mathrm{DNN}}\sqrt{\log(2d_{\max}^2)}\log(2k)\big\}.
$$
The constant $C$ includes the fixed Lipschitz constant. The explicit width dependence is logarithmic; the matrix norms may themselves depend on width.

\paragraph{Example 2 (Transformers: \citealp{edelman2022inductive}).}

Corollary~4.5 in \cite{edelman2022inductive} gives a covering bound for a self-attention head; Theorem~4.7 in \cite{edelman2022inductive} extends it to multilayer architectures with the stated normalization. We use only the single-head bound here for illustration.

Let $L_{\mathrm{seq}}$ be the number of tokens and $d_{\mathrm{emb}}$ the embedding dimension. The input-response token matrix $\mathbf S\in\mathbb R^{L_{\mathrm{seq}}\times d_{\mathrm{emb}}}$ has $j$th row $\mathbf s_j^\top$, with fixed input and positional embeddings and $\|\mathbf s_j\|_2\le1$. Fix a query position $j_0\in\{1,\ldots,L_{\mathrm{seq}}\}$ and the output weights $\mathbf c_{\mathrm{out}}\in\mathbb R^{d_{\mathrm{emb}}}$ with $\|\mathbf c_{\mathrm{out}}\|_2\le B_{\mathrm{out}}$. The scalar head output is
$$
h_\theta(\mathbf S)=\mathbf c_{\mathrm{out}}^\top\operatorname{Act}_{\mathrm{out}}\big\{\mathbf W_V^\top\mathbf S^\top\operatorname{softmax}(\mathbf S\mathbf W_{QK}^\top\mathbf s_{j_0})\big\},
$$
Here $\mathbf s_{j_0}$ is the query-token vector, $\mathbf W_V$ is the value matrix, $\mathbf W_{QK}$ is the combined query-key matrix, and the output activation $\operatorname{Act}_{\mathrm{out}}:\mathbb R^{d_{\mathrm{emb}}}\to\mathbb R^{d_{\mathrm{emb}}}$ is fixed, satisfies $\operatorname{Act}_{\mathrm{out}}(0)=0$, and is $\kappa_{\mathrm{out}}$-Lipschitz in Euclidean norm. Both matrices belong to $\mathbb R^{d_{\mathrm{emb}}\times d_{\mathrm{emb}}}$, and $\theta$ collects their trainable entries. The softmax map sends a vector to the probabilities proportional to its exponentiated coordinates. Assume the fixed norm bounds
$$
\|\mathbf W_V\|_{\mathrm{op}}\le s_V,
\qquad
\|\mathbf W_V\|_{2,1}\le\zeta_V,
\qquad
\|\mathbf W_{QK}^{\top}\|_{2,1}\le\zeta_{QK},
$$
where $s_V,\zeta_V,\zeta_{QK},B_{\mathrm{out}}>0$ are constants. Let $\mathcal F_{\mathrm{TF}}$ denote the resulting scalar output class, and define
$$
\Gamma_{\mathrm{TF}}=\kappa_{\mathrm{out}}B_{\mathrm{out}}\big\{\zeta_V^{2/3}+(\zeta_{QK}s_V)^{2/3}\big\}^{3/2}.
$$
For deterministic token matrices $\mathbf S_1,\ldots,\mathbf S_k$, define
$$
d_{\infty,k}(h,h')=\max_{1\le i\le k}|h(\mathbf S_i)-h'(\mathbf S_i)|.
$$
Let $\mathcal N_\infty(\tau,\mathcal F_{\mathrm{TF}},k)$ denote the supremum, over all admissible samples of size $k$, of the covering number under $d_{\infty,k}$. Choose $B_{\mathrm{env}}$ large enough to bound the head outputs uniformly. The proof of Corollary~4.5 in \citet{edelman2022inductive} yields the following bound, with logarithmic factors displayed explicitly, for $0<\tau\le B_{\mathrm{env}}$:
$$
\log\mathcal N_\infty(\tau,\mathcal F_{\mathrm{TF}},k)\le\frac{C\Gamma_{\mathrm{TF}}^2}{\tau^2}\log\bigg(2+\frac{C_0d_{\mathrm{emb}}kL_{\mathrm{seq}}}{\tau}\bigg).
$$
Here $C_0$ depends on fixed norm and envelope bounds. Truncated Dudley integration therefore gives
$$
\widehat{\mathfrak R}_k(\mathcal F_{\mathrm{TF}})\le\frac C{\sqrt k}\big\{B_{\mathrm{env}}+\Gamma_{\mathrm{TF}}\sqrt{\log(2+C_0d_{\mathrm{emb}}kL_{\mathrm{seq}})}\log(2k)\big\}.
$$
Thus the explicit embedding-dimension and context-length dependence is logarithmic, with no parameter-count factor; the bound still depends on the stated norm constraints.

\paragraph{Verification of Assumption~\ref{assump:model-complexity}.}
Examples 1 and 2 provide sufficient conditions for the DB-DPO complexity bounds and the DIPO human-correction bound in Appendix~\ref{sec:rad-complexity}. Let $\theta$ denote the trainable parameters, and let $h_\theta(X,y)$ be the network score of response $y$. On a fixed finite response set $\mathcal Y$, define
$$
\pi_\theta(y\mid X)
=\frac{\pi_{\mathrm{ref}}(y\mid X)\exp\{h_\theta(X,y)\}}
{\sum_{y'\in\mathcal Y}\pi_{\mathrm{ref}}(y'\mid X)\exp\{h_\theta(X,y')\}}.
$$
For DNNs, $h_\theta(X,\cdot)$ is the vector of response logits; for Transformers, $h_\theta(X,y)$ is the scalar score of the input-response sequence. Assume uniformly bounded logits and retain the corresponding theorem's reward, coverage, and moment conditions. For DB-DPO, also assume that the conditional disagreement probability given $(X,Y^{(1)},Y^{(2)})$ and this training sample is bounded by $C\nu_n$ almost surely, with fixed $C$.

For DB-DPO, the policy parameterization gives, on the reference support,
$$
\log\frac{\pi_\theta(y\mid X)}{\pi_{\mathrm{ref}}(y\mid X)}=h_\theta(X,y)-\log\sum_{y'\in\mathcal Y}\pi_{\mathrm{ref}}(y'\mid X)\exp\{h_\theta(X,y')\}.
$$
The log normalizer and the term $\beta\log W(X)$ in \eqref{eq:rpi} depend only on $X$, so both cancel when comparing two responses to the same prompt. Consequently,
$$
r^{\pi_\theta}(Y^{(1)},X)-r^{\pi_\theta}(Y^{(2)},X)=\beta\{h_\theta(X,Y^{(1)})-h_\theta(X,Y^{(2)})\}.
$$
The oracle reward difference in $f_\pi$ is fixed across policies. Thus, for $U=(X,Y^{(1)},Y^{(2)})$ and any $\theta,\theta'$, the triangle inequality gives
\begin{align*}
	|f_{\pi_\theta}(U)-f_{\pi_{\theta'}}(U)|\le\beta\sum_{a=1}^2|h_\theta(X,Y^{(a)})-h_{\theta'}(X,Y^{(a)})|\le2\beta\max_{y\in\mathcal Y}|h_\theta(X,y)-h_{\theta'}(X,y)|.
\end{align*}
Moreover, we have $(\log\sigma)'(u)=\sigma(-u)\in(0,1)$, so log-sigmoid is $1$-Lipschitz. For each fixed observation, $|2\hat Z_i-1|=1$ and the centering term $\ell_i(\pi^*,\hat Z_i)$ in $a_i(\pi)$ does not depend on $\theta$. Hence
\begin{align*}
	&|a_i(\pi_\theta)-a_i(\pi_{\theta'})|=|\ell_i(\pi_\theta,\hat Z_i)-\ell_i(\pi_{\theta'},\hat Z_i)|\le|f_{\pi_\theta}(U_i)-f_{\pi_{\theta'}}(U_i)|\\
	\le&2\beta\max_{y\in\mathcal Y} |h_\theta(X_i,y)-h_{\theta'}(X_i,y)|.
\end{align*}
These pointwise inequalities transfer the output covers to both policy classes, enlarging the radius by at most $2\beta$: for DNNs, the maximum coordinate difference is bounded by the output Euclidean norm; for Transformers, use a cover of the scalar scores at all required input-response pairs. The fixed oracle terms leave covering distances unchanged. Together with the bounded reward envelopes, the transferred covering bounds yield the corresponding ordinary Rademacher bounds by Dudley integration.

For DIPO, conditional on the human observations and nuisance fitting, the labels and fitted generation probabilities are fixed. Each fitted joint weight depends on the policy only through $\pi(Y_i^{(a)}\mid X_i)$. On the reference support, log-softmax gives
$$
|\log\pi_\theta(y\mid X)-\log\pi_{\theta'}(y\mid X)|
\le2\max_{y'\in\mathcal Y}|h_\theta(X,y')-h_{\theta'}(X,y')|.
$$
The envelope $\bar w$ in \eqref{eq:dipo-weight-moment} is independent of the policy index and dominates all true and fitted joint weights. Using $|a-b|\le\max(a,b)|\log a-\log b|$ for $a,b>0$ therefore yields
$$
\bigg|\frac{B_i(\pi_\theta)-B_i(\pi_{\theta'})}{\bar w(U_i)}\bigg|\le C\max_{y\in\mathcal Y}|h_\theta(X_i,y)-h_{\theta'}(X_i,y)|.
$$
Terms with zero reference probability vanish. Hence the network output covers transfer to the normalized human-correction class. For DNNs, the output Euclidean norm dominates the maximum coordinate norm, so the joint cover introduces no additional response-count factor. For Transformers, covering the $k|\mathcal Y|$ input-response sequences replaces $k$ by $k|\mathcal Y|$ inside the logarithm. Choosing covering centers within the restricted class enlarges the radius by at most a constant factor. A sufficient common complexity index is
\begin{equation}\label{eq:norm-based-v}
	\begin{split}
		v\ge C\log^2\{2(n+N)\}\max\Big\{&1+B_{\mathrm{in}}^2\Gamma_{\mathrm{DNN}}^2\log(2d_{\max}^2),\\
		&1+\Gamma_{\mathrm{TF}}^2\log\{2+C_0d_{\mathrm{emb}}L_{\mathrm{seq}}|\mathcal Y|(n+N)\}\Big\}.
	\end{split}
\end{equation}
For either example alone, retain its corresponding entry. Fixed envelopes and Lipschitz constants are absorbed into $C$, and all displayed logarithmic factors are included in $v$. The output bounds give $C\sqrt{v/k}$ for the bounded base classes on every admissible deterministic sample of size $1\le k\le n+N$. This verifies \eqref{eq:moment-base-ddpo} and \eqref{eq:moment-base-dipo}; averaging over the observations also verifies \eqref{eq:rad-ddpo-ai}.

For the weighted classes, apply the $\delta>2$ case of \eqref{eq:moment-multiplier}. Taking $(V,D,h_\pi)=(\hat w,Z-\hat Z,f_\pi)$ gives \eqref{eq:rad-ddpo-human}, and taking $(V,D,h_\pi)=(\bar w,1,B(\pi)/\bar w)$ gives \eqref{eq:rad-dipo-human}. Use $\hat w=w$ when the weight is known. Thus, Examples 1 and 2 verify all three complexity conditions under the stated coverage, moment, and disagreement conditions. The DIPO direct term uses the original sampled preference scores: conditional independence of each fresh batch gives \eqref{eq:dipo-fresh-concentration}.

For $1<\delta<2$, the same samplewise bounds and the corresponding multiplier rate verify the complexity conditions of Corollary~\ref{cor:ddpo-moment}(i) and Corollary~\ref{cor:dipo-moment}. The local refinement in Corollary~\ref{cor:ddpo-moment}(ii) requires the separate verification in Appendix \ref{sec:local-ddpo}.

\subsection{Local Rademacher refinement for DB-DPO}
\label{sec:local-ddpo}

Local Rademacher complexity restricts ordinary Rademacher averages to a neighborhood of the oracle; see \citet{bartlett2005local}. Here the neighborhood is defined by the population $L^2$ norm of the reward-difference error. Retain $f_\pi$, $\sigma_\pi$, $a_i(\pi)$, and $b_i(\pi)$ from Appendix~\ref{sec:rad-complexity}. For $v\ge1$, define
$$
\Pi(t)=\{\pi\in\Pi:\sigma_\pi\le t\},\qquad
\mu_n=\sqrt{\frac vN}+\sqrt{\frac{v\nu_n}{n}}.
$$
Conditional on the independent nuisance-training sample, assume that the following bounds hold almost surely, simultaneously for all $t\ge\mu_n$:
\begin{align}
	\mathbb E\sup_{\pi\in\Pi(t)}\bigg|\frac1N\sum_{i=n+1}^{n+N}\varepsilon_i a_i(\pi)\bigg|&\le Ct\sqrt{\frac vN},\label{eq:local-rad-ddpo-ai}\\
	\mathbb E\sup_{\pi\in\Pi(t)}\bigg|\frac1n\sum_{i=1}^n\varepsilon_i b_i(\pi)\bigg|&\le Ct\sqrt{\frac{v\nu_n}{n}}.\label{eq:local-rad-ddpo-human}
\end{align}
The expectations are over the evaluation observations and Rademacher signs, conditional on nuisance fitting. The constant $C$ is uniform in $t$, $n$, $N$, and $\nu_n$. In particular, \eqref{eq:local-rad-ddpo-human} concerns the correction class formed with the weights actually used in the estimator. Any logarithmic factors required by these bounds are included in $v$.

\begin{proof}[Proof of Corollary~\ref{cor:ddpo-local}]
	Write $\hat\pi=\hat\pi_{\mathrm{DB\text{-}DPO}}$ and condition on the nuisance-training sample $\mathcal T$. Define the centered empirical process
	\begin{align*}
		\mathbb G_{n,N}(\pi)=\mathbb E_{n,N}l(\pi)-\mathbb E_{n,N}l(\pi^*)-\mathbb E\{\mathbb E_{n,N}l(\pi)-\mathbb E_{n,N}l(\pi^*)\mid\mathcal T\}.
	\end{align*}
	Applying symmetrization separately to the two evaluation samples and using \eqref{eq:local-rad-ddpo-ai}--\eqref{eq:local-rad-ddpo-human} gives
	\begin{equation}\label{eq:local-ddpo-process}
		\mathbb E\big[\sup_{\pi\in\Pi(t)}|\mathbb G_{n,N}(\pi)|\mid\mathcal T\big]\le Ct\mu_n,\qquad t\ge\mu_n.
	\end{equation}
	The curvature and population-bias bounds in the proof of Theorem~\ref{thm:subopt-gap}, together with empirical optimality, yield
	$$
	c\sigma^2\le|\mathbb G_{n,N}(\hat\pi)|+O_p(1)d_n\sigma,
	$$
	where $\sigma=\sigma_{\hat\pi}$ and $d_n=\nu_n\epsilon^{-2}\|\hat w-w\|_2$. By the uniform conditional disagreement assumption, the $O_p(1)$ coefficient is measurable with respect to $\mathcal T$. For each fixed $\eta>0$, it is bounded by a constant $D_\eta$ on a $\mathcal T$-measurable event with probability at least $1-\eta$, for all sufficiently large sample sizes.
	
	On this event, choose $A\ge\max\{1,2D_\eta/c\}$ and set $R=A(\mu_n+d_n)$. The radius $R$ is fixed conditional on $\mathcal T$. Whenever $\sigma>R$, the bias term is at most $c\sigma^2/2$. Partitioning $\{\sigma>R\}$ into the shells $2^jR<\sigma\le2^{j+1}R$, $j\ge0$, and applying Markov's inequality and \eqref{eq:local-ddpo-process}, we obtain
	\begin{align*}
		\mathbb P(\sigma>R\mid\mathcal T)
		&\le\sum_{j=0}^{\infty}\mathbb P\bigg(\sup_{\pi\in\Pi(2^{j+1}R)}|\mathbb G_{n,N}(\pi)|\ge\frac c2\,2^{2j}R^2\bigg|\mathcal T\bigg)\\
		&\le C\sum_{j=0}^{\infty}\frac{2^{j+1}R\mu_n}{2^{2j}R^2}\le\frac{C\mu_n}{R}\le\frac CA.
	\end{align*}
	Every radius $2^{j+1}R$ exceeds $\mu_n$, so the stated local bounds apply throughout. Removing the conditioning and including the exceptional training event gives
	$$
	\mathbb P\{\sigma>A(\mu_n+d_n)\}\le\eta+C/A
	$$
	for sufficiently large sample sizes. Taking $\eta$ arbitrarily small and then $A$ sufficiently large gives
	$$
	\sigma=O_p(\mu_n+d_n)=O_p\bigg\{\sqrt{\frac vN}+\sqrt{\frac{v\nu_n}{n}}+\nu_n^{1/2}\|\hat w-w\|_2\bigg\},
	$$
	where the final, weaker bound uses $0<\nu_n\le1$ and fixed $\epsilon$. Finally, the reward-to-policy value bound in the proof of Theorem~\ref{thm:subopt-gap} gives
	$$
	|J(\pi^*)-J(\hat\pi)|\le\frac C{\beta\epsilon}\sigma,
	\qquad
	J(\pi)=\mathbb E_{X\sim\mathbb P_X,\,Y\sim\pi(\cdot\mid X)}r(X,Y).
	$$
	This proves \eqref{eq:subopt-dpo-local}.
	\hfill $\blacksquare$
\end{proof}

\paragraph{Examples satisfying the local conditions.}
The following examples verify \eqref{eq:local-rad-ddpo-ai}--\eqref{eq:local-rad-ddpo-human} for a trainable linear reward head applied to a frozen representation. Let $\varphi(X,Y)\in\mathbb R^d$ be bounded and either fixed or pretrained on data independent of the evaluation samples. Include any pretraining data in $\mathcal T$, condition on $\mathcal T$, and consider
$$
\pi_\theta(Y\mid X)=\frac{\pi_{\mathrm{ref}}(Y\mid X)\exp\{\theta^\top\varphi(X,Y)/\beta\}}{\sum_{Y'\in\mathcal Y}\pi_{\mathrm{ref}}(Y'\mid X)\exp\{\theta^\top\varphi(X,Y')/\beta\}}.
$$
Here $\mathcal Y$ is the fixed finite response set in Appendix~\ref{sec:rad-examples}. Restrict the head parameters to satisfy the reward bounds in Assumption~\ref{assump:realizability}, and retain the coverage conditions. Assume that the restricted class contains $\theta^*$ such that $\pi_{\theta^*}=\pi^*$ and
$$
r^*(Y^{(1)},X)-r^*(Y^{(2)},X)
=\theta^{*\top}\{\varphi(X,Y^{(1)})-\varphi(X,Y^{(2)})\}.
$$
For the initial verification, assume that the weights $w_i$ actually used in the estimator satisfy $|w_i|\le C_w$ almost surely and that
$$
\mathbb P(Z\ne\hat Z\mid X,Y^{(1)},Y^{(2)},\mathcal T)\le C\nu_n,
$$
almost surely, with fixed constants $C_w$ and $C$. This uniform bound with a deterministic constant strengthens Assumption~\ref{assump:aif_hf} and suffices for the local conditions to hold almost surely. Write $\mathbb E_{\mathrm{AI}}$ and $\mathbb E_{\mathrm{Hum}}$ for expectations under the two comparison laws, conditional on $\mathcal T$, and set
$$
\chi=\varphi(X,Y^{(1)})-\varphi(X,Y^{(2)}),\qquad
\Sigma=\mathbb E_{\mathrm{AI}}[\chi\chi^\top].
$$
Choose a common index $v\ge\max\{1,\operatorname{rank}(\Sigma)\}$ almost surely, enlarging it as needed to satisfy the other complexity bounds.

\paragraph{Example 3 (Deep ReLU DNN).}
Let $z_0(X,Y)$ be a bounded input encoding and
$$
z_\ell=\operatorname{ReLU}(A_\ell z_{\ell-1}+b_\ell),
\qquad \ell=1,\ldots,L,\qquad \varphi(X,Y)=z_L(X,Y).
$$
The depth $L$ and hidden-layer widths may be any fixed finite values. Freeze all hidden-layer weights and biases after independent pretraining, and train only the linear reward head $\theta$ by DB-DPO. Bounded inputs and finite fixed parameters ensure that $\varphi$ is bounded. The verification below uses the rank of $\Sigma$.

\paragraph{Example 4 (Multilayer, multihead Transformer).}
For the input-response sequence, let
$$
H_0=\operatorname{Embed}(X,Y),\qquad
H_\ell=\operatorname{TF}_\ell(H_{\ell-1}),\quad \ell=1,\ldots,L,
\qquad \varphi(X,Y)=\operatorname{Pool}(H_L)\in\mathbb R^d.
$$
Here $\operatorname{Embed}$ includes token and positional embeddings. Each block $\operatorname{TF}_\ell$ contains multihead softmax self-attention, a positionwise feedforward network with ReLU or GELU activations, residual connections, and layer normalization. The number of layers and heads may be any fixed finite values. The fixed map $\operatorname{Pool}$ selects the hidden state at a designated token position or forms a fixed linear combination of the final hidden states. A fixed finite context length, bounded embeddings, finite fixed parameters, and layer-normalization denominators bounded away from zero ensure that $\varphi$ is bounded. Freeze the embeddings, Transformer blocks, normalization parameters, and pooling after independent pretraining, and train only the linear reward head $\theta$ by DB-DPO. The same verification uses $\operatorname{rank}(\Sigma)\le d$.

For both architectures, the bounded-weight calculation depends on the representation through $\operatorname{rank}(\Sigma)$, which may depend on $d$, without an explicit parameter-count factor for the frozen network. These examples establish the local bounds for linear-head training. The finite-moment extension below gives an additional feature condition sufficient for unbounded weights.

\begin{proof}[Verification of Examples 3 and 4]
	The policy normalizer and the $X$-dependent term in \eqref{eq:rpi} cancel in pairwise reward differences. The factor $1/\beta$ in the policy parameterization also cancels the factor $\beta$ in \eqref{eq:rpi}. Thus, with $u=\theta-\theta^*$, we have
	$$
	f_\pi=u^\top\chi,\qquad \sigma_\pi^2=u^\top\Sigma u.
	$$
	Let $\Sigma^{\dagger/2}$ denote the positive semidefinite square root of the Moore--Penrose inverse of $\Sigma$, and put $\widetilde\chi=\Sigma^{\dagger/2}\chi$. If $u$ lies in the null space of $\Sigma$, then $\mathbb E_{\mathrm{AI}}(u^\top\chi)^2=0$. Hence $\chi$ lies in the range of $\Sigma$ almost surely under the AI law. The coverage conditions give $w\ge\epsilon^4$, so every AI-null set is also human-null, and the same range property holds under the human law. Consequently,
	$$
	u^\top\chi=(\Sigma^{1/2}u)^\top\widetilde\chi,
	\qquad
	\|\Sigma^{1/2}u\|_2=\sigma_\pi\le t
	\quad\text{for }\pi\in\Pi(t).
	$$
	For an i.i.d. sample from either source, let $c_i$ be the value of a common scalar multiplier $c$ at observation $i$. The multiplier may depend on $\mathcal T$ but not on the policy index or the Rademacher signs. Cauchy--Schwarz, Jensen's inequality, and independence of the signs give
	\begin{align*}
		\mathbb E\sup_{\pi\in\Pi(t)}
		\bigg|\frac1m\sum_{i=1}^m\varepsilon_i c_i f_\pi(U_i)\bigg|
		\le t\mathbb E\bigg\|\frac1m\sum_{i=1}^m
		\varepsilon_i c_i\widetilde\chi_i\bigg\|_2\le\frac t{\sqrt m}
		\big\{\mathbb E[c^2\|\widetilde\chi\|_2^2]\big\}^{1/2}.
	\end{align*}
	Under the AI law, take $c=1$ and use
	$$
	\mathbb E_{\mathrm{AI}}\|\widetilde\chi\|_2^2=\operatorname{tr}(\Sigma^\dagger\Sigma)=\operatorname{rank}(\Sigma)\le v.
	$$
	For each observation, the centered log-sigmoid defining $a_i(\pi)$ is a $1$-Lipschitz function of $f_\pi(U_i)$ that vanishes at zero. Rademacher contraction therefore gives \eqref{eq:local-rad-ddpo-ai}, with its universal constant absorbed into $C$. For the human correction, take $c_i=w_i(Z_i-\hat Z_i)$. The bounds on the actual weights and conditional disagreement, followed by change of measure using $w\ge\epsilon^4$, give
	\begin{align*}
		\mathbb E_{\mathrm{Hum}}[c^2\|\widetilde\chi\|_2^2]
		&\le C C_w^2\nu_n\mathbb E_{\mathrm{Hum}}\|\widetilde\chi\|_2^2\le C C_w^2\nu_n\epsilon^{-4}
		\mathbb E_{\mathrm{AI}}\|\widetilde\chi\|_2^2\le C C_w^2\epsilon^{-4}v\nu_n.
	\end{align*}
	Substituting this bound into the preceding inequality with $m=n$ proves \eqref{eq:local-rad-ddpo-human}. Both local bounds hold simultaneously for every $t>0$, and hence at all radii required by Corollary~\ref{cor:ddpo-local}. Under the remaining conditions of that corollary, both examples therefore satisfy \eqref{eq:subopt-dpo-local}.
	\hfill $\blacksquare$
\end{proof}

\paragraph{Verification with finite-moment weights.}
For Examples 3 and 4, the bounded-weight restriction can be replaced by the corresponding $\delta$-moment condition together with the weighted-feature condition below. Retain the coverage bounds, the uniform conditional disagreement bound with a deterministic constant, and the remaining conditions of the corresponding corollary, including its global complexity requirements. Set $\hat w=w$ when the weight is known, and let $p=\min\{2,\delta\}$ for either $\delta>2$ or $1<\delta<2$. With $\widetilde\chi=\Sigma^{\dagger/2}\chi$, assume
\begin{equation}\label{eq:local-feature-moment}
	\mathbb E_{\mathrm{Hum}}
	[\hat w^p\|\widetilde\chi\|_2^p]
	\le C M_\delta^{p/\delta}v^{p-1}
	\quad\text{almost surely}.
\end{equation}
The expectation is conditional on $\mathcal T$, and the constant is uniform in $n$, $N$, and $\nu_n$. A sufficient condition is $\|\widetilde\chi\|_2\le\Lambda$ almost surely under the human law, for a fixed $\Lambda$, with
$$
v\ge\max\{1,\operatorname{rank}(\Sigma),\Lambda^{p/(p-1)}\}.
$$
Indeed, the scalar weight moment gives $\mathbb E_{\mathrm{Hum}}\hat w^p\le M_\delta^{p/\delta}$, so the left-hand side of \eqref{eq:local-feature-moment} is at most $\Lambda^p M_\delta^{p/\delta}\le M_\delta^{p/\delta}v^{p-1}$. To verify the local conditions, use the preceding range-space argument and the following inequality, valid for $1<p\le2$ and arbitrary vectors $z_i$:
$$
\mathbb E_\varepsilon\bigg\|\sum_i\varepsilon_i z_i\bigg\|_2^p\le\bigg(\sum_i\|z_i\|_2^2\bigg)^{p/2}\le\sum_i\|z_i\|_2^p.
$$
Since $|Z-\hat Z|^p=\mathbb I\{Z\ne\hat Z\}$ and $\hat w$ and $\widetilde\chi$ are functions of the comparison tuple and $\mathcal T$, conditional disagreement and \eqref{eq:local-feature-moment} give, for every $t>0$,
\begin{align*}
	\mathbb E\sup_{\pi\in\Pi(t)}
	\bigg|\frac1n\sum_{i=1}^n\varepsilon_i b_i(\pi)\bigg|
	&\le t\mathbb E\bigg\|\frac1n\sum_{i=1}^n
	\varepsilon_i w_i(Z_i-\hat Z_i)\widetilde\chi_i\bigg\|_2\\
	&\le\frac{t}{n^{1-1/p}}
	\big\{\mathbb E_{\mathrm{Hum}}[
	\hat w^p|Z-\hat Z|^p\|\widetilde\chi\|_2^p]\big\}^{1/p}\le CtM_\delta^{1/\delta}\nu_n^{1/p}
	\bigg(\frac vn\bigg)^{1-1/p}.
\end{align*}
The AI local bound is unchanged. For $\delta>2$, take $p=2$ and absorb the fixed moment constant into $C$ to obtain \eqref{eq:local-rad-ddpo-human}. For $1<\delta<2$, take $p=\delta$ to obtain \eqref{eq:moment-local-ddpo-human}. Thus, under \eqref{eq:local-feature-moment} and the retained conditions, both frozen architectures satisfy the local complexity bounds required by Corollary~\ref{cor:ddpo-local} and Corollary~\ref{cor:ddpo-moment}(ii) in their respective moment regimes.

\subsection{Weighted Rademacher bounds and finite-moment extensions}
\label{sec:moment-weights}

This subsection connects the neural-network examples in Appendix~\ref{sec:rad-examples} to the weighted Rademacher conditions in Appendix~\ref{sec:rad-complexity}. We first state samplewise bounds for the unweighted DB-DPO class and the normalized DIPO correction class, then prove a multiplier inequality that yields the required weighted bounds for $\delta>2$. We then establish policy guarantees for $1<\delta<2$, including a local refinement for DB-DPO under additional local complexity conditions and a finite-step bound for DIPO. Throughout, we retain the sampling and norm conventions stated at the beginning of this appendix and condition on the independent nuisance-training sample $\mathcal T$. Moment and Rademacher constants are fixed, and $v\ge1$ may include the logarithmic factors allowed in Appendix~\ref{sec:rad-complexity}.

{\emergencystretch=3em
For DB-DPO, assume that every admissible deterministic comparison sample $U_1,\ldots,U_m$, with $1\le m\le\max(n,N)$, satisfies\par}

\begin{equation}\label{eq:moment-base-ddpo}
	\mathbb E_\varepsilon\sup_{\pi\in\Pi}
	\left|\frac1m\sum_{i=1}^m\varepsilon_i f_\pi(U_i)\right|
	\le C\sqrt{\frac vm}.
\end{equation}
Appendix~\ref{sec:rad-examples} verifies this bound for the stated norm-constrained classes. Bounded rewards provide a fixed envelope for $f_\pi$, and contraction yields the corresponding bound for the centered AI log-likelihood class. For DIPO, assume that, for almost every $X$, $\pi(\cdot\mid X),\pi_{\mathrm{ref}}(\cdot\mid X)\ll\pi_{\mathrm{Gen}}^{\mathrm{Hum}}(\cdot\mid X)$ for $\pi\in\Pi$. Define the true joint weights by
\begin{align*}
	w_\pi^{(1)}(U)=\frac{\pi(Y^{(1)}\mid X)\pi_{\mathrm{ref}}(Y^{(2)}\mid X)}{\pi_{\mathrm{Gen}}^{\mathrm{Hum}}(Y^{(1)}\mid X)\pi_{\mathrm{Gen}}^{\mathrm{Hum}}(Y^{(2)}\mid X)},\qquad
	w_\pi^{(2)}(X,Y^{(1)},Y^{(2)})=w_\pi^{(1)}(X,Y^{(2)},Y^{(1)}).
\end{align*}
The fitted weights $\hat w_\pi^{(a)}$ replace the generation probabilities with estimates fitted on $\mathcal T$ for $a=1,2$, with positive denominators on the evaluation support. Define the common envelope and impose the moment bound
\begin{equation}\label{eq:dipo-weight-moment}
	\begin{split}
		\bar w(U)=1\vee\sup_{\pi\in\Pi}\max_{a=1,2}
		\{w_\pi^{(a)}(U),\hat w_\pi^{(a)}(U)\},\quad
		\mathbb E_{\mathrm{Hum}}\bar w^\delta
		\le M_{\delta}<\infty\quad\text{almost surely}.
	\end{split}
\end{equation}
When the generation probabilities are known, set $\hat w_\pi^{(a)}=w_\pi^{(a)}$. The correction summand used in the estimator is
$$
B_i(\pi)=\frac12(Z_i-\hat Z_i)\{\hat w_\pi^{(1)}(U_i)-\hat w_\pi^{(2)}(U_i)\}.
$$
Assume that every admissible deterministic evaluation sample of size $1\le m\le n$, on the support where the denominators are positive and $\bar w$ is finite, satisfies
\begin{equation}\label{eq:moment-base-dipo}
	\mathbb E_\varepsilon\sup_{\pi\in\Pi}\bigg|\frac1m\sum_{i=1}^m\varepsilon_i\frac{B_i(\pi)}{\bar w(U_i)}\bigg|\le C\sqrt{\frac vm}.
\end{equation}
This bound concerns the normalized class formed with the actual fitted weights; its envelope satisfies $|B_i(\pi)/\bar w(U_i)|\le1$. Measure the weight-estimation error by
\begin{equation}\label{eq:moment-dipo-nuisance}
	e_{\delta}=\sup_{\pi\in\Pi}\max_{a=1,2}\|\hat w_\pi^{(a)}-w_\pi^{(a)}\|_{\delta,\mathrm{Hum}}.
\end{equation}
The common-envelope condition \eqref{eq:dipo-weight-moment} controls weight tails uniformly over policies and is stronger than a separate moment bound for each policy.

For the two moment regimes considered here, define
\begin{equation*}
	\rho_\delta(x)=
	\begin{cases}
		x^{1-1/\delta},&1<\delta<2,\\
		x^{1/2},&\delta>2,
	\end{cases}\qquad x>0.
\end{equation*}
Let $h_\pi$ be uniformly bounded and satisfy a samplewise bound of the form \eqref{eq:moment-base-ddpo} for every $1\le m\le n$. Let $V\ge0$ be a function of the comparison tuple $U$ and $\mathcal T$, with $\mathbb E V^\delta\le M$, and let $D\in\{-1,0,1\}$ satisfy $\mathbb P(D\ne0\mid U,\mathcal T)\le C\nu$ for $0<\nu\le1$. The function $h_\pi$ may depend on labels, provided its samplewise bound holds for every such labeled sample. For conditionally i.i.d. observations, write $V_i$, $D_i$, and $h_{\pi,i}$ for the corresponding values at observation $i$. The Rademacher signs are independent of these observations. Then
\begin{equation}\label{eq:moment-multiplier}
	\mathbb E\sup_{\pi\in\Pi}\bigg|\frac1n\sum_{i=1}^n\varepsilon_i V_iD_i h_{\pi,i}\bigg|\le C_\delta M^{1/\delta}\nu\rho_\delta\bigg(\frac{v}{n\nu}\bigg).
\end{equation}
To prove \eqref{eq:moment-multiplier}, introduce an auxiliary threshold $\tau>0$. The identity
$$
V\wedge\tau=\int_0^\tau\mathbb I\{V>s\}\,ds
$$
and the triangle inequality reduce the truncated Rademacher average to averages over the $K_s=\sum_{i=1}^n\mathbb I\{V_i>s,D_i\ne0\}$ selected observations. Conditional on the observations, the nonzero signs of $D_i$ can be absorbed into the independent Rademacher signs. The samplewise bound therefore gives
$$
\mathbb E_\varepsilon\sup_{\pi\in\Pi}
\bigg|\frac1n\sum_{i=1}^n
\varepsilon_i\mathbb I\{V_i>s\}D_i h_{\pi,i}\bigg|
\le\frac{C\sqrt{vK_s}}n,
$$
where the left-hand side is zero when $K_s=0$. Since $V$ depends only on $U$ and $\mathcal T$, the conditional bound on $D$ implies
$$
\mathbb E K_s
=n\mathbb P(V>s,D\ne0)
\le Cn\nu\,\mathbb P(V>s).
$$
Jensen's inequality then bounds the truncated part by
$$
C\sqrt{\frac{v\nu}{n}}\int_0^\tau\sqrt{\mathbb P(V>s)}\,ds.
$$
The envelope bound controls the remaining tail by
$$
C\mathbb E[V\mathbb I\{V>\tau,D\ne0\}]
\le C\nu\mathbb E[V\mathbb I\{V>\tau\}]
\le C\nu M\tau^{1-\delta}.
$$
Put $b=M^{1/\delta}$ and $x=v/(n\nu)$. If $M=0$, then $V=0$ almost surely and the result is immediate. Otherwise, for $x\ge1$, the envelope bound $C\nu\mathbb E V\le C\nu b$ proves \eqref{eq:moment-multiplier}. For $0<x<1$ and $1<\delta<2$, use
$$
\mathbb P(V>s)\le\min\{1,Ms^{-\delta}\}
$$
and choose $\tau=bx^{-1/\delta}$. Splitting the integral at $b$ gives
$$
\int_0^\tau\sqrt{\mathbb P(V>s)}\,ds
\le b+\frac{b^{\delta/2}\tau^{1-\delta/2}}{1-\delta/2}.
$$
Since $\sqrt{v\nu/n}=\nu\sqrt{x}$, both the truncated part and the tail are bounded by $C_\delta b\nu x^{1-1/\delta}$. For $\delta>2$, let $\tau\to\infty$. The tail vanishes, and
$$
\int_0^\infty\sqrt{\mathbb P(V>s)}\,ds
\le b+\frac b{\delta/2-1},
$$
which yields $C_\delta b\nu\sqrt{x}$. This proves \eqref{eq:moment-multiplier}. The threshold is used only in the proof; the estimator remains untruncated.

For $\delta>2$, this multiplier bound completes the verification of the ordinary complexity conditions in Appendix~\ref{sec:rad-complexity}. For DB-DPO, take $(V,D,h_\pi,\nu)=(\hat w,Z-\hat Z,f_\pi,\nu_n)$:the samplewise bound \eqref{eq:moment-base-ddpo}, the weight moment condition, and the conditional disagreement bound imply \eqref{eq:rad-ddpo-human}, while contraction and averaging imply \eqref{eq:rad-ddpo-ai}. For DIPO, take $(V,D,h_\pi,\nu)=(\bar w,1,B(\pi)/\bar w,1)$: \eqref{eq:dipo-weight-moment} and \eqref{eq:moment-base-dipo} imply \eqref{eq:rad-dipo-human}. Thus, the samplewise bounds verified for the neural-network examples in Appendix~\ref{sec:rad-examples} yield all three ordinary complexity conditions under the stated assumptions. For $1<\delta<2$, the same argument instead gives the slower weighted rates used in Corollaries~\ref{cor:ddpo-moment}(i) and~\ref{cor:dipo-moment}; the local DB-DPO refinement requires the additional conditions below.

We next state the additional local Rademacher conditions required by Corollary~\ref{cor:ddpo-moment}(ii). For $1<\delta<2$, retain $\Pi(t)$, $a_i(\pi)$, and $b_i(\pi)$ from Appendix~\ref{sec:local-ddpo}, and put
$$
\mu_{n,\delta}=\sqrt{\frac vN}+M_\delta^{1/\delta}\nu_n^{1/\delta}\bigg(\frac vn\bigg)^{1-1/\delta}.
$$
Conditional on $\mathcal T$, assume that the following bounds hold almost surely, simultaneously for all $t\ge\mu_{n,\delta}$:
\begin{align}
	\mathbb E\sup_{\pi\in\Pi(t)}
	\bigg|\frac1N\sum_{i=n+1}^{n+N}\varepsilon_i a_i(\pi)\bigg|
	&\le Ct\sqrt{\frac vN},\label{eq:moment-local-ddpo-ai}\\
	\mathbb E\sup_{\pi\in\Pi(t)}
	\bigg|\frac1n\sum_{i=1}^n\varepsilon_i b_i(\pi)\bigg|
	&\le CtM_\delta^{1/\delta}\nu_n^{1/\delta}
	\bigg(\frac vn\bigg)^{1-1/\delta}.
	\label{eq:moment-local-ddpo-human}
\end{align}
The expectations are over the evaluation observations and Rademacher signs, conditional on $\mathcal T$, and $C$ is uniform in $t$, $n$, $N$, and $\nu_n$. These are additional local conditions on the actual summands, including the correction class formed with fitted weights.

\paragraph{Theoretical guarantees for $1<\delta<2$.}
The main results allow unbounded weights with a finite moment of order $\delta>2$; \citet {cortes2010importance} study related moment-based guarantees for importance weighting. The following corollaries allow $1<\delta<2$, with slower human-sample rates. Expectations and norms remain conditional on $\mathcal T$, and $v$ retains its original meaning.

\begin{corollary}[DB-DPO with $1<\delta<2$]
	\label{cor:ddpo-moment}
	Retain the BT model, Assumptions~\ref{assump:realizability} and \ref{assump:aif_hf}, and the coverage conditions in Assumption~\ref{assump:coverage}, but replace its moment order $\delta>2$ by $1<\delta<2$. The empirical minimizer of the original, untruncated DB-DPO objective satisfies the following bounds.
	
	\textnormal{(i) Ordinary Rademacher complexity.}
	Replace the DB-DPO complexity conditions in Appendix~\ref{sec:rad-complexity} by the samplewise bound \eqref{eq:moment-base-ddpo}. Then
	\begin{equation}\label{eq:ddpo-moment-rate}
		\begin{split}
			&|\mathbb E_X[\mathbb E_{Y\sim\pi^*(\cdot\mid X)}r(X,Y)-\mathbb E_{Y\sim\hat\pi_{\mathrm{DB\text{-}DPO}}(\cdot\mid X)}r(X,Y)]|\\
			&=O_p\bigg[\frac1{\beta\epsilon}\bigg\{
			\bigg(\frac vN\bigg)^{1/4}
			+M_\delta^{1/(2\delta)}\nu_n^{1/(2\delta)}
			\bigg(\frac vn\bigg)^{(\delta-1)/(2\delta)}
			+\bigg(\nu_n\|\hat w-w\|_1\bigg)^{1/2}
			\bigg\}\bigg],
		\end{split}
	\end{equation}
	where $\|\hat w-w\|_1=\mathbb E_{\mathrm{Hum}}|\hat w-w|$.
	
	\textnormal{(ii) Local Rademacher complexity.}
	If the local bounds \eqref{eq:moment-local-ddpo-ai}--\eqref{eq:moment-local-ddpo-human} also hold, then
	\begin{equation}\label{eq:ddpo-moment-local-rate}
		\begin{split}
			&|\mathbb E_X[\mathbb E_{Y\sim\pi^*(\cdot\mid X)}r(X,Y)-\mathbb E_{Y\sim\hat\pi_{\mathrm{DB\text{-}DPO}}(\cdot\mid X)}r(X,Y)]|\\
			&=O_p\bigg[\frac1{\beta\epsilon}\bigg\{
			\sqrt{\frac vN}
			+M_\delta^{1/\delta}\nu_n^{1/\delta}
			\bigg(\frac vn\bigg)^{1-1/\delta}
			+\bigg(\nu_n\|\hat w-w\|_1\bigg)^{1/2}
			\bigg\}\bigg].
		\end{split}
	\end{equation}
	In either bound, if $\|\hat w-w\|_2<\infty$, the last term may be replaced by $\nu_n^{1/2}\|\hat w-w\|_2$.
\end{corollary}

The retained coverage conditions permit conversion from reward error to policy value. The local conditions improve the sampling terms, while the $L^1$ nuisance term remains valid without second moments. Appendix~\ref{sec:local-ddpo} verifies the local conditions for the two frozen neural representations under the stated assumptions, including \eqref{eq:local-feature-moment}.

\begin{corollary}[Finite-step DIPO with $1<\delta<2$]
	\label{cor:dipo-moment}
	Retain the conditions and fresh-evaluation protocol of Theorem~\ref{thm:regret_dipo}, replacing $\delta>2$ in Assumption~\ref{assump:weakcoverage} by $1<\delta<2$ and \eqref{eq:rad-dipo-human} by the normalized samplewise bound \eqref{eq:moment-base-dipo}. With the computational gap $\varepsilon_{\mathrm{opt},T}$ defined in \eqref{eq:dipo-computation-gap},
	\begin{equation}\label{eq:dipo-moment-rate}
		\begin{split}
			&|P(\pi^*)
			-P(\pi^{(T)})|\\
			&\quad\le\varepsilon_{\mathrm{opt},T}
			+O_p\bigg\{\beta K+\sqrt{\frac{\log\{2(T+1)\}}N}
			+M_\delta^{1/\delta}\bigg(\frac vn\bigg)^{1-1/\delta}
			+e_\delta\|\tilde g-g\|_{\delta/(\delta-1),\mathrm{Hum}}\bigg\}.
		\end{split}
	\end{equation}
	This bound requires no nuisance-consistency assumption. When the true generation weights are used, $e_\delta=0$, even if evaluator bias persists, and
	\begin{align*}
		&|P(\pi^*)
		-P(\pi^{(T)})|\\
		&\quad\le\varepsilon_{\mathrm{opt},T}
		+O_p\bigg\{\beta K+\sqrt{\frac{\log\{2(T+1)\}}N}
		+M_\delta^{1/\delta}\bigg(\frac vn\bigg)^{1-1/\delta}\bigg\}.
	\end{align*}
	If the fitted weights satisfy $e_\delta=o_p(1)$, the product term also vanishes without evaluator consistency, since $|\tilde g-g|\le1$.
\end{corollary}

Relative to Theorem~\ref{thm:regret_dipo}, only the human sampling term changes. The corollary concerns policy regret; fixed-policy expansion and efficiency retain their separate moment and nuisance conditions.

\begin{proof}[Proof of Corollary~\ref{cor:ddpo-moment}]
	\textnormal{(i) Ordinary Rademacher complexity.}
	Condition on $\mathcal T$. Under the uniform $O_p(\nu_n)$ disagreement condition, restrict first to a $\mathcal T$-measurable event on which the normalized disagreement bound has a fixed upper bound. Such events can have arbitrarily high probability for sufficiently large sample sizes. Apply \eqref{eq:moment-multiplier} with $V=\hat w$, $D=Z-\hat Z$, $h_\pi=f_\pi$, and $\nu=\nu_n$. Symmetrization and Markov's inequality, together with contraction for the AI summands, give
	$$
	\sup_{\pi\in\Pi}|\mathbb G_{n,N}(\pi)|=O_p\bigg\{\sqrt{\frac vN}+M_\delta^{1/\delta}\nu_n\rho_\delta\!\left(\frac{v}{n\nu_n}\right)\bigg\},
	$$
	where $\mathbb G_{n,N}$ is the centered process defined in Appendix~\ref{sec:local-ddpo}. Absolute continuity and integrability preserve the change-of-measure identity used in the proof of Theorem~\ref{thm:subopt-gap}. Boundedness of $f_\pi$ and the conditional disagreement bound yield, uniformly over $\pi\in\Pi$,
	$$
	|\mathbb E_{\mathrm{Hum}}[ (\hat w-w)(Z-\hat Z)f_\pi ]|\le O_p(\nu_n)\|\hat w-w\|_1.
	$$
	Likelihood curvature and empirical optimality therefore imply
	$$
	\sigma_{\hat\pi_{\mathrm{DB\text{-}DPO}}}^2=O_p\bigg\{\sqrt{\frac vN}+M_\delta^{1/\delta}\nu_n\rho_\delta\bigg(\frac{v}{n\nu_n}\bigg)+\nu_n\|\hat w-w\|_1\bigg\}.
	$$
	Taking square roots and applying the value bound $C\sigma_{\hat\pi_{\mathrm{DB\text{-}DPO}}}/(\beta\epsilon)$ from the proof of Theorem~\ref{thm:subopt-gap} proves \eqref{eq:ddpo-moment-rate}, since
	$$
	\nu_n\rho_\delta\bigg(\frac{v}{n\nu_n}\bigg)
	=\nu_n^{1/\delta}\bigg(\frac vn\bigg)^{1-1/\delta}.
	$$
	If $\|\hat w-w\|_2<\infty$, the coverage bound $w\ge\epsilon^4$ instead gives the population-bias bound $O_p(\nu_n\epsilon^{-2})\|\hat w-w\|_2\sigma_\pi$. Solving the resulting quadratic inequality and using $\nu_n\le1$ gives the alternative nuisance term, with fixed coverage constants.
	
	\textnormal{(ii) Local Rademacher complexity.}
	Symmetrization and \eqref{eq:moment-local-ddpo-ai}--\eqref{eq:moment-local-ddpo-human} give
	$$
	\mathbb E\big[\sup_{\pi\in\Pi(t)}|\mathbb G_{n,N}(\pi)|\mid\mathcal T\big]\le Ct\mu_{n,\delta},\qquad t\ge\mu_{n,\delta}.
	$$
	Write $\sigma=\sigma_{\hat\pi_{\mathrm{DB\text{-}DPO}}}$ and $b_n=\nu_n\|\hat w-w\|_1$. The same curvature and bias bounds give
	$$
	c\sigma^2\le|\mathbb G_{n,N}(\hat\pi_{\mathrm{DB\text{-}DPO}})|+O_p(1)b_n.
	$$
	For each $\eta>0$, the $\mathcal T$-measurable $O_p(1)$ coefficient is bounded by a constant $D_\eta$ on a training event of probability at least $1-\eta$, for all sufficiently large sample sizes. On this event, set $R=A(\mu_{n,\delta}+\sqrt{b_n})$, where $A\ge\max\{1,\sqrt{2D_\eta/c}\}$. The radius $R$ is fixed conditional on $\mathcal T$, and $\sigma>R$ implies that the bias term is at most $c\sigma^2/2$. Partitioning into shells $2^jR<\sigma\le2^{j+1}R$ and applying Markov's inequality gives
	$$
	\mathbb P(\sigma>R\mid\mathcal T)
	\le C\sum_{j\ge0}
	\frac{2^{j+1}R\mu_{n,\delta}}{2^{2j}R^2}
	\le\frac{C\mu_{n,\delta}}R\le\frac CA.
	$$
	All radii used exceed $\mu_{n,\delta}$. Removing the conditioning gives
	$$
	\mathbb P\{\sigma>A(\mu_{n,\delta}+\sqrt{b_n})\}
	\le\eta+C/A
	$$
	for sufficiently large sample sizes. Taking $\eta$ arbitrarily small and then $A$ sufficiently large yields $\sigma=O_p(\mu_{n,\delta}+\sqrt{b_n})$. The same value bound proves \eqref{eq:ddpo-moment-local-rate}. If $\|\hat w-w\|_2<\infty$, replace the bias term by $O_p(1)d_n\sigma$, where $d_n=\nu_n\epsilon^{-2}\|\hat w-w\|_2$. Repeating the argument with $R=A(\mu_{n,\delta}+d_n)$ gives $\sigma=O_p(\mu_{n,\delta}+d_n)$ and the stated alternative, using $\nu_n\le1$ and fixed coverage constants.
	\hfill $\blacksquare$
\end{proof}

\begin{proof}[Proof of Corollary~\ref{cor:dipo-moment}]
	Condition on $\mathcal T$ and apply \eqref{eq:moment-multiplier} with $V=\bar w$, $D=1$, $\nu=1$, and $h_\pi=B(\pi)/\bar w$. For $1<\delta<2$,
	$$
	\mathbb E\sup_{\pi\in\Pi}\bigg|
	\frac1n\sum_{i=1}^n\varepsilon_iB_i(\pi)\bigg|
	\le C_\delta M_\delta^{1/\delta}
	\bigg(\frac vn\bigg)^{1-1/\delta}.
	$$
	Symmetrization and Markov's inequality imply
	$$
	\sup_{\pi\in\Pi}\bigg|
	\frac1n\sum_{i=1}^nB_i(\pi)-\mathbb E_{\mathrm{Hum}}B(\pi)\bigg|
	=O_p\bigg\{M_\delta^{1/\delta}\bigg(\frac vn\bigg)^{1-1/\delta}\bigg\}.
	$$
	Since $\delta>1$, the weights are integrable. Pair exchangeability, the comparison convention for $g$ and $\tilde g$, and change of measure therefore give the identity used in the proof of Theorem~\ref{thm:regret_dipo}:
	$$
	\overline P_{\mathrm{DM}}(\pi)+\mathbb E_{\mathrm{Hum}}B(\pi)-P(\pi)
	=\mathbb E_{\mathrm{Hum}}[(\tilde g-g)
	\{w_\pi^{(1)}-\hat w_\pi^{(1)}\}].
	$$
	H\"older's inequality with exponents $\delta$ and $\delta/(\delta-1)$ bounds the absolute bias uniformly by
	$e_\delta\|\tilde g-g\|_{\delta/(\delta-1),\mathrm{Hum}}$.
	Thus the uniform human-sample error $e_{\mathrm H}$ defined in that proof satisfies
	$$
	e_{\mathrm H}=O_p\bigg\{
	M_\delta^{1/\delta}\bigg(\frac vn\bigg)^{1-1/\delta}
	+e_\delta\|\tilde g-g\|_{\delta/(\delta-1),\mathrm{Hum}}\bigg\}.
	$$
	The direct scores remain in $[0,1]$, so the fresh-batch concentration bound \eqref{eq:dipo-fresh-concentration} is unchanged. Substituting these bounds into the deterministic comparison \eqref{eq:dipo-trajectory-comparison} proves \eqref{eq:dipo-moment-rate}. The proof uses no nuisance-consistency assumption. The remaining conclusions follow from $e_\delta=0$ for true weights and $\|\tilde g-g\|_{\delta/(\delta-1),\mathrm{Hum}}\le1$.
\end{proof}

For $1<\delta<2$, finite second moments are not guaranteed, so these corollaries do not extend the fixed-policy asymptotic expansion or efficiency claims without their separate assumptions. They apply to untruncated weights; clipping at a fixed threshold introduces an additional bias term that must be accounted for separately.

\subsection{Computational errors}
\label{sec:computational-errors}

This subsection quantifies errors from Monte Carlo evaluation, finite optimization, and loss approximation. The Monte Carlo result concerns fixed-policy evaluation under the stated $\delta>2$ assumptions. The optimization extensions apply to both moment regimes under the conditions of the corresponding policy bounds.

\subsubsection{Monte Carlo evaluation}
\label{sec:dipo-mc}

Fix $\pi$ and $\pi_{\mathrm{ref}}$ as in
\eqref{eq:dipo-probability}. When exact evaluator probabilities are unavailable, draw $m$ conditionally independent AI labels for each direct comparison and set
$$
\bar Z_i^{(a,m)}=\frac1m\sum_{j=1}^m\widehat Z_{i,j}^{(a)},
\qquad
\widehat Z_{i,j}^{(a)}\mid X_i,Y_i,Y_i^{(a)}\sim\operatorname{Bernoulli}\{\tilde g(Y_i,Y_i^{(a)}\mid X_i)\}.
$$
Replacing the corresponding probability in \eqref{eq:PDM} by this average defines its Monte Carlo approximation. The human correction is approximated analogously below.

For each human pair, obtain fresh AI labels $\widehat Z_{i,j}\sim\operatorname{Bernoulli}\{\tilde g(Y_i^{(1)},Y_i^{(2)}\mid X_i)\}$ and let $\bar Z_i^{(m)}=m^{-1}\sum_{j=1}^m\widehat Z_{i,j}$. Conditionally on the fitted nuisances, all evaluation prompts and responses, and the human labels, assume all simulation calls used in these averages and in $\bar Z_i^{(a,m)}$ are independent, with the stated Bernoulli means. The Monte Carlo version is
\begin{equation*}
	\begin{aligned}
		\hat P_{\mathrm{DIPO}}^{(m)}(\pi)
		&=\frac1{2N}\sum_{i=n+1}^{n+N}\sum_{a=1}^2\bar Z_i^{(a,m)}-\frac1{2n}\sum_{i=1}^n\{\bar Z_i^{(m)}-Z_i\}\\
		&\times\frac{\pi(Y_i^{(1)}\mid X_i)\pi_{\mathrm{ref}}(Y_i^{(2)}\mid X_i)-\pi(Y_i^{(2)}\mid X_i)\pi_{\mathrm{ref}}(Y_i^{(1)}\mid X_i)}{\hat\pi_{\mathrm{Gen}}^{\mathrm{Hum}}(Y_i^{(1)}\mid X_i)\hat\pi_{\mathrm{Gen}}^{\mathrm{Hum}}(Y_i^{(2)}\mid X_i)}.
	\end{aligned}
\end{equation*}
Exact evaluator probabilities are used in \eqref{eq:dipo-probability}; they are not identified with the binary labels $\hat Z_i$. The simulation size $m=m_{n,N}$ may grow; its effect on fixed-policy evaluation is quantified in \eqref{eq:dipo-mc-error} below.

For the Monte Carlo version, the exact decomposition is
\begin{equation*}
	\hat P_{\mathrm{DIPO}}^{(m)}(\pi)=\hat P_{\mathrm{DIPO}}^{\mathrm{prob}}(\pi)+R_{\mathrm{MC}}(\pi),
\end{equation*}
where
\begin{equation*}
	\begin{split}
		R_{\mathrm{MC}}(\pi)
		=&\frac1{2N}\sum_{i=n+1}^{n+N}\sum_{a=1}^2\{\bar Z_i^{(a,m)}-\tilde g(Y_i,Y_i^{(a)}\mid X_i)\}-\frac1{2n}\sum_{i=1}^n\{\bar Z_i^{(m)}-\tilde g(Y_i^{(1)},Y_i^{(2)}\mid X_i)\}\\
		&\times\frac{\pi(Y_i^{(1)}\mid X_i)\pi_{\mathrm{ref}}(Y_i^{(2)}\mid X_i)-\pi(Y_i^{(2)}\mid X_i)\pi_{\mathrm{ref}}(Y_i^{(1)}\mid X_i)}{\hat\pi_{\mathrm{Gen}}^{\mathrm{Hum}}(Y_i^{(1)}\mid X_i)\hat\pi_{\mathrm{Gen}}^{\mathrm{Hum}}(Y_i^{(2)}\mid X_i)}.
	\end{split}
\end{equation*}
Let $\mathcal G$ contain the fitted nuisances, evaluation prompts and responses, and human labels, but not the simulation calls. The conditional independence above and Assumption~\ref{assump:weakcoverage} give
\begin{equation}\label{eq:dipo-mc-error}
	\mathbb E(R_{\mathrm{MC}}\mid\mathcal G)=0,
	\qquad
	\mathbb E R_{\mathrm{MC}}^2\le\frac1{8Nm}+\frac{M_{\delta}^{2/\delta}}{4nm}.
\end{equation}
Thus, for $N>n$, $m\to\infty$ suffices for $R_{\mathrm{MC}}=o_{L^2}(n^{-1/2})$. A fixed simulation size generally leaves a leading human-correction noise term. The proof is given below.

\paragraph{Extension of Corollary~\ref{cor:efficiency_comparison}.}
Under all the conditions of Corollary~\ref{cor:efficiency_comparison}, its conclusions also hold for $\hat P_{\mathrm{DIPO}}^{(m)}(\pi)$ under the conditional simulation independence above and $m=m_{n,N}\to\infty$. In particular, its regularity requirement applies to the estimator under consideration.

\begin{proof}
	For \eqref{eq:dipo-mc-error}, condition on $\mathcal G$. Each simulation average has variance $\tilde g(1-\tilde g)/m\le1/(4m)$. The simulation groups are conditionally independent, so
	$$
	\operatorname{Var}(R_{\mathrm{MC}}\mid\mathcal G)\le\frac1{8Nm}+\frac1{16n^2m}\sum_{i=1}^n\{\hat w_\pi^{(1)}(U_i)-\hat w_\pi^{(2)}(U_i)\}^2.
	$$
	Here the joint weights are those in \eqref{eq:dipo-weight-moment}. Since $|\hat w_\pi^{(1)}-\hat w_\pi^{(2)}|/2\le\bar w$, the conditional moment bound gives
	$\mathbb E_{\mathrm{Hum}} \{\hat w_\pi^{(1)}-\hat w_\pi^{(2)}\}^2/4\le M_{\delta}^{2/\delta}$. Taking the expectation proves \eqref{eq:dipo-mc-error}, hence $R_{\mathrm{MC}}=o_{L^2}(n^{-1/2})$ when $N>n$ and $m\to\infty$. Because $\hat P_{\mathrm{DIPO}}^{\mathrm{prob}}(\pi)$ is $\mathcal G$-measurable and the simulation error has conditional mean zero, there is also the exact identity
	$$
	\mathrm{MSE}(\hat P_{\mathrm{DIPO}}^{(m)}(\pi))=\mathrm{MSE}(\hat P_{\mathrm{DIPO}}^{\mathrm{prob}}(\pi))+\mathbb E R_{\mathrm{MC}}^2.
	$$
	The additional term is $o(n^{-1})$ under the stated simulation condition. Therefore, the expansion, strict MSE comparison, and fixed-policy efficiency conclusion carry over to $\hat P_{\mathrm{DIPO}}^{(m)}(\pi)$, without identifying binary labels with probabilities.
\end{proof}

\subsubsection{Finite optimization for DB-DPO}
The displayed bounds concern the empirical minimizer. If the returned policy $\pi^{(T)}\in\Pi$ satisfies
$$
\hat L_{\mathrm{DB\text{-}DPO}}(\pi^{(T)};\pi_{\mathrm{ref}})\le\inf_{\pi\in\Pi}\hat L_{\mathrm{DB\text{-}DPO}}(\pi;\pi_{\mathrm{ref}})+\varepsilon_{\mathrm{opt}}
$$
for a deterministic tolerance $\varepsilon_{\mathrm{opt}}\ge0$, Theorem~\ref{thm:subopt-gap} and Corollaries~\ref{cor:ddpo-local} and~\ref{cor:ddpo-moment} hold for that output with the additional term $O_p\{\sqrt{\varepsilon_{\mathrm{opt}}}/(\beta\epsilon)\}$. The proof follows below. We do not assert a convergence rate for the optimization tolerance.

\paragraph{Proof of the finite-optimization extension for DB-DPO.}
Write $\sigma=\sigma_{\pi^{(T)}}$ and use the notation of the corresponding local proof. For an $\varepsilon_{\mathrm{opt}}$-minimizer, empirical optimality adds $\varepsilon_{\mathrm{opt}}$ to each curvature inequality. The global cases follow by solving the resulting quadratic inequalities. For the local $L^2$ case, on the same training event as in the proof of Corollary~\ref{cor:ddpo-local}, use $R=A(\mu_n+d_n+\sqrt{\varepsilon_{\mathrm{opt}}})$ with  $A\ge\max\{1,4D_\eta/c,2/\sqrt c\}$. Then $\sigma>R$ implies $D_\eta d_n\sigma+\varepsilon_{\mathrm{opt}}\le c\sigma^2/2$, and the same peeling argument gives $\sigma=O_p(\mu_n+d_n+\sqrt{\varepsilon_{\mathrm{opt}}})$. For Corollary~\ref{cor:ddpo-moment}(ii), the $L^1$ case uses $R=A(\mu_{n,\delta}+\sqrt{b_n}+\sqrt{\varepsilon_{\mathrm{opt}}})$ and $A\ge\max\{1,\sqrt{4D_\eta/c},2/\sqrt c\}$; the optional $L^2$ case uses the first radius with $\mu_n$ replaced by $\mu_{n,\delta}$. These radii remain measurable with respect to the nuisance-training sample because the tolerance is deterministic. The unchanged value-conversion inequality proves the stated additional term in all cases.

\subsubsection{Finite optimization and loss approximation for DIPO}
Recall $\overline J$ from \eqref{eq:dipo}. Conditional on nuisance fitting and the human sample, this is a fixed objective over $\Pi$; it averages over new prompts and responses but retains the empirical human correction. For each $t=0,\ldots,T$, define the untruncated corrected score
\begin{equation*}
	\widehat J_t=\hat P_{\mathrm{DM},t}(\pi^{(t)})-\widehat{\mathrm{Bias}}(\pi^{(t)})-\beta D_{\mathrm{KL}}(\pi^{(t)}\|\pi_{\mathrm{ref}}),
\end{equation*}
where $\hat P_{\mathrm{DM},t}$ is \eqref{eq:PDM} evaluated on a fresh batch generated after $\pi^{(t)}$ is determined. In Algorithm~\ref{alg:dipo}, the batches used for updates evaluate $\pi^{(0)},\ldots,\pi^{(T-1)}$; an additional fresh batch evaluates $\pi^{(T)}$. All evaluations use the same fitted nuisances and human correction. Define the computational gap
\begin{equation*}
	\begin{split}
		\varepsilon_{\mathrm{opt},T}=\bigg\{\sup_{\pi\in\Pi}\overline J(\pi)-\max_{0\le t\le T}\overline J(\pi^{(t)})\bigg\}+\bigg\{\max_{0\le t\le T}\widehat J_t-\widehat J_T\bigg\}.
	\end{split}
\end{equation*}
Both terms are nonnegative. The first is the optimization error of the best visited policy for the conditional mean objective; the second is the final iterate's shortfall in recorded score. This gap is a theoretical remainder, not an observable certificate of global optimality. Its definition allows arbitrary adaptive updates and does not assume that SGD makes it small.

With $e_{\mathrm H}$ and $\Delta_{N,T}$ defined in the proof of Theorem~\ref{thm:regret_dipo}, the deterministic comparison is
\begin{equation}\label{eq:dipo-optimization-bound}
	|P(\pi^*)-P(\pi^{(T)})|\le\beta K+2e_{\mathrm H}+2\Delta_{N,T}+\varepsilon_{\mathrm{opt},T}.
\end{equation}
Indeed, replace $\overline J(\pi^{(t)})$ by $\widehat J_t$ at the best visited policy and at the final policy, incurring at most $2\Delta_{N,T}$. The first term of \eqref{eq:dipo-computation-gap} then bounds the remaining gap to $\sup_{\pi\in\Pi}\overline J(\pi)$. Replacing $\overline J$ by the population objective at $\pi^*$ and $\pi^{(T)}$ costs at most $2e_{\mathrm H}$, and the KL conversion costs at most $\beta K$.

If the output is instead a checkpoint maximizing $\widehat J_t$, the second term in \eqref{eq:dipo-computation-gap} is zero. The simultaneous concentration event already covers this data-dependent selection, so \eqref{eq:dipo-optimization-bound} holds for that output with only the first gap term. A fixed iteration budget alone does not control this first term. In particular, small gradients or unbiased stochastic gradients do not establish global optimization of a neural policy class.

For approximate recorded scores $\widetilde J_t$, suppose
$\max_{0\le t\le T}|\widetilde J_t-\widehat J_t|\le a$.
Define $\widetilde\varepsilon_{\mathrm{opt},T}$ by retaining the first term of \eqref{eq:dipo-computation-gap}, with the same ideal objective $\overline J$, and replacing only the recorded scores in its second term by $\widetilde J_t$. Then
$$
\max_t\widehat J_t-\widehat J_T\le\max_t\widetilde J_t-\widetilde J_T+2a,
\qquad
\varepsilon_{\mathrm{opt},T}\le\widetilde\varepsilon_{\mathrm{opt},T}+2a.
$$
Thus Theorem~\ref{thm:regret_dipo} and Corollary~\ref{cor:dipo-moment} hold with the added term
$\widetilde\varepsilon_{\mathrm{opt},T}+2a$ in place of $\varepsilon_{\mathrm{opt},T}$. This algebraic comparison permits random $a$ and applies to clipping, approximate KL evaluation, and other score approximations when the displayed bound is available. Changes to the updates themselves are reflected in the unchanged first gap term. The score-function surrogate is used to compute gradients; its numerical values are not preference scores and are not assumed to satisfy this approximation bound.

The term $\sqrt{\log\{2(T+1)\}/N}$ controls fresh evaluations along the trajectory. It does not establish a uniform sampled-class bound over $\Pi$, and $N$ counts observations per batch rather than all $(T+1)N$ AI observations. If only target responses are refreshed on reused prompts, conditional concentration instead centers at the direct term integrated over those responses; an additional uniform prompt/reference-sample bound would then be required. The Monte Carlo calculation in Appendix~\ref{sec:dipo-mc} concerns evaluator-label simulation for a fixed policy and is separate from this trajectory argument.

\section{Implementation Details of DIPO and Extensions}

\subsection{Implementation Details}
\label{sec:dipo-implementation}

Discrete response sampling motivates a score-function estimator of the policy gradient. It does not require changing the preference-probability estimator \eqref{eq:est_dipo}. Write $\pi=\pi_\theta$ for a differentiable policy with common support, and condition on independently fitted nuisance functions, including the fixed evaluator $\tilde g$. Assume the relevant derivatives are integrable and differentiation may be interchanged with expectation. Put
$$
\overline P_{\mathrm{DM}}(\pi_\theta)=\mathbb E_{X,\,Y\sim\pi_\theta(\cdot\mid X),Y'\sim\pi_{\mathrm{ref}}(\cdot\mid X)}\tilde g(Y,Y'\mid X).
$$
Let $\mathcal F_{t-1}$ contain the fitted nuisances, the human data, and the optimization history before iteration $t$. For the fresh prompts and responses generated at that iteration of Algorithm~\ref{alg:dipo}, define
\begin{equation*}
	\hat P_{\mathrm{DM}}^{*,t}(\pi_\theta)=\frac1{2N}\sum_{i=n+1}^{n+N}\log\pi_\theta(Y_i\mid X_i)\sum_{a=1}^2\tilde g(Y_i,Y_i^{(a)}\mid X_i),
\end{equation*}
where $Y_i\sim\pi_{\theta_{t-1}}(\cdot\mid X_i)$ and $Y_i^{(a)}\sim\pi_{\mathrm{ref}}(\cdot\mid X_i)$. Hold the sampled responses and evaluator scores fixed when differentiating. The conditional gradient identity is
\begin{equation}\label{eq:dipo-score-identity}
	\mathbb E\bigg[\nabla_\theta\hat P_{\mathrm{DM}}^{*,t}(\pi_\theta)|_{\theta=\theta_{t-1}}\bigg|\mathcal F_{t-1}\bigg]=\nabla_\theta\overline P_{\mathrm{DM}}(\pi_\theta)|_{\theta=\theta_{t-1}}.
\end{equation}
Indeed, the conditional expectation of each sampled score term is
$$
\mathbb E_{X,Y'}\sum_y\pi_\theta(y\mid X)\nabla_\theta\log\pi_\theta(y\mid X)\tilde g(y,Y'\mid X)=\mathbb E_{X,Y'}\sum_y\nabla_\theta\pi_\theta(y\mid X)\tilde g(y,Y'\mid X),
$$
evaluated at $\theta=\theta_{t-1}$. Interchanging differentiation and expectation proves \eqref{eq:dipo-score-identity}. Unbiased Monte Carlo evaluator scores give the same identity by iterated expectation.

The score-function update uses
$$
\hat P_{\mathrm{DIPO}}^{*,t}(\pi)=\hat P_{\mathrm{DM}}^{*,t}(\pi)-\widehat{\mathrm{Bias}}_{[c_{\min},c_{\max}]}(\pi),
$$
where the subscript denotes the clipped ratios in Algorithm~\ref{alg:dipo}; with $c_{\min}=0$ and $c_{\max}=\infty$ this is the original $\widehat{\mathrm{Bias}}(\pi)$. The correction weights retain their policy dependence during differentiation, and we differentiate the KL term as well. In the untruncated case, with exact correction and KL gradients, the conditional mean gradient direction is the gradient of $\overline P_{\mathrm{DM}}(\pi)-\widehat{\mathrm{Bias}}(\pi)-\beta D_{\mathrm{KL}}(\pi\|\pi_{\mathrm{ref}})$.

This identity concerns gradients at the policy generating the current responses. It does not identify surrogate values with preference probabilities or transfer their asymptotic distributions. With fresh prompts and responses, untruncated weights, and exact correction and KL gradients, the conditional mean update is the gradient of $\overline J$ in \eqref{eq:dipo}, the objective used in the computational gap \eqref{eq:dipo-computation-gap}. The recorded scores in \eqref{eq:dipo-checkpoint-score} use the preference values in \eqref{eq:PDM}. Reusing responses from an earlier policy requires an appropriate importance correction; the unweighted gradient identity applies only at their generating policy.

Theorem~\ref{thm:mse} and Corollary~\ref{cor:efficiency_comparison} retain their fixed-policy interpretations and their separate nuisance and Monte Carlo conditions. Theorem~\ref{thm:regret_dipo} instead controls the finite-step policy using fresh evaluations, uniform control of the reused human correction, and the explicit computational gap. The final iterate must be evaluated on a batch drawn after it is formed. We do not assert that the neural implementation makes this gap negligible. Clipping and approximate criterion values require the additional control described in Appendix~\ref{sec:computational-errors}.

\section{Additional Numerical Experiment Details}

\subsection{Dataset Statistics for the Additional Studies}\label{sec:previous-data}
Table~\ref{tab:dataset_stats} presents the number of training and testing samples for both the supervised fine-tuning and preference datasets across all three experimental tasks. 

To compare the offline methods (DPO, Sampled IPO, DB-DPO) and the online methods (IPO, DIPO, DIPO+) on an equal footing, every method within a task is trained on the same set of preference prompts. The online methods are considerably more expensive, so we fix the preference set at a size that keeps online training feasible and use it for all methods. For sentiment, $\mathcal{D}_{\mathrm{Hum}}$ is built from human-written IMDb reviews in the training and unlabeled splits. Reviews that open with the same four words are grouped. Within each group, we rank the reviews by their gold reward scores and iteratively pair the highest- and lowest-scoring remaining reviews until the difference in gold log-odds between the pair falls below $2$. This gives $5{,}277$ disjoint pairs over $2{,}238$ distinct prompts, from which we draw $5{,}000$ at random. For summarization, the training split of the \texttt{summarize-from-feedback} comparison data contains $92{,}858$ human comparisons over $14{,}767$ distinct TL;DR posts. We keep one comparison per post, drawn at random, and remove three posts whose two candidate summaries are identical, leaving $14{,}764$ prompts. For dialogue, we start from the $43{,}835$ conversations in the \texttt{helpful-base} training split of the Anthropic HH data and keep the $12{,}797$ single-turn ones. We remove pairs in which a response contains a stray speaker tag, is empty, or coincides with the other response. We keep one pair per prompt and discard a small number of prompts without a usable pair of SFT samples, leaving $12{,}478$ prompts. 

\begin{table}[ht]
	\centering
	\begin{tabular}{ccccccc}
		\hline
		& \multicolumn{2}{c}{Sentiment} & \multicolumn{2}{c}{Summarization} & \multicolumn{2}{c}{Dialogue} \\
		& SFT         & Preference      & SFT           & Preference        & SFT        & Preference      \\ \hline
		Training (DB-DPO) & 25000       & 5000           & 107218        & 14764             & 49269      & 12478           \\
		Training (DIPO) & 25000       & 5000           & 107218        & 14764             & 49269      & 12478           \\
		Testing  & -           & 5000            & -             & 5888              & -          & 2602            \\ \hline
	\end{tabular}
	\caption{Dataset statistics for the three experimental tasks.}
	\label{tab:dataset_stats}
\end{table}

\subsection{Training details}\label{sec:training-details}
\paragraph{Controlled judge bias.}
To precisely control the source and magnitude of label bias, we construct each AI judge by augmenting a weak base model with a feature-gated preference rule. When two responses differ beyond a pre-calibrated threshold along a target feature, that feature dictates the label, whereas otherwise the base model's native preference prevails. This fixed rule applies to both offline labeling and online scoring. For sentiment generation (\texttt{twitter-roberta-base-sentiment-latest}), the judge selects the first-presented review when predicted positive probabilities differ by less than $0.5$. For summarization (\texttt{Qwen3-1.7B}), it favors the longer summary when word counts differ by more than $4.5$ words (half the standard deviation of length differences across SFT summary pairs). For dialogue (\texttt{Qwen3-1.7B}), it favors the response with higher mean per-token log-likelihood under itself when they differ by more than $0.95$ nats/token (two-thirds the standard deviation across SFT response pairs). These judges agree with the gold reward model on $63.0\%$ of $\mathcal{D}_{\mathrm{AI}}$ pairs for sentiment generation, and with the human proxy on $63.1\%$ and $67.1\%$ of human-labeled pairs for summarization and dialogue, respectively.

\paragraph{Density-ratio estimation.}
We estimate the density ratio with a probabilistic classifier. For every prompt in $\mathcal{D}_{\mathrm{Hum}}$ we additionally sample two responses from the SFT policy, and train a classifier to separate these SFT responses from the two dataset responses given the prompt. The classifier is \texttt{Qwen2.5-1.5B-Instruct} with a sequence-classification head and LoRA adapters of rank $8$, trained for one epoch with learning rate $2\times10^{-4}$ and weight decay $0.01$. Its logit $z(x,y)$ estimates $\log\{\pi_{\mathrm{Gen}}^{\mathrm{AI}}(y\mid x)/\pi_{\mathrm{Gen}}^{\mathrm{Hum}}(y\mid x)\}$, and the weight of a pair is $\log\widehat w(x,y^{(1)},y^{(2)})=\{z(x,y^{(1)})+z(x,y^{(2)})\}/T$, where $T$ is a calibration temperature.

The estimate is cross-fitted. The prompts of $\mathcal{D}_{\mathrm{Hum}}$ are split at random into two halves. A classifier trained on one half is used only to score the pairs of the other half, so that no pair is weighted by a model that saw it during training. Within each training half, $10\%$ of the prompts are held out to fit the temperature $T$.

\paragraph{Supervised fine-tuning.}
For sentiment, we fine-tune \texttt{gpt-neo-125m} \citep{gpt_neo} for three epochs on the $25{,}000$ IMDb training reviews. For summarization, we fine-tune \texttt{Qwen2.5-1.5B} \citep{qwen25} for three epochs on \texttt{summarize\_\allowbreak from\_\allowbreak feedback\_\allowbreak tldr\_\allowbreak 3\_\allowbreak filtered} \citep{summarization_sft_data}, after removing the samples that overlap with the preference dataset. For dialogue, we extract the single-turn conversations of the Anthropic HH training data and fine-tune \texttt{Qwen2.5-1.5B} for three epochs on their chosen responses. In each task, the resulting SFT model initializes every policy and also serves as the reference policy $\pi_{\mathrm{ref}}$.

\paragraph{Policy training.}
All methods are implemented with the Hugging Face TRL library and use its default hyperparameters unless stated otherwise. Every method is trained for one epoch over the preference set with an effective batch size of $32$. For the online methods, each update samples one response per prompt from the current policy, with at most $128$ new tokens for summarization and $512$ for sentiment and dialogue. A sampled response that reaches the token limit without terminating is excluded from the loss. The coefficient of the KL regularizer is set to $\beta=0.04$. For DB-DPO, the estimated density ratio $\widehat w$ is clipped to $[0.8,1.25]$ and rescaled to have mean one over $\mathcal{D}_{\mathrm{Hum}}$. For DIPO and DIPO+, the clipbound is $[0.8,1.25]$.

\paragraph{Hardware.}
Each run uses a single GPU: an NVIDIA A40 or A100 for sentiment, and an NVIDIA H100 for summarization and dialogue.

\subsection{Prompts for the Additional Studies}\label{sec:previous-prompts}
In this section, we present the prompts utilized to assess win rates for summarization and dialogue tasks. To mitigate potential position bias, the order of candidate summaries or responses is randomized for each evaluation instance. For evaluations employing \texttt{claude-5.5-haiku} and \texttt{gpt-6-luna} as the judge, we adopt the prompt templates proposed by \cite{rafailov2023direct, xu2025doubly}, which are detailed in Table~\ref{tab:summarization_prompt} and Table~\ref{tab:dialogue_prompt}. While utilizing models such as \texttt{Qwen3-1.7B}, we instruct the models to provide the answer directly without providing an explanation. This format facilitates automatic parsing of the evaluation outputs. The corresponding prompt templates are provided in Table~\ref{tab:summarization_prompt_qwen} and Table~\ref{tab:dialogue_prompt_qwen}.

\begin{table}[ht]
	\centering
	\begin{mdframed}
		Which of the following summaries does a better job of summarizing the most important points in the given forum post, without including unimportant or irrelevant details? A good summary is both precise and concise.
		
		Post: \\
		\{post\}
		
		Summary A: \\
		\{summary\_0\}
		
		Summary B: \\
		\{summary\_1\}
		
		FIRST provide a one-sentence comparison of the two summaries, explaining which you prefer and why. SECOND, on a new line, state only ``A'' or ``B'' to indicate your choice. Your response should use the format:  \\ 
		Comparison: \textless one-sentence comparison and explanation\textgreater \\
		Preferred: \textless``A'' or ``B''\textgreater
	\end{mdframed}
	\caption{Prompt template of the summarization task for Claude-5.5-Haiku and GPT-6-Luna.}
	\label{tab:summarization_prompt}
\end{table}

\begin{table}[ht]
	\centering
	\begin{mdframed}
		For the following query to a chatbot, which response is more helpful? A helpful response addresses the human's query in a way that moves the conversation in a beneficial direction, provides useful information or assistance, and avoids escalating problematic requests.
		
		Query: \\
		\{query\} 
		
		Response A: \\
		\{response\_0\}
		
		Response B: \\
		\{response\_1\}
		
		FIRST provide a one-sentence comparison of the two responses and explain which you feel is more helpful. SECOND, on a new line, state only ``A'' or ``B'' to indicate which response is more helpful. Your response should use the format: \\
		Comparison: \textless one-sentence comparison and explanation\textgreater \\
		More helpful: \textless ``A'' or ``B''\textgreater
	\end{mdframed}
	\caption{Prompt template of the single-turn dialogue task for Claude-5.5-Haiku and GPT-6-Luna.}
	\label{tab:dialogue_prompt}
\end{table}

\begin{table}[ht]
	\centering
	\begin{mdframed}
		Which of the following summaries does a better job of summarizing the most important points in the given forum post, without including unimportant or irrelevant details? A good summary is both precise and concise. 
		
		Post: \\
		\{post\}
		
		Summary A: \\
		\{summary\_0\}
		
		Summary B: \\
		\{summary\_1\}
		
		Answer with ``A'' or ``B'' only, without any other text.
	\end{mdframed}
	\caption{Prompt template of the summarization task for Qwen3-1.7B.}
	\label{tab:summarization_prompt_qwen}
\end{table}

\begin{table}[ht]
	\centering
	\begin{mdframed}
		For the following query to a chatbot, which response is more helpful? A helpful response addresses the human's query in a way that moves the conversation in a beneficial direction, provides useful information or assistance, and avoids escalating problematic requests.
		
		Query: \\
		\{query\} 
		
		Response A: \\
		\{response\_0\}
		
		Response B: \\
		\{response\_1\}
		
		Answer with ``A'' or ``B'' only, without any other text.
	\end{mdframed}
	\caption{Prompt template of the single-turn dialogue task for Qwen3-1.7B.}
	\label{tab:dialogue_prompt_qwen}
\end{table}

\end{document}